\pdfoutput=1
\documentclass{article}
\usepackage{iclr2027_conference,times}
\usepackage{threeparttable}
\usepackage{etoolbox,placeins}
\usepackage{visual/preprint}
\hypersetup{pdftitle={Pay Once, Certify Every Round: Anytime-Valid Verdicts for Self-Improving Programs}}

\hypersetup{pdfauthor={Ziming Wang, Jingmin Zhu, Yuheng He, Qingyu Yin, Xiuyu Li}}
\author{\parbox{\dimexpr\textwidth-2\tabcolsep\relax}{\raggedright
Ziming Wang\textsuperscript{1}\quad Jingmin Zhu\textsuperscript{2}\quad Yuheng He\textsuperscript{3}\\Qingyu Yin\textsuperscript{4}\quad Xiuyu Li\textsuperscript{5}\\[0.6em]
\normalfont\small
\textsuperscript{1}The Hong Kong University of Science and Technology\\\textsuperscript{2}Monash University\\\textsuperscript{3}Sun Yat-sen University\\\textsuperscript{4}National University of Singapore\\\textsuperscript{5}Renmin University of China}}
\newcommand{\preprintauthornames}{Ziming Wang\textsuperscript{1}, Jingmin Zhu\textsuperscript{2}, Yuheng He\textsuperscript{3}, Qingyu Yin\textsuperscript{4}, and Xiuyu Li\textsuperscript{5}}
\newcommand{\preprintaffiliations}{\textsuperscript{1}The Hong Kong University of Science and Technology, \textsuperscript{2}Monash University, \textsuperscript{3}Sun Yat-sen University, \textsuperscript{4}National University of Singapore, \textsuperscript{5}Renmin University of China}

\declaretheorem[name=Theorem]{theorem}
\declaretheorem[name=Proposition]{proposition}
\declaretheorem[name=Lemma]{lemma}

\newcommand{\method}{BranchCert}
\newcommand{\E}{\mathbb E}
\newcommand{\accept}{\textup{\textsc{Accept}}}
\newcommand{\nonebetter}{\textup{\textsc{None Better}}}
\newcommand{\undecided}{\textup{\textsc{Undecided}}}
\graphicspath{{figures/}}
\newcommand{\MatrixCertificateCost}{37}
\newcommand{\MatrixBettingCost}{25}
\newcommand{\MatrixHalvingCost}{48}
\newcommand{\MatrixBatchCost}{165}
\newcommand{\MatrixRaceCost}{78}
\newcommand{\MatrixBettingDecisionRate}{94}

\newcommand{\SweepNoneBetterCostRoundThree}{174}

\newcommand{\WrongSigmaWithoutRowShuffle}{20.8}
\newcommand{\SigmaShare}{38.2}
\newcommand{\SigmaRuns}{384}
\newcommand{\SigmaWrongRuns}{2}
\newcommand{\ExactShare}{44.7}
\newcommand{\WrappedShareMin}{39.4}
\newcommand{\WrappedShareMax}{41.4}
\newcommand{\WrappedHedgedRoundSaving}{5.2}
\newcommand{\WrappedHedgedRoundsLower}{23}
\newcommand{\WrappedAlphaRoundSaving}{2.7}
\newcommand{\WrappedAlphaRoundsLower}{22}

\newcommand{\ChainRounds}{48}
\newcommand{\ChainNoneBetterRounds}{41}
\newcommand{\ChainCells}{27{,}000}
\newcommand{\ChainExactShare}{39.9}
\newcommand{\ChainSigmaShare}{23.2}
\newcommand{\ChainHistoryShare}{19.4}
\newcommand{\ChainHistoryLateSaving}{22.0}
\newcommand{\ChainHistoryRuns}{384}
\newcommand{\ChainOrders}{8}
\newcommand{\ChainHeldoutTasks}{300}
\newcommand{\ChainRequirement}{0.02}
\newcommand{\ChainRankRounds}{42}
\newcommand{\ChainRankRho}{$-$0.64}
\newcommand{\ChainWeakest}{6}
\newcommand{\ChainWeakestResidual}{73}
\newcommand{\ChainResidualOthersMax}{60}
\newcommand{\ChainUnitSaving}{16.7}
\newcommand{\ChainUnitSavingError}{2.9}

\newcommand{\AblChainDecidesSigmaExactOnSigmaTrace}{1}
\newcommand{\PairedHistoryShare}{19.1}
\newcommand{\BehaviourHolding}{15}
\newcommand{\BehaviourScenarios}{15}
\newcommand{\BettingGridTotalRuns}{3,072}

\newcommand{\AlphaLowest}{0.01}
\newcommand{\AlphaHighest}{0.2}

\newcommand{\BanditGoodArmLucbShare}{94.9}
\newcommand{\BanditGoodArmLucbDecides}{54}
\newcommand{\BanditMinceShare}{13.6}
\newcommand{\BanditMinceWrong}{59.1}

\newcommand{\FreshHeldoutControl}{0.4}
\newcommand{\FreshHeldoutLargest}{13.8}
\newcommand{\FreshHeldoutLargestChain}{6}
\newcommand{\FreshHeldoutRequired}{4.9}
\newcommand{\FreshHeldoutTasksPerTarget}{500}
\newcommand{\LiveMismatchHistoryPercent}{5.7}
\newcommand{\LiveMismatchSigmaPercent}{6.6}
\newcommand{\LiveRescoredHistory}{8{,}860}
\newcommand{\LiveRescoredSigma}{11{,}136}
\newcommand{\LiveRoundHistoryCells}{7{,}856}
\newcommand{\LiveRoundSigmaCells}{9{,}595}
\newcommand{\FreshHeldoutControlMax}{1.6}
\newcommand{\FreshHeldoutControlMin}{0.0}

\newcommand{\ExampleCellsSigma}{4{,}886}
\newcommand{\ExampleCellsHistory}{4{,}729}
\newcommand{\HelmSigmaBelowExact}{23}
\newcommand{\HelmRoundCount}{24}
\newcommand{\HelmShareSigmaHistory}{37.8}
\newcommand{\HelmShareBranchcertBonferroniEagerParent}{43.1}
\newcommand{\HelmSequenceMin}{43.6}
\newcommand{\HelmSequenceMax}{44.7}
\newcommand{\HelmOrderSpreadSigma}{3.6}
\newcommand{\HelmOrderErrorSigma}{0.20}
\newcommand{\HelmRunMeanSigma}{37.6}
\newcommand{\LevelHistoryWrongMax}{1.6}
\newcommand{\LevelBeaconWrongMin}{32.7}
\newcommand{\LevelHalvingWrongMin}{75.5}
\newcommand{\SweepMedianMin}{9.4}
\newcommand{\SweepMedianMax}{56.8}
\newcommand{\SweepOthersMin}{47.0}
\newcommand{\SweepOthersMax}{52.8}
\newcommand{\GapChainsWithoutFinitePopulation}{7.6}
\newcommand{\GapChainsSigmaHZeroPredictionHistoryRanges}{3.8}
\newcommand{\GapHelmParentBilledPerCandidate}{49.8}
\newcommand{\PairedSigmaShare}{23.2}
\newcommand{\FreshGainError}{1.9}
\newcommand{\FreshChainsChangedWord}{five}
\newcommand{\LiveWindowMin}{64}
\newcommand{\LiveWindowMax}{256}
\newcommand{\LadderBudget}{60}
\newcommand{\LadderHighBudget}{90}
\newcommand{\LadderSigmaCorrect}{96.9}
\newcommand{\LadderOthersCorrect}{62.8}
\newcommand{\LadderHighWrongTiny}{15.1}
\newcommand{\LadderHighWrongMince}{16.4}
\newcommand{\LadderBanditFirstDecision}{70}
\newcommand{\LadderBanditFullCorrect}{54.2}
\newcommand{\LiveAcceptCells}{53}
\newcommand{\ReexecutionQwen}{37, 37 and 38}
\newcommand{\ReexecutionGranite}{53, 53 and 58}
\newcommand{\GridSpread}{2.4}
\newcommand{\GatesUnsettled}{4, 1, 2 and 4}
\newcommand{\GatesPositive}{2, 39, 17 and 51}
\newcommand{\GatesNegative}{10, 0, 16 and 0}
\newcommand{\RolloutsBinomial}{17}
\newcommand{\RolloutsUnion}{35}
\newcommand{\GateMinibatchTenWrongAccept}{78.6}
\newcommand{\GateMinibatchThirtyWrongAccept}{77.1}
\newcommand{\QualityOracleChange}{45.2}
\newcommand{\SettingsLargestDrop}{0.2}

\newcommand{\CertificateCpuMillisecondsPerRead}{0.7}
\newcommand{\CostCompleteSecondsHigh}{3{,}212}
\newcommand{\CostCompleteSecondsLow}{2{,}385}
\newcommand{\CostFastestSeconds}{1{,}735}
\newcommand{\CostFreshFlipsHigh}{1{,}309}
\newcommand{\CostFreshFlipsLow}{1{,}263}
\newcommand{\CostReplayRuns}{32}
\newcommand{\CostTokenShareMax}{62}
\newcommand{\CostTokenShareMin}{48}
\newcommand{\GateMinibatchWrongAccept}{57.8}
\newcommand{\GatePipelineMissed}{40}
\newcommand{\GateQualifyingRuns}{56}
\newcommand{\RefreshFullSaving}{0.0}
\newcommand{\RefreshHalfSaving}{6.1}
\newcommand{\RefreshQuarterSaving}{10.2}
\newcommand{\RefreshQuarterSavingMax}{15.0}
\newcommand{\RefreshQuarterSavingMin}{5.8}
\newcommand{\WrapperChainAlphaSaving}{15.6}
\newcommand{\WrapperChainControlVariateSaving}{18.7}
\newcommand{\WrapperChainControlVariateSavingError}{2.6}
\newcommand{\WrapperChainGainMax}{36}
\newcommand{\WrapperChainGainMin}{29}
\newcommand{\WrapperChainHedgedSaving}{8.0}

\newcommand{\BoundaryRuns}{288{,}000}
\newcommand{\GateLoopCompleteMinutesPerRound}{58}
\newcommand{\GateLoopCompletePerSeed}{+0.2, +7.8, and +9.8}
\newcommand{\GateLoopCompleteRounds}{4}
\newcommand{\GateLoopGepaMinibatchAccepted}{22}
\newcommand{\GateLoopGepaMinibatchWrong}{18}
\newcommand{\GateLoopGepaScoreMinutesPerRound}{10}
\newcommand{\GateLoopGepaScorePerSeed}{+0.3, +1.3, and +11.6}
\newcommand{\GateLoopSigmaHMinutesPerRound}{50}
\newcommand{\GateLoopSigmaHRounds}{12}
\newcommand{\QualityAdversarialCost}{5.8}
\newcommand{\QualityAllRuns}{15{,}360}
\newcommand{\StaleDifferHigh}{54}
\newcommand{\StaleDifferLow}{46}
\newcommand{\StaleHalted}{255}
\newcommand{\StaleOrders}{256}
\newcommand{\StaleWidenedSavingHigh}{12}
\newcommand{\StaleWidenedSavingLow}{11}
\newcommand{\SweControlTasks}{194}
\newcommand{\SweFinalCertificate}{77.6}
\newcommand{\SweFinalComplete}{78.8}
\newcommand{\SweMinibatchWrongAccept}{26.4}
\newcommand{\SweOrders}{512}
\newcommand{\SwePipelineMissed}{2{,}605}
\newcommand{\SwePipelineQualifyingRounds}{6{,}849}
\newcommand{\SweRequiredTasks}{306}
\newcommand{\SweRounds}{17}
\newcommand{\SweRuns}{8{,}704}
\newcommand{\SweSigmaShareOfComplete}{37}
\newcommand{\SweSystems}{135}

\newcommand{\ConstantsPseudoShareMax}{19.7}
\newcommand{\ConstantsPseudoShareMin}{19.4}
\newcommand{\ConstantsThreeQuartersHistoryChange}{1.01}
\newcommand{\SizeHistoryGrowth}{2.70}
\newcommand{\SizeMediumSigmaShareChainTwo}{54.7}
\newcommand{\SizeSigmaGrowth}{2.94}
\newcommand{\WidthFortyThreeSavingExact}{38.1}
\newcommand{\WidthFortyTwoQualifying}{24}
\newcommand{\WidthRounds}{24}
\newcommand{\WidthRuns}{384}
\newcommand{\WidthTenTwoQualifying}{9}
\newcommand{\WidthWrongMax}{0.52}

\title{Pay Once, Certify Every Round:\\Anytime-Valid Verdicts for Self-Improving Programs}
\preprintlink{\faGithub}{Code and data}{https://github.com/ZenAlexa/pay-once-certify-every-round}
\iclrfinalcopy
\begin{document}
\lhead{Preprint}
\begin{abstract}
A program that improves itself works in rounds: it proposes a batch of edited copies of itself as candidates, scores
them and the program it currently runs, the parent, on a fixed set of tasks, and keeps at most one candidate. Each score
costs one execution of a program on a task, so scoring every candidate on every task makes a round expensive, and deciding
from a small sample risks keeping the wrong program. We present \method{}, a certificate that decides a round from part of
its score table. It returns \accept{} for a named candidate that meets the required improvement over the parent,
\nonebetter{} when no candidate does, or \undecided{} when the budget runs out first. \method{} reads every candidate and the
shared parent along one random order of the tasks, so each parent score is paid for once and the tasks read so far form a
uniform random sample. Confidence sequences for sampling without replacement then let the round stop as soon as it is
decided, and they bound the error of both verdicts at any stopping time. A second form also reuses the scores that earlier
rounds already paid for. Before the round starts it predicts the score difference of every candidate on every
task, and it tests only what the prediction misses. Any prediction fixed in advance keeps the guarantee, and an accurate
one lowers the cost. On public evaluation tables, recorded rounds of a running self-improvement loop and live
re-executions, \method{} decides each round from part of its table with fewer wrong verdicts than the stated level allows,
and its mean cost falls as the loop accumulates paid scores. On six recorded loops in which Qwen3.5-9B rewrites its own
system prompt, the second form decides every round after reading \ChainHistoryShare\% of the score table on average, with no
wrong verdict, against \ChainExactShare\% when every unread score is bounded only by its lowest and highest value.
\end{abstract}

\section{Introduction}
\label{sec:introduction}

Self-improving programs propose edits of themselves and keep the edits their acceptance rule admits
\citep{duan2026recursive,zhang2026dgm,helix,agrawal2025gepa}. In the loops we study, a round proposes a batch of candidate
edits, evaluates them against the version the program runs, the parent, and keeps at most one. Each evaluation runs one
program on one task, so a round that scores every candidate on every task pays for the whole score table before it decides
anything. Two observations suggest that much of this table is unnecessary: a candidate far below the parent is decided after
a few tasks, and every candidate is compared with the same parent. This paper presents \method{}, a certificate that reads
the table one score at a time and stops as soon as the round is decided.

\begin{figure}[t]
\centering
\includegraphics[width=\textwidth]{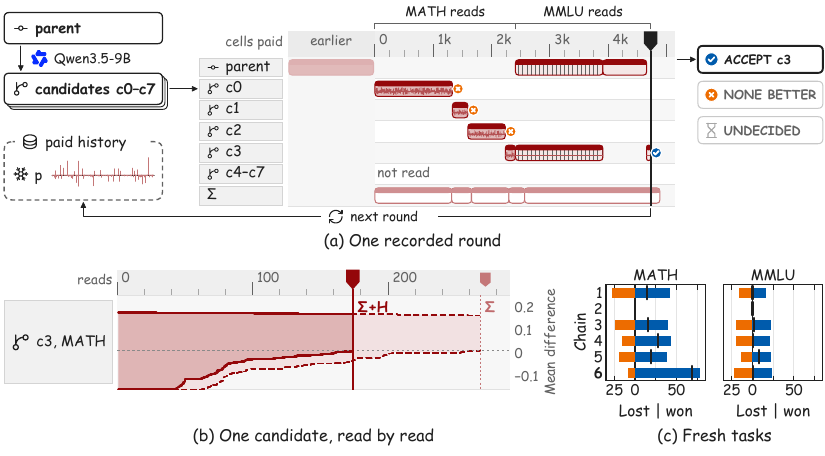}
\caption{One round decided from part of its table: round 3 of chain 6, under the reading order whose costs lie closest to
the round's medians. \method{}-$\Sigma$+H pays for \ExampleCellsHistory{} and \method{}-$\Sigma$ for \ExampleCellsSigma{} of the
\ChainCells{} cells. (a) The paid reads of \method{}-$\Sigma$+H in order; the clip before zero was paid in earlier rounds,
waveforms are the frozen prediction $p$, and the last lane is \method{}-$\Sigma$. (b) The interval of c3's mean paired MATH
difference after each read, solid for \method{}-$\Sigma$+H and dashed for \method{}-$\Sigma$; the grey dashed line is
the requirement. (c) Fresh tasks, of \FreshHeldoutTasksPerTarget{}, solved only by the final or only by the initial prompt;
ticks mark the net change.}
\label{fig:overview}
\end{figure}

We study rounds with a fixed, finite set of tasks and a score of zero or one for each program on each task. A candidate
qualifies when it meets the required improvement over the parent on every group of tasks, for example a gain on the targeted
tasks and no loss on control tasks. The round must name a qualifying candidate, conclude that none qualifies, or report that
its budget ran out. A small sample can give the wrong answer, and more so when the rule looks at the scores before it decides
to stop. Confidence sequences control this error for a single comparison under early stopping
\citep{howard2021confseq,waudbysmith2023betting,waudbysmith2020wor}, and the gates of current loops test one candidate at a
time \citep{ma2026validation,cai2026safeharness,shawn2026pace}, so neither controls the error of a decision about the whole
batch. Our question is whether a round can decide the whole batch at a controlled error and pay for each score only once.

\method{} has three forms. The exact form, \method{}-E, bounds every unread score by its lowest and highest value, which gives conditions
for both verdicts that hold with certainty. \method{}-$\Sigma$ reads every candidate along one shared random order of the
tasks. Each parent score is then paid for once and reused by every candidate, and the tasks a candidate has read form a
uniform sample of all its tasks, so a confidence sequence for sampling without replacement can decide the comparison early.
\method{}-$\Sigma$+H also uses the scores that earlier rounds paid for: before the order is drawn it predicts every paired
score difference of the current batch, and it bets only on what the prediction misses. The error guarantee holds whenever the
predictions, and ranges that contain the true differences, are fixed before the order is drawn, so an inaccurate prediction costs reads and never
validity. The guarantee covers the score table each round realizes, and the rounds of a loop share one error level.
\Cref{fig:overview} follows one recorded round from the first read to the verdict.

We test the certificates on public model-evaluation tables, on recorded chains in which a model rewrites its own system
prompt, and on live tool-use rounds. On the public tables the \method{} certificates read the least among the rules that
control their error and decide every round, and published confidence sequences read less inside the wrapper of
\method{} than one candidate at a time. On the chains the saving of \method{}-$\Sigma$+H over \method{}-$\Sigma$ grows as the
loop accumulates paid scores, with no wrong verdict, and at an equal evaluation budget a loop gated by \method{}-$\Sigma$+H
runs three times as many rounds as a loop that scores every candidate on every task. The accepted edits raise accuracy on
fresh tasks, and because the certificate reads only the score table, it applies unchanged to a loop whose candidates are the
submitted agents of the SWE-bench Verified leaderboard in date order \citep{chowdhury2024swebenchverified}.

\section{Related work}
\label{sec:related}

A self-improvement loop keeps the edit that its gate accepts, so the gate shapes how the loop evolves
\citep{romeroparedes2023funsearch,novikov2025alphaevolve,zelikman2023stop,robeyns2025sica,zhang2026dgm}. Published gates test
one candidate against a retained parent with fresh paired tests, retention tests, sequential tests or e-values
\citep{ma2026validation,cai2026safeharness,sea,wu2025sgm}, certify one candidate or one model update at a stated error level
\citep{shawn2026pace,balachandran2026discern,zhang2026slicewise}, or combine quality checks, rollback and validators
\citep{jiang2026harnessevolve,mao2026safeevolve,liu2026pairwisevalidator}, and practical prompt optimizers accept a candidate
on a small minibatch of tasks \citep{agrawal2025gepa}. Run once per candidate, such a gate pays for the parent's scores in
every comparison and reuses no score of earlier rounds. \method{} decides a whole batch against one shared parent row at one
error level, gives \nonebetter{} its own error bound, and reuses the paid history of the loop.

Confidence sequences built by betting hold at every stopping time \citep{howard2021confseq,ramdas2022savi,vovk2021evalues},
and their forms for sampling without replacement rest on the conditional law of a random permutation
\citep{waudbysmith2020wor,serfling1974sampling,stark2022alpha,shekhar2023riskaudits}. Recent evaluators run such sequences to
score one model on a fixed question set or to compare two agents, adding a surrogate, an active sampling rule or a learned
control variate \citep{zhou2026celeus,hsu2026sequentialeval,li2026avaivat}; fixed-budget evaluators predict the unread scores with a factor or
low-rank model of the score matrix and state their guarantee at the end of the budget \citep{wu2026faq,tolochinsky2026bestmodel}, and prediction-powered inference corrects model predictions with a small labeled
sample \citep{angelopoulos2023ppi,zrnic2024activeinference,tenzer2026semisupervisedbetting}. Each of them serves one stream
of scores under its own rule for choosing the next observation. \method{}-$\Sigma$ runs such
sequences for $K$ candidates and $Q$ targets jointly on one shared order, and \method{}-$\Sigma$+H builds its control variate
from the paid history of the loop and freezes it before the order is drawn.

Racing and best-arm identification select the best of several candidates
\citep{kalyanakrishnan2012lucb,karnin2013successivehalving,garivier2016trackandstop}, and good-arm identification,
1-identification and feasibility-constrained identification can also report that no arm passes a threshold
\citep{kano2019goodarm,li2026oneidentification,cai2025constrainedbai}. Comparisons against a shared control go back to
\citet{dunnett1955multiple}, and stochastic function evaluation asks which costly observations determine the value of a
function \citep{charikar2000priced,deshpande2013sbfe,ghuge2026correlations}. The bandit formulations sample arms with
replacement from fixed laws. A round of \method{} reveals a fixed finite table along a random order, accepts any candidate that meets its
paired requirements without ranking the candidates, and lets one parent cell serve every comparison.

Subset evaluation estimates the benchmark score of a model from a small set of tasks calibrated on earlier models
\citep{polo2024tinybenchmarks,das2026mince,huang2026zipbench}, and a developer can choose a fixed task panel under a monetary
budget \citep{conn2026deltaselect}. These methods fix which tasks are read and estimate scores with no error bound on a
decision about a batch. \method{} reads adaptively, decides a whole candidate set at a stated level and returns the exact
totals after a full read.

\section{Set-level certificates on a finite table}
\label{sec:problem}

A \emph{round} fixes a parent program $E_0$, candidates $E_1,\ldots,E_K$ written by a proposer, and $Q$ targets. Target $q$
owns $N_q$ tasks and a requirement $b_q$ on the mean paired score difference: a required target has $b_q>0$, and a control
target has $b_q\le0$ and asks for non-inferiority. Running candidate $k$ on task $i$ of target $q$ reveals a score
$y_{kqi}\in\{0,1\}$, and running the parent reveals $z_{qi}$; the complete score table $x$ collects all these scores. A
candidate's total paired difference on a target, the total requirement it has to clear and the set of qualifying candidates
are
\begin{equation}
 D_{kq}=\sum\nolimits_{i=1}^{N_q}\bigl(y_{kqi}-z_{qi}\bigr),\qquad T_q=N_qb_q,\qquad
 \mathcal S(x)=\{k:\ D_{kq}\ge T_q\ \ \forall q\}.
 \label{eq:eligible}
\end{equation}

A \emph{cell} is one program on one task, and a \emph{rollout} is one execution of a program on a task, whose score the table
assigns to its cell. Reading a cell means paying for its score. A rule reads cells one at a time, each chosen after seeing the
earlier ones; the \emph{history} $h$ lists the cells read so far in the round with their scores, and a budget $B$ caps the
reads. A round returns \accept{} for a named candidate, the claim that it lies in $\mathcal S(x)$; \nonebetter{}, the claim
that $\mathcal S(x)$ is empty; or \undecided{} when the budget runs out first, in which case the round keeps the parent. At level
$\alpha$, a wrong \accept{} and a wrong \nonebetter{} may each have probability at most $\alpha/2$, so a round errs with
probability at most $\alpha$. A \emph{chain} is a sequence of rounds $\rho=1,2,\ldots$ in which
each accepted candidate becomes the next parent. The \emph{paid history} of a round is the set of scores paid for before it
starts: in a chain, the cells that the rule read in its earlier rounds; on a public table, the published scores of other
models. A cell in the paid history costs nothing when read, and the paid parent cells enter $h$ before the round's first read.

The exact certificate asks which verdicts the unread cells leave open. Write $\mathcal X(h)$ for the tables that agree with $h$
on every read cell and keep every unread score in $[0,1]$. Let $O_{kq}(h)$ be the tasks of target $q$ on which candidate $k$
has been read and $O_q(h)$ those on which the parent has, and let
$R_{kq}(h)=\sum_{i\in O_{kq}(h)}y_{kqi}-\sum_{i\in O_q(h)}z_{qi}$ be the observed part of $D_{kq}$. An unread candidate score
adds at most one to the total and an unread parent score subtracts at most one, so
\begin{equation}
 L_{kq}(h)=R_{kq}(h)-\bigl(N_q-|O_q(h)|\bigr)\le D_{kq}\le R_{kq}(h)+\bigl(N_q-|O_{kq}(h)|\bigr)=U_{kq}(h).
 \label{eq:endpoints}
\end{equation}
Reading the parent on one task raises $L_{kq}$ by $1-z_{qi}$ and lowers $U_{kq}$ by $z_{qi}$ for all $K$ candidates at once, so
one parent row serves every comparison on its target.

\begin{restatable}[Exact set-level certificates]{theorem}{restatecertificates}
\label{thm:certificates}
For every history $h$ and candidate $k$, $k\in\mathcal S(x')$ for all $x'\in\mathcal X(h)$ exactly when $L_{kq}(h)\ge T_q$
for every $q$, and $\mathcal S(x')=\varnothing$ for all $x'\in\mathcal X(h)$ exactly when every $k$ has a target with
$U_{kq}(h)<T_q$. Both equivalences hold at every step of any adaptive reading rule, and the complete table yields one of the
two verdicts.
\end{restatable}

The completion that sets every unread candidate score to zero and every unread parent score to one attains every lower end of
\cref{eq:endpoints}, and the opposite completion every upper end, which gives necessity. The certificate depends on the
observed history alone, so every reading schedule and stopping rule keep it valid. Each round realizes one table $x$, which
stores the first execution or the cached score of every cell, so under a nondeterministic evaluator the verdicts are exact for
that realized table. \method{}-E runs the exact certificate with a planner: among
the plans of further reads that fit the remaining budget, it finds the one most likely, under a frozen prior on the unread
scores, to reach \accept{}, and reads that plan's first cell.

Rules that draw tasks with replacement can execute one task many times, and a repeated draw adds no information about a fixed
table.

\begin{restatable}[Distinct-cell cost]{proposition}{restateresampling}
\label{prop:resampling}
If a protocol draws task indices of a target with $N$ tasks independently of the table and reads the candidate and the
parent on each draw, every decision it computes is a function of the paired outcomes on the distinct tasks drawn, so $g$
draws resolve at most $\min\{g,N\}$ comparisons at a cost of at most $2\min\{g,N\}$ distinct cells.
\end{restatable}

Every rule is therefore charged one unit per distinct cell it reads outside its paid history. A round holds
$\sum_q(K+1)N_q$ distinct cells, the parent row counted once for all $K$ candidates, and the \emph{share read} of a run is the
number of cells it is charged for divided by this total.

\section{\method{}: certificates that stop early and reuse paid scores}
\label{sec:method}

\subsection{A shared random order and a lazy parent row}
\label{sec:sigma}

The exact interval of \cref{eq:endpoints} fills every unread cell with its worst value, so the exact certificate closes only
after most of a row has been read. \method{}-$\Sigma$ replaces the worst case by a confidence set that a random reading order
makes valid. Before any cell of the round is read, it draws for each target $q$ a \emph{reading order} $\pi_q$, uniform over
the $N_q!$ permutations of its tasks and independent across targets, and every row of that target reads a prefix of $\pi_q$.
The parent cell of a task is read the first time any candidate reaches the task, and once a candidate has read half of a
target the rest of the parent row is read, which tightens the exact intervals of all $K$ candidates at once. Given its prefix,
the unread tasks of a row are in uniformly random order among the remaining tasks of the target, which is the law of a finite
population drawn without replacement \citep{serfling1974sampling,waudbysmith2020wor}.

\subsection{Residual betting on the paid history}
\label{sec:history}

Fix a candidate $k$ and a target $q$, and write $d_i=y_{kqi}-z_{qi}$ for the paired difference on task $i$, so that
$D_{kq}=\sum_id_i$. Before the orders of a round are drawn, \method{}-$\Sigma$+H fixes from the paid history a prediction
$p_i$ of every $d_i$ and a range $[a_i,c_i]$ that contains it. In a chain, let $n_i$ be the number of earlier programs read on
task $i$, $s_i$ the sum of their scores, and $\bar s$ the mean score over all paid cells of the target. When the current parent
was read on task $i$ in an earlier round,
\begin{equation}
 p_i=\frac{s_i+\kappa\bar s}{n_i+\kappa}-z_{qi},\qquad [a_i,c_i]=[-z_{qi},\,1-z_{qi}]:
 \label{eq:prediction}
\end{equation}
the observed difficulty of the task, shrunk toward $\bar s$ with a pseudo-count $\kappa>0$, minus the known parent score. A
task without a paid parent cell keeps $p_i=0$ and $[a_i,c_i]=[-1,1]$, and setting every $p_i=0$ and every range to $[-1,1]$
recovers \method{}-$\Sigma$. \Cref{sec:exp-helm} gives the predictor for a public table.

The bets run on the residuals $e_t=d_{\pi(t)}-p_{\pi(t)}$ of the row, read along its order $\pi=\pi_q$. After $t$ reads the row
has observed $S_t=\sum_{j\le t}d_{\pi(j)}$ and $P_t=\sum_{j\le t}p_{\pi(j)}$, and $P=\sum_ip_i$ is known. Under the hypothesis
$D_{kq}=m$ the next task is uniform over the unread ones, so the residual has the known conditional mean
$\nu_t(m)=\bigl((m-S_{t-1})-(P-P_{t-1})\bigr)/(N_q-t+1)$, whatever the prediction. Two mixtures of betting processes test $m$
from each side,
\begin{equation}
 M^{\pm}_t(m)=\sum\nolimits_{\vartheta\in\Theta}w_\vartheta\prod\nolimits_{j\le t}\bigl(1\pm\lambda_{j,\vartheta}^{\pm}(m)\,(e_j-\nu_j(m))\bigr),
 \label{eq:sigma}
\end{equation}
with fixed weights $w_\vartheta\ge0$ summing to one over a finite grid $\Theta$ and predictable fractions
$\lambda_{j,\vartheta}^{\pm}(m)$ that scale the grid point $\vartheta$ by the range of the unread residuals, so every factor
stays nonnegative. $M^{+}$ grows when the true total exceeds $m$, and $M^{-}$ when it falls short. The lower set
$\mathcal C^{+}_t$ keeps the totals $m$ with $M^{+}_s(m)<2K/\alpha$ for all $s\le t$, the upper set $\mathcal C^{-}_t$ keeps
those with $M^{-}_s(m)<2Q/\alpha$, and each set is intersected with the exact interval $[L_{kq},U_{kq}]$, falling back to that
interval when the intersection is empty. The round returns \accept{} for $k$ when every lower set of $k$ lies at or above
$T_q$, excludes $k$ when some upper set of $k$ lies below $T_q$, and returns \nonebetter{} once every candidate is excluded.
The shared order, the lazy parent row and these one-sided levels form the set-level \emph{wrapper} of \method{}, inside which
any confidence sequence for sampling without replacement can take the place of $M^{\pm}$.
Every prediction keeps the bets fair at the true total, since $\nu_t(m)$ is exact for every prediction; an accurate one shrinks
the spread of the residuals, so the wealth at a false total tends to grow faster and the sets to close earlier.

\begin{restatable}[Finite-population certificates with paid history]{theorem}{restatesigma}
\label{thm:sigma}
Fix a round with a binary table. Fix predictions, ranges containing every paired difference, and mixture weights before the
orders are drawn, and draw each order uniformly given the table and this side information. Then (i) for each candidate $k$
and target $q$, $M^{\pm}_t(D_{kq})$ are nonnegative martingales starting at one in the row's own filtration, so $D_{kq}$
leaves the lower sets at some time with probability at most $\alpha/(2K)$ and the upper sets with probability at most
$\alpha/(2Q)$; (ii) under any schedule across rows and any stopping rule on the joint history, a wrong \accept{} and a wrong
\nonebetter{} each have probability at most $\alpha/2$; (iii) on the same cells the certificate decides whenever the exact
certificate of \cref{thm:certificates} decides, and a full read returns the exact totals; (iv) when the rounds
$\rho=1,2,\ldots$ of a chain each fix their side information from their paid history, rounds at levels $\alpha_\rho$ with
$\sum_\rho\alpha_\rho\le\alpha$ err in any round with probability at most $\alpha$.
\end{restatable}

At $m=D_{kq}$ both mixtures in \cref{eq:sigma} are martingales, since $\nu_t(D_{kq})$ is the exact conditional mean of $e_t$.
Ville's inequality \citep{ville1939etude,howard2021confseq} then bounds the chance that the true total ever leaves the lower
sets by $\Pr[\exists t:M^{+}_t(D_{kq})\ge2K/\alpha]\le\alpha/(2K)$, and the upper sets by $\alpha/(2Q)$. Part (ii) needs one
more step. A global schedule stops each row at a random time that depends on every other row, which is in general no stopping
time of the row's own filtration.

\begin{restatable}[Path lemma]{lemma}{restatepath}
\label{lem:path}
Let $(\mathcal C_t)_{t\le N}$ be random sets and $\xi$ a fixed value with $\Pr[\exists t\le N:\xi\notin\mathcal C_t]\le\delta$.
Then $\Pr[\xi\notin\mathcal C_\tau]\le\delta$ for every random index $\tau\in\{0,\ldots,N\}$, from any filtration.
\end{restatable}

The lemma holds because the event $\{\xi\notin\mathcal C_\tau\}$ lies inside the event that $\xi$ leaves the set at some
time. A wrong \accept{} must exclude the true total at one failing target of a non-qualifying candidate, a witness fixed by
the complete table before the orders are drawn, and a union over at most $K$ such witnesses costs $\alpha/2$. A wrong
\nonebetter{} must exclude one qualifying candidate through one of its $Q$ targets, again at cost $\alpha/2$. The witnesses
belong to the proof; the evaluator uses the observed confidence sets alone.

\subsection{Conditions on the paid history}
\label{sec:history-conditions}

\Cref{thm:sigma} asks six things of the paid history and the evaluator. (1) Predictions, ranges and mixture weights come
from the paid history, which is fixed before the round starts, and each order is then drawn uniformly given the table and this
side information. (2) Under a nondeterministic evaluator the realized table does not depend on the reading order, which holds
when a score does not depend on the batch in which it is decoded. (3) Every range contains the
true paired difference: the evaluator caches scores by program and task, so a recorded parent score equals the current one,
and a task without a paid parent score keeps the range $[-1,1]$. (4) Bets and residual ranges use the frozen predictions and
the row's own prefix, so parent cells read during the round enter the exact interval and leave the bets unchanged. (5) Scores
are binary, so the integer grid of hypothesized totals contains the true total; the total requirement $T_q$ may be
fractional. (6) The levels of the rounds of a chain sum to at most $\alpha$.

\section{Experiments}
\label{sec:experiments}

Every rule reads the same complete tables, is charged one unit per distinct cell and spends one level per round, so a
comparison differs only in the rule. The compared rules form five families: without-replacement (WoR) betting sequences
\citep{waudbysmith2023betting,waudbysmith2020wor,stark2022alpha,shekhar2023riskaudits}, sequential and active evaluators
\citep{zhou2026celeus,hsu2026sequentialeval,li2026avaivat,tolochinsky2026bestmodel,wu2026faq}, per-candidate gates
\citep{balachandran2026discern,zhang2026slicewise,shawn2026pace,ma2026validation,sea,agrawal2025gepa}, bandit, subset and racing
rules \citep{kano2019goodarm,li2026oneidentification,polo2024tinybenchmarks,das2026mince,karnin2013successivehalving,kalyanakrishnan2012lucb},
and rules without error control, score-and-keep evaluation \citep{zhang2026dgm,robeyns2025sica} and a BEACON-style stopping
rule that we construct after \citet{wan2025beacon}. A dagger marks every rule we construct from a cited idea, and the exact
certificate reads either the parent row first (parent-first) or the candidate rows first (candidate-first). A rule that judges
one candidate runs inside the \emph{Bonferroni wrapper}, which gives it level $\alpha/(KQ)$ on each candidate--target pair,
audits the candidates in order, accepts the first one certified on every target and returns \nonebetter{} when every
candidate is excluded. The published without-replacement sequences run in both wrappers, the wrapper of \method{} and the
Bonferroni wrapper with two-sided levels and an \emph{eager} parent, which reads the whole parent row before any candidate.

A run \emph{decides} when it returns \accept{} or \nonebetter{}, and a decided run is \emph{wrong} when it accepts a candidate
outside $\mathcal S(x)$ or returns \nonebetter{} when $\mathcal S(x)$ is nonempty, the two errors of \cref{thm:sigma}. We report
the share read, the share of runs that decide and the share of decided runs that are wrong. The \emph{saving} of a rule $A$
over a rule $B$ is $100(1-C_A/C_B)$, with $C$ the number of cells charged. On the chains, shares read and savings are averaged
over runs unless the six chains are taken as units, and on HELM Lite they are the median over the orders of a round,
averaged over rounds.

\subsection{Public tables}
\label{sec:exp-helm}

HELM Lite publishes per-instance scores of 95 models on six benchmark families \citep{liang2023helmlite,liang2023helm}. A
round takes one family as the required target ($b=0.02$) and another as the control target ($b=-0.02$), a parent at one of
four quantiles of required accuracy, and the parent's ten nearest neighbors in required accuracy as candidates. Six
(required, control) pairs and four quantiles give 24 rounds, each replayed under 16 reading orders at $\alpha=0.05$. The paid
history of a round is the published scores of the other models: \method{}-$\Sigma$+H predicts a program's score on a task by
the other models' scores on it, averaged with weights that grow with each model's agreement with the program on the families
outside the round, and takes $p_i$ as the candidate's prediction minus the parent's, times a shrinkage factor fitted on the
other models and clipped to $[0,1]$, with every range $[-1,1]$.

\begin{figure}[t]
\centering
\includegraphics[width=\textwidth]{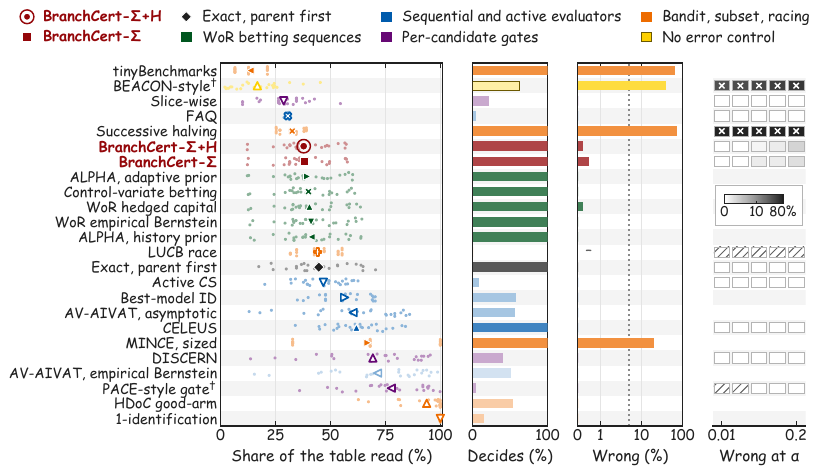}
\caption{Every rule on the 24 HELM Lite rounds at $\alpha=0.05$ (16 reading orders each), sorted by share read, with the
published without-replacement sequences inside the wrapper of \method{}. Dots are rounds and the large marker is their mean; a
hollow mark decides only part of its runs. The dashed line is $\alpha$, and a dash marks a rule that never decides. Last
column: wrong verdicts at five levels for the rules of the level sweep, crossed where above the level and hatched where the
rule decides no run.}
\label{fig:helm}
\end{figure}

On these tables the set-level certificates are the cheapest rules that decide every run within their error level
(\Cref{fig:helm}). \method{}-$\Sigma$ reads \SigmaShare\% of a table, decides every run and errs in \SigmaWrongRuns{} of
\SigmaRuns{} runs. \method{}-$\Sigma$+H reads almost the same, \HelmShareSigmaHistory\%. Each compared rule that reads less than the two certificates either errs in more than $\alpha$
of its decided runs (tinyBenchmarks, successive halving and BEACON-style stopping) or decides a minority of its runs
(slice-wise certification and FAQ). When every rule stops at \LadderBudget\% of each table, the smallest common budget at which
\method{}-$\Sigma$ returns the complete-table verdict in at least 95\% of the runs, it does so in \LadderSigmaCorrect\% and no
bandit or subset rule in more than \LadderOthersCorrect\%.

Deciding the batch as a set carries most of the saving over the exact certificate, which reads \ExactShare\%. In the
Bonferroni wrapper with an eager parent and two-sided levels, the published sequences read \HelmSequenceMin\% to
\HelmSequenceMax\%, and \method{}-$\Sigma$ reads \HelmShareBranchcertBonferroniEagerParent\%. Inside the wrapper of
\method{}, the same published sequences read \WrappedShareMin\% to \WrappedShareMax\%, and \method{}-$\Sigma$ reads less than
each of them, on average \WrappedHedgedRoundSaving\% less per round than hedged capital and \WrappedAlphaRoundSaving\% less than
adaptive ALPHA, the strongest of them on these tables.

The error control holds across levels. From $\alpha=\AlphaLowest$ to \AlphaHighest{} both certificates decide every run and
keep their wrong verdicts within $\alpha/2$, at most \LevelHistoryWrongMax\%, and BEACON-style stopping and successive
halving err in at least \LevelBeaconWrongMin\% and \LevelHalvingWrongMin\% of their decided runs at every level.

The rounds above are the hardest setting of a difficulty sweep of 216 rounds over three ways of choosing the ten
candidates (nearest neighbors, a uniform draw and an ability-stratified draw) and mean requirements of 0.02, 0.05 and 0.10.
With nearest candidates the median saving of \method{}-$\Sigma$ over the exact certificate grows from \SweepMedianMin\% at
$b=0.02$, the smallest of the sweep, to \SweepMedianMax\% at $b=0.10$, and with the other two choices it lies between
\SweepOthersMin\% and \SweepOthersMax\% at every requirement. With 5 to 40 nearest candidates and two or three targets,
\method{}-$\Sigma$ reads less than the exact certificate and than hedged capital and adaptive ALPHA in the Bonferroni wrapper
with an eager parent in every setting, and its saving over the exact certificate is largest, \WidthFortyThreeSavingExact\% per round on average, with 40
candidates and three targets.

\subsection{Self-improvement chains}
\label{sec:exp-chain}

Qwen3.5-9B \citep{qwen2026qwen35} rewrites its own system prompt in six chains of eight rounds. Each round proposes eight
candidates from the current parent and scores every program on 1{,}500 required MATH tasks \citep{hendrycks2021math} of levels
3 to 5 and 1{,}500 control MMLU tasks \citep{hendrycks2021mmlu}, a table of \ChainCells{} cells, at mean requirements
$b=\ChainRequirement$ on MATH and $-\ChainRequirement$ on MMLU. We record the complete table of every round and replay each
certificate on it under \ChainOrders{} reading orders at levels $\alpha_\rho=0.05/(\zeta(2)\rho^2)$, with
$\zeta(2)=\sum_{\rho\ge1}\rho^{-2}$, which sum to $0.05$ over a chain. Each certificate starts every chain from an empty paid
history and adds the cells it reads in a round to its own paid history for the later rounds.

\begin{figure}[t]
\centering
\includegraphics[width=\textwidth]{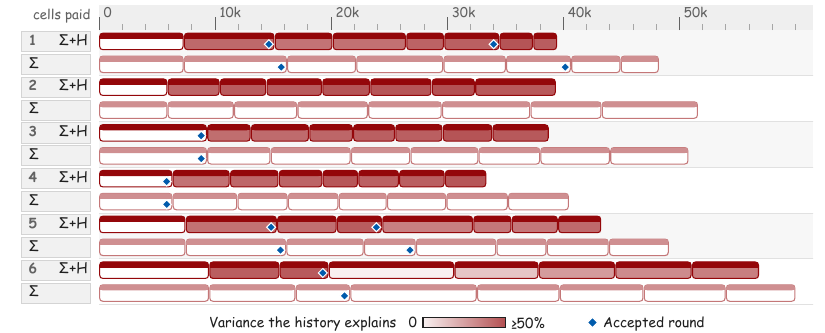}
\caption{The six recorded chains, numbered 1 to 6 from the top, as timelines of the cells each certificate pays for, one
clip per round (the mean over \ChainOrders{} reading orders). A \method{}-$\Sigma$+H clip is shaded by the share of the paired-difference variance its prediction explains.}
\label{fig:chains}
\end{figure}

The paid history makes the later rounds of a chain cheaper (\Cref{fig:chains}). In the first round no history exists and the
two certificates coincide. Over its eight rounds \method{}-$\Sigma$+H reads less than \method{}-$\Sigma$ in every chain,
\ChainHistoryShare\% of a table against \ChainSigmaShare\%, and \ChainExactShare\% for Exact, $\Sigma$ order, the exact
interval read along the schedule of \method{}-$\Sigma$ (\Cref{tab:chain}). Taking the six chains as units, the saving of \method{}-$\Sigma$+H over
\method{}-$\Sigma$ averages \ChainUnitSaving\% (standard error \ChainUnitSavingError{} over the chains); pooled over chains and
orders, it is \ChainHistoryLateSaving\% over rounds five to eight. The saving comes from the residual bet: when both
certificates start each round from the cells \method{}-$\Sigma$ read before, \method{}-$\Sigma$+H reads
\PairedHistoryShare\% against \PairedSigmaShare\%.

\begin{table}[t]
\caption{The six chains. Accepted counts the rounds of the recorded chain that accepted a candidate. Shares read average the
rounds and the \ChainOrders{} reading orders of a chain, with standard errors over the orders in parentheses; Exact, $\Sigma$
order reads the exact interval along the schedule of \method{}-$\Sigma$. Saving is the reduction in cells read by
\method{}-$\Sigma$+H against \method{}-$\Sigma$, a ratio of means that the All row takes over the pooled runs. MATH gives the
accuracy of the first and the last prompt on the \ChainHeldoutTasks{} held-out MATH tasks of the chain. Each row's lowest share
is bold, the second underlined.}
\label{tab:chain}
\centering
\small
\setlength{\tabcolsep}{3.6pt}
\begin{tabular}{@{}l r r r r r r r@{}}
\toprule
& & \multicolumn{3}{c}{Share read (\%)\,$\downarrow$} & & \multicolumn{2}{c}{MATH (\%)} \\
\cmidrule(lr){3-5}\cmidrule(l){7-8}
Chain & Accepted & Exact, $\Sigma$ order & \method{}-$\Sigma$ & \method{}-$\Sigma$+H & Saving (\%)\,$\uparrow$ & First & Last \\
\midrule
1 & 2 & 65.8 (0.1) & {\secondbest{22.3}} (0.4) & \textbf{18.3} (0.3) & 18.2 (1.2) & 78.3 & 79.7 \\
2 & 0 & 37.9 (0.1) & {\secondbest{23.9}} (0.4) & \textbf{18.2} (0.3) & 23.7 (1.5) & 78.7 & 78.7 \\
3 & 1 & 36.2 (0.1) & {\secondbest{23.5}} (0.5) & \textbf{17.9} (0.3) & 23.7 (1.5) & 82.0 & 83.3 \\
4 & 1 & 33.0 (0.0) & {\secondbest{18.7}} (0.8) & \textbf{15.5} (1.0) & 17.6 (3.5) & 81.7 & 83.7 \\
5 & 2 & 33.9 (0.1) & {\secondbest{22.7}} (0.3) & \textbf{20.0} (0.4) & 11.9 (2.1) & 82.0 & 84.7 \\
6 & 1 & 32.4 (0.1) & {\secondbest{27.8}} (0.7) & \textbf{26.3} (0.4) & 5.2 (2.4) & 81.0 & 93.7 \\
\midrule
All & 7 & 39.9 (0.0) & {\secondbest{23.2}} (0.2) & \textbf{19.4} (0.1) & 16.4 (0.7) & 80.6 & 83.9 \\
\bottomrule
\end{tabular}
\end{table}

The saving follows how much of the paired differences the prediction explains. Over the \ChainRankRounds{} rounds with paid
history it falls as the unexplained share of their variance rises (Spearman correlation \ChainRankRho{}), and chain
\ChainWeakest{}, whose history leaves \ChainWeakestResidual\% of the variance unexplained against at most
\ChainResidualOthersMax\% in the other chains, saves the least. The quality of the prediction moves the cost and leaves the
verdicts unchanged: a prediction that negates the true differences reads \QualityAdversarialCost\% more than no prediction,
one equal to them reads \QualityOracleChange\% less, and none of the \QualityAllRuns{} runs of five predictors returns a wrong
verdict.

Inside the wrapper of \method{}, with the chains as units, \method{}-$\Sigma$+H reads \WrapperChainControlVariateSaving\%
less (standard error \WrapperChainControlVariateSavingError) than control-variate betting with the same history prediction as
its control variate, the strongest published sequence on the chains, and less in every chain; \method{}-$\Sigma$ reads
\WrapperChainHedgedSaving\% less than hedged capital and \WrapperChainAlphaSaving\% less than adaptive ALPHA. Moving any of
these sequences, the two of \method{} included, from the Bonferroni wrapper with an eager parent into the wrapper of \method{}
lowers its share read by \WrapperChainGainMin\% to \WrapperChainGainMax\%.

No run of either certificate disagrees with the complete table, in \ChainHistoryRuns{} runs each, and \ChainNoneBetterRounds{}
of the \ChainRounds{} rounds end in \nonebetter{}, a verdict outside the answers of a two-model gate. The confidence sequences
make these verdicts: on the cells \method{}-$\Sigma$ reads, the exact interval alone decides
\AblChainDecidesSigmaExactOnSigmaTrace\% of the runs.

\subsection{Gates of practical loops}
\label{sec:exp-gates}

Practical loops admit candidates on small minibatches. The gate of GEPA$^{\dagger}$ \citep{agrawal2025gepa}, which we extend
with a control minibatch, accepts the first candidate that strictly beats the parent on a fresh minibatch of three required
tasks with no loss on the control minibatch, and its pipeline$^{\dagger}$ scores every candidate that passes the minibatch on all tasks and accepts the qualifier with the largest
required gain. Replayed on the recorded chain rounds (\Cref{tab:gates}), the minibatch gate reads under 1\% of a
table and accepts a non-qualifying candidate in \GateMinibatchWrongAccept\% of the runs, and with minibatches of 10 and 30
tasks in \GateMinibatchTenWrongAccept\% and \GateMinibatchThirtyWrongAccept\%. The pipeline reads about as much as \method{}-$\Sigma$+H and never accepts wrongly, and it returns \nonebetter{} in
\GatePipelineMissed{} of the \GateQualifyingRuns{} runs whose round holds a qualifying candidate, because the minibatch rejects
every qualifying candidate first.

\begin{table}[t]
\caption{Gates on the Qwen3.5-9B chains, with the GEPA rules$^{\dagger}$ of \citet{agrawal2025gepa} extended with a control minibatch. Replay: 384 runs on the recorded rounds (\%), where full evaluation defines the correct verdict and Qual.\ keeps the 56 runs of rounds with a qualifying candidate; the random subset of matched size reads one random task subset per target, of the size \method{}-$\Sigma$+H read. Own loop: accuracies are means over three seeds of 12 rounds, 4 for full evaluation; Accepted totals the acceptances over the seeds, with the non-qualifying ones in parentheses; Required and Control give the MATH and MMLU test accuracy (\%) of the final parent with its change against the loop's own evaluation of the initial prompt; and the last column the required difference in points $\pm$ twice its standard error, with its verdict: ahead beyond twice the standard error with control accuracy at most 2 points lower, tied within it.}
\label{tab:gates}
\centering
\footnotesize
\setlength{\tabcolsep}{2.55pt}
\begin{tabular}{@{}l r r r r r r r l@{}}
\toprule
& \multicolumn{4}{c}{Replay on recorded rounds (\%)} & \multicolumn{4}{c}{Own loop at one budget} \\
\cmidrule(lr){2-5}\cmidrule(l){6-9}
Gate & Read & Correct\,$\uparrow$ & \stack{Wrong}{\accept{}} & Qual. & Accepted & Required\,$\uparrow$ & Control & $\Sigma$+H${}-{}$gate \\
\midrule
\method{}-$\Sigma$+H & 19.4 & \textbf{100.0} & 0.0 & 100.0 & 3 (0) & \textbf{92.4} (+11.2) & 86.4 (+1.2) & \textemdash \\
GEPA$^{\dagger}$ & & & & & & & & \\
\quad pipeline & 18.0 & {\secondbest{89.6}} & 0.0 & 28.6 & 4 (0) & {\secondbest{88.0}} (+7.5) & 85.5 ($-$0.1) & +4.4 $\pm$ 7.2, tied \\
\quad gate, 3 tasks & 0.2 & 38.5 & 57.8 & 26.8 & 22 (18) & 80.3 ($-$0.8) & 86.1 (+0.7) & +12.1 $\pm$ 4.8, ahead \\
\quad gate, 10 tasks & 0.6 & 19.3 & 78.6 & 25.0 & \textemdash & \textemdash & \textemdash & \textemdash \\
\quad gate, 30 tasks & 1.9 & 21.9 & 77.1 & 14.3 & \textemdash & \textemdash & \textemdash & \textemdash \\
\begin{tabular}[t]{@{}l@{}}Random subset,\\matched size\end{tabular} & 19.3 & 78.9 & 16.9 & 62.5 & \textemdash & \textemdash & \textemdash & \textemdash \\
Full evaluation & \textemdash & \textemdash & \textemdash & \textemdash & 3 (0) & 86.5 (+5.3) & 86.6 (+1.2) & +5.9 $\pm$ 5.8, ahead \\
\bottomrule
\end{tabular}
\end{table}

In a loop of its own, each gate also chooses the parents of the next candidates. For each of three seeds, four loops start from the initial prompt and task split of a recorded chain and propose eight candidates per round
with the proposer of the chains; only the gate differs, and every loop has a budget of four complete tables, counting every
executed cell, and at most 12 rounds. Full evaluation, which scores every candidate on every task, pays for
\GateLoopCompleteRounds{} rounds, and the \method{}-$\Sigma$+H loop for all \GateLoopSigmaHRounds{}. The \method{}-$\Sigma$+H loop ends ahead of full evaluation in MATH test accuracy, by more than twice the standard error of
the per-seed differences, and tied with the GEPA pipeline.
The minibatch gate accepts \GateLoopGepaMinibatchAccepted{} candidates over the three seeds, \GateLoopGepaMinibatchWrong{} of them non-qualifying,
and ends below its initial prompt.

The certificate reads only the score table, so it applies unchanged to loops that edit code against fixed tasks
\citep{zhang2026dgm}. On the public SWE-bench Verified leaderboard
\citep{jimenez2024swebench,chowdhury2024swebenchverified}, \SweSystems{} submitted agents, ordered by date, form the stream:
the oldest is the initial parent, and the others arrive eight per round as candidates. Each gate runs its own loop with the issues annotated as taking at least fifteen minutes to fix as
the required target and the shorter ones as control. \method{}-$\Sigma$ reads \SweSigmaShareOfComplete\% of the cells that
full evaluation reads per loop, and none of the certificates returns a wrong verdict in any of its \SweRuns{} runs; the
minibatch gate accepts a non-qualifying agent in \SweMinibatchWrongAccept\% of the runs, and the GEPA pipeline misses a
qualifying agent in \SwePipelineMissed{} of the \SwePipelineQualifyingRounds{} runs whose round holds one. The certificate
loops end at an agent that resolves \SweFinalCertificate\% of the issues, and full evaluation, which accepts the qualifying
agent with the largest required gain, at \SweFinalComplete\%.

\subsection{Removals of the mechanism}
\label{sec:exp-ablation}

The finite-population correction, the single paid parent row and the residual bet each carry part of the saving
(\Cref{tab:ablation}). Betting against the with-replacement mean of the unread residuals drops the finite-population
correction, and with it a step of the proof; the share read of \method{}-$\Sigma$+H rises by
\GapChainsWithoutFinitePopulation{} points on the chains. Zero predictions with the history ranges kept add
\GapChainsSigmaHZeroPredictionHistoryRanges{} points, which returns the share of \method{}-$\Sigma$, so the paid history saves
through the residual bet. Charging each parent cell once per candidate, as a per-candidate gate does, costs most where many
candidates share a parent: \GapHelmParentBilledPerCandidate{} points on HELM Lite, with ten candidates per round. An eager parent, two-sided levels and
paired-only bounds, in which a parent cell tightens only the interval of the candidate that read it, also raise the share,
and unscaled betting fractions change it little. A fixed reading order breaks the uniform order that \cref{thm:sigma}
assumes and errs in \WrongSigmaWithoutRowShuffle\% of the decided HELM runs. Running \method{}-$\Sigma$ at level $10\alpha$
gives up the level-$\alpha$ guarantee for a smaller share read, and its wrong verdicts on HELM Lite stay below $\alpha/2$. Four fixed grids of betting fractions move the HELM share by at most \GridSpread{} points, and for
pseudo-counts $\kappa$ from 0.5 to 8 \method{}-$\Sigma$+H reads \ConstantsPseudoShareMin\% to \ConstantsPseudoShareMax\% of
each chain table.

\begin{table}[t]
\caption{One change to the mechanism at a time, on the 384 runs of the chains and the 24 HELM Lite rounds (16 orders each).
Bold and underline rank the shares of the variants that keep the guarantee and decide every run; a dash marks a change not
run on the chains.}
\label{tab:ablation}
\centering
\small
\setlength{\tabcolsep}{2.7pt}
\begin{tabular}{@{}l r r r r r r@{}}
\toprule
& \multicolumn{3}{c}{Chains (\%)} & \multicolumn{3}{c}{HELM Lite (\%)} \\
\cmidrule(lr){2-4}\cmidrule(l){5-7}
Variant & Read\,$\downarrow$ & Decides\,$\uparrow$ & Wrong\,$\downarrow$ & Read\,$\downarrow$ & Decides\,$\uparrow$ & Wrong\,$\downarrow$ \\
\midrule
\method{}-$\Sigma$+H & \textbf{19.4} & 100 & 0.0 & \textbf{37.8} & 100 & 0.3 \\
\quad zero prediction, history ranges & {\secondbest{23.2}} & 100 & 0.0 & {\secondbest{38.2}} & 100 & 0.5 \\
\quad parent cells charged per candidate & 27.0 & 100 & 0.0 & 87.6 & 100 & 0.3 \\
\midrule
\method{}-$\Sigma$ & {\secondbest{23.2}} & 100 & 0.0 & {\secondbest{38.2}} & 100 & 0.5 \\
\quad parent cells charged per candidate & 30.5 & 100 & 0.0 & 87.9 & 100 & 0.5 \\
\quad eager parent & \textemdash & \textemdash & \textemdash & 41.5 & 100 & 0.5 \\
\quad two-sided levels & \textemdash & \textemdash & \textemdash & 40.3 & 100 & 0.0 \\
\quad unscaled bets & \textemdash & \textemdash & \textemdash & 38.4 & 100 & 0.3 \\
\quad paired-only bounds & \textemdash & \textemdash & \textemdash & 62.1 & 100 & 0.0 \\
\quad exact, $\Sigma$ order & 39.9 & 100 & 0.0 & 44.6 & 100 & 0.0 \\
\quad exact interval on the $\Sigma$ trace & 23.2 & 1 & 0.0 & 38.2 & 10 & 0.0 \\
\midrule
\multicolumn{7}{@{}l}{\textit{Changes outside the proof of \Cref{thm:sigma}}} \\
\method{}-$\Sigma$+H, no finite-population correction & 27.0 & 100 & 0.0 & 42.1 & 100 & 0.0 \\
\method{}-$\Sigma$, no finite-population correction & \textemdash & \textemdash & \textemdash & 42.2 & 100 & 0.0 \\
\method{}-$\Sigma$, fixed reading order & \textemdash & \textemdash & \textemdash & 27.7 & 100 & 20.8 \\
\midrule
\multicolumn{7}{@{}l}{\textit{A change that gives up the level-$\alpha$ guarantee}} \\
\method{}-$\Sigma$ at level $10\alpha$ & \textemdash & \textemdash & \textemdash & 32.8 & 100 & 2.3 \\
\bottomrule
\end{tabular}
\end{table}

\subsection{Live rounds, fresh executions and fresh tasks}
\label{sec:exp-live}

Granite-4.2-8B \citep{granite2026model} proposes candidates for a tool-use program, and each round scores the parent and up
to twelve candidates on 15 required and 5 control tasks of ToolFailBench \citep{toolfailbench2026}, disjoint from the tasks
used to build the candidates. The seven complete live tables, of 40 to 260 cells, are the four rounds of a first Granite run
(R1 to R4), the ten candidates of R4 scored by Qwen3.5-9B, and rounds 6 and 8 of a later Granite run (R6, R8).

\begin{figure}[t]
\centering
\includegraphics[width=\textwidth]{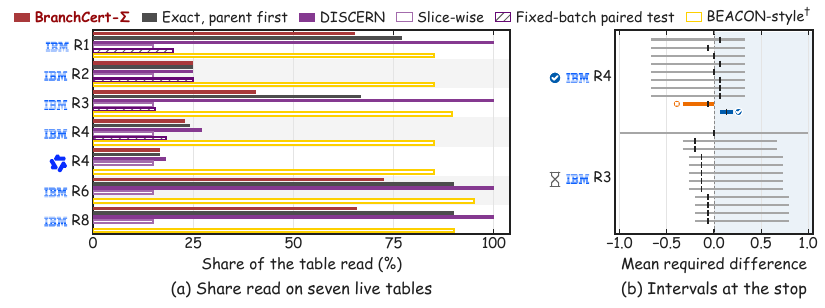}
\caption{Live tool-use rounds. The IBM logo marks tables scored by Granite-4.2-8B and the Qwen logo the table scored by
Qwen3.5-9B. (a) An outlined bar decides no run; the fixed-batch paired test, hatched, ran on Granite R1 to R4. (b) Required intervals of
every candidate when the exact certificate stops, \method{}-E on Granite R4 (\accept{}) and the candidate-first order on R3
(\undecided{}): blue certified, orange excluded, grey open; ink ticks are complete-table values, and the shaded side of the
dashed requirement qualifies.}
\label{fig:live}
\end{figure}

The certificates hold on live tables and return all three verdicts (\Cref{fig:live}). \method{}-$\Sigma$ agrees with the
complete table on all seven and reads fewer cells than the parent-first exact certificate on six and as many on the seventh; on
Granite R3 it certifies \nonebetter{} for twelve candidates, none of which qualifies. Granite R3 ran live at a budget of 220 of
its 260 cells and Granite R4 at 80 of its 220: \method{}-E accepts on Granite R4 after \LiveAcceptCells{} cells, and on Granite
R3 the candidate-first order spends its budget and returns \undecided{}. DISCERN reads the whole table whenever no candidate
qualifies, and slice-wise certification and BEACON-style stopping decide no live round.

Replay reuses recorded scores, so both certificates also ran on a fresh execution of round 2 of chain 1, a
\ChainCells-cell table of one parent and eight candidate prompts, with every response generated again under the frozen
decoding settings and the evaluator cache serving the cells paid in round 1. Each run executes cells in windows ahead of its
reads: \method{}-$\Sigma$ executed \LiveRescoredSigma{} cells and decided after reading \LiveRoundSigmaCells{}, and
\method{}-$\Sigma$+H executed \LiveRescoredHistory{} and decided after \LiveRoundHistoryCells{}. At temperature zero the fresh
scores differ from the recorded ones on \LiveMismatchSigmaPercent\% and \LiveMismatchHistoryPercent\% of the executed cells.
In this execution both certificates accept c7, which qualifies on the recorded table and under the fresh scores.

A second execution of the same round, end to end on one H200 GPU per run, shows what a read saves in tokens and time. Counting executed cells that
are never read, the certificates generate \CostTokenShareMin\% to \CostTokenShareMax\% of the output tokens of the complete
table. The batch schedule sets the wall-clock time: with windows that grow from \LiveWindowMin{} tasks \method{}-$\Sigma$ takes about as long
as the complete table on the same GPU, and with a full first batch it finishes in \CostFastestSeconds{}~s, against
\CostCompleteSecondsLow{}~s and \CostCompleteSecondsHigh{}~s for two runs of the complete table.

A certificate that admits a candidate should leave the loop better off on tasks it has not seen. For each chain we score the
initial prompt and the prompt held after round eight on the same \FreshHeldoutTasksPerTarget{} fresh MATH and
\FreshHeldoutTasksPerTarget{} fresh MMLU tasks, drawn outside every partition of the loop. The gain on a target is
$100(W-L)/\FreshHeldoutTasksPerTarget$ points, with $W$ and $L$ the numbers of tasks that only the final or only the initial
prompt solves. The final prompts gain \FreshHeldoutRequired{} points on MATH on average (standard error \FreshGainError{} over
the six chains), and chain \FreshHeldoutLargestChain{} gains the most, \FreshHeldoutLargest{} points, with the candidate c3
accepted in the round of \Cref{fig:overview} as its final prompt. Accuracy rises in all \FreshChainsChangedWord{} chains that
accept an edit; chain 2 accepts none. MMLU changes by +\FreshHeldoutControl{} points on average and by
\FreshHeldoutControlMin{} to +\FreshHeldoutControlMax{} points per chain.

\section{Conclusion}
\label{sec:conclusion}

A round of a self-improvement loop decides a whole batch of candidates against one shared parent on a finite table, and both
of its verdicts have exact conditions in the bounds of the cells the round has not read. One shared random order turns those
bounds into confidence sequences that hold at every stopping time and across the whole batch, and betting on the residuals
of a frozen prediction lowers the cost of later rounds as far as the paid history predicts the current tasks. On public tables the set-level certificates read less than every compared rule that controls its error and decides
every round, and published confidence sequences read less inside the wrapper of \method{} than one candidate at a time; a running loop
lowers its own evaluation cost as its paid history grows, and at an equal budget a loop gated by the certificate runs
three times as many rounds as full evaluation. Since \accept{} names any qualifying candidate, a loop that needs the best
one ranks the candidates by reading on.

\section*{Statements}

\subsection*{Ethics}
The experiments score language models on public benchmarks and on tool-use tasks in simulated environments, with no human
subjects and no personal data. A certificate that admits candidates faster can speed up self-improving systems, and the
control target and the level $\alpha$ give the loop a stated check against regressions.

\subsection*{Reproducibility}
\Cref{sec:experiments} states the rounds, task sets, budgets and comparators. The appendix proves the results of
\Cref{sec:problem,sec:method} and gives the algorithms, the betting parameters, task construction, decoding settings,
sampling rules and cost accounting. The code and the machine-readable records behind every number, table and figure are
released at \url{https://github.com/ZenAlexa/pay-once-certify-every-round}.

\subsection*{Use of generative AI}
Generative AI tools were used in the research and in preparing this paper. The authors verified all content and take full
responsibility for it.

\bibliography{references}

\begin{thebibliography}{93}
\providecommand{\natexlab}[1]{#1}
\providecommand{\url}[1]{\texttt{#1}}
\expandafter\ifx\csname urlstyle\endcsname\relax
  \providecommand{\doi}[1]{doi: #1}\else
  \providecommand{\doi}{doi: \begingroup \urlstyle{rm}\Url}\fi

\bibitem[Agrawal et~al.(2026)Agrawal, Tan, Soylu, Ziems, Khare, Opsahl-Ong,
  Singhvi, Shandilya, Ryan, Jiang, Potts, Sen, Dimakis, Stoica, Klein, Zaharia,
  and Khattab]{agrawal2025gepa}
Lakshya~A. Agrawal, Shangyin Tan, Dilara Soylu, Noah Ziems, Rishi Khare, Krista
  Opsahl-Ong, Arnav Singhvi, Herumb Shandilya, Michael~J. Ryan, Meng Jiang,
  Christopher Potts, Koushik Sen, Alexandros~G. Dimakis, Ion Stoica, Dan Klein,
  Matei Zaharia, and Omar Khattab.
\newblock {{GEPA}: Reflective Prompt Evolution Can Outperform Reinforcement
  Learning}.
\newblock In \emph{International Conference on Learning Representations}, pp.\
  8479--8565, 2026.
\newblock URL
  \url{https://proceedings.iclr.cc/paper_files/paper/2026/hash/0e9e708b6f48e14fd0ac29e167413f76-Abstract-Conference.html}.

\bibitem[Angelopoulos et~al.(2023)Angelopoulos, Bates, Fannjiang, Jordan, and
  Zrnic]{angelopoulos2023ppi}
Anastasios~N. Angelopoulos, Stephen Bates, Clara Fannjiang, Michael~I. Jordan,
  and Tijana Zrnic.
\newblock {Prediction-Powered Inference}.
\newblock \emph{Science}, 382\penalty0 (6671):\penalty0 669--674, 2023.
\newblock \doi{10.1126/science.adi6000}.

\bibitem[Ao et~al.(2026)Ao, Chen, Gao, Li, and Simchi-Levi]{ao2026llmjudgebai}
Ruicheng Ao, Hongyu Chen, Siyang Gao, Hanwei Li, and David Simchi-Levi.
\newblock {Best Arm Identification with LLM Judges and Limited Human Audits}.
\newblock \emph{arXiv preprint arXiv:2601.21471}, 2026.
\newblock URL \url{https://arxiv.org/abs/2601.21471}.

\bibitem[Ashouritaklimi et~al.(2026)Ashouritaklimi, Kilian, Huang, Rainforth,
  and Caron]{ashouritaklimi2026ppactivetesting}
Kianoosh Ashouritaklimi, Valentin Kilian, Daolang Huang, Tom Rainforth, and
  Fran{\c{c}}ois Caron.
\newblock {Prediction-Powered Active Testing}.
\newblock In \emph{Advances in Neural Information Processing Systems}, 2026.
\newblock URL \url{https://openreview.net/forum?id=7YBe8ZysCG}.

\bibitem[Balachandran(2026)]{balachandran2026discern}
Vishnu~Bindu Balachandran.
\newblock {Pay Only for Disagreement: Certified No-Regression Verdicts for
  Model Updates with Matching Label-Complexity Bounds}.
\newblock \emph{arXiv preprint arXiv:2609.17560}, 2026.
\newblock URL \url{https://arxiv.org/abs/2609.17560}.

\bibitem[Berrada et~al.(2025)Berrada, Kossen, Bickford~Smith, Razzak, Gal, and
  Rainforth]{berrada2025activetestingllm}
Gabrielle Berrada, Jannik Kossen, Freddie Bickford~Smith, Muhammed Razzak,
  Yarin Gal, and Thomas Rainforth.
\newblock {Scaling Up Active Testing to Large Language Models}.
\newblock In \emph{Advances in Neural Information Processing Systems},
  volume~38, pp.\  173670--173697, 2025.
\newblock \doi{10.52202/085713-5777}.

\bibitem[Cai et~al.(2026)Cai, Zhang, Nie, Ran, Zheng, Song, Tian, Guo, and
  Xue]{cai2026safeharness}
Qianshu Cai, Yonggang Zhang, Jun Nie, Maohao Ran, Huajiang Zheng, Jun Song,
  Xinmei Tian, Yike Guo, and Wei Xue.
\newblock {A Theory of Reliable Self-Evolution for Agent Harnesses}.
\newblock \emph{arXiv preprint arXiv:2609.08175}, 2026.
\newblock URL \url{https://arxiv.org/abs/2609.08175}.

\bibitem[Cai \& Kandasamy(2026)Cai and Kandasamy]{cai2025constrainedbai}
Ting Cai and Kirthevasan Kandasamy.
\newblock {Constrained Best Arm Identification with Tests for Feasibility}.
\newblock \emph{Proceedings of the AAAI Conference on Artificial Intelligence},
  40\penalty0 (24):\penalty0 19808--19816, 2026.
\newblock \doi{10.1609/aaai.v40i24.39063}.

\bibitem[Charikar et~al.(2000)Charikar, Fagin, Guruswami, Kleinberg, Raghavan,
  and Sahai]{charikar2000priced}
Moses Charikar, Ronald Fagin, Venkatesan Guruswami, Jon Kleinberg, Prabhakar
  Raghavan, and Amit Sahai.
\newblock {Query Strategies for Priced Information (Extended Abstract)}.
\newblock In \emph{ACM Symposium on Theory of Computing}, pp.\  582--591, 2000.
\newblock \doi{10.1145/335305.335382}.

\bibitem[Chen et~al.(2026)Chen, Mahboob, and Weng]{mahboob2026simtoreal}
Yujia Chen, Zaid Mahboob, and Bowen Weng.
\newblock {Betting for Sim-to-Real Performance Evaluation}.
\newblock In \emph{Robotics: Science and Systems}, 2026.
\newblock \doi{10.15607/RSS.2026.XXII.090}.
\newblock URL \url{https://www.roboticsproceedings.org/rss22/p090.html}.

\bibitem[Chi et~al.(2026)Chi, Li, Hong, Wang, Gao, Yang, He, Zheng, Xiao, and
  Na]{chi2026ai4aibench}
Yizhe Chi, Wenyi Li, Deyao Hong, Xiaoqiu Wang, Mingju Gao, Kaisen Yang,
  Bingxiang He, Youjie Zheng, Calvin Xiao, and Qinhuai Na.
\newblock {{AI4AI-Bench}: Benchmarking {LLM} Agents in Algorithmic Design for
  Recursive Self-Improvement}.
\newblock \emph{arXiv preprint arXiv:2608.20318}, 2026.
\newblock URL \url{https://arxiv.org/abs/2608.20318}.

\bibitem[Chowdhury et~al.(2024)Chowdhury, Aung, Shern, Jaffe, Sherburn,
  Starace, Mays, Dias, Aljubeh, Glaese, Jimenez, Yang, Ho, Patwardhan, Liu, and
  Madry]{chowdhury2024swebenchverified}
Neil Chowdhury, James Aung, Chan~Jun Shern, Oliver Jaffe, Dane Sherburn, Giulio
  Starace, Evan Mays, Rachel Dias, Marwan Aljubeh, Mia Glaese, Carlos~E.
  Jimenez, John Yang, Leyton Ho, Tejal Patwardhan, Kevin Liu, and Aleksander
  Madry.
\newblock {Introducing {SWE}-bench Verified}.
\newblock OpenAI, 2024.
\newblock URL \url{https://openai.com/index/introducing-swe-bench-verified/}.

\bibitem[Clerico et~al.(2026)Clerico, Wegel, Azangulov, and
  Rebeschini]{clerico2026timesensitive}
Eugenio Clerico, Tobias Wegel, Iskander Azangulov, and Patrick Rebeschini.
\newblock {Time-Sensitive Anytime-Valid Testing}.
\newblock In \emph{Advances in Neural Information Processing Systems}, 2026.
\newblock URL \url{https://openreview.net/forum?id=ugSaVm063Y}.

\bibitem[Conn(2026)]{conn2026deltaselect}
Nicholas~J. Conn.
\newblock {DeltaSelect}: Affordable {A/B} testing for coding agents.
\newblock \emph{arXiv preprint arXiv:2609.19607}, 2026.

\bibitem[Cowen-Breen et~al.(2026)Cowen-Breen, Agarwal, Bates, Cohen,
  Eisenstein, Globerson, and Fisch]{cowenbreen2026multipleppi}
Charlie Cowen-Breen, Alekh Agarwal, Stephen Bates, William~W. Cohen, Jacob
  Eisenstein, Amir Globerson, and Adam Fisch.
\newblock {Multiple-Prediction-Powered Inference}.
\newblock In \emph{International Conference on Learning Representations}, pp.\
  132989--133033, 2026.
\newblock URL
  \url{https://proceedings.iclr.cc/paper_files/paper/2026/hash/d79ead018586772763ee1e486bd12b3a-Abstract-Conference.html}.

\bibitem[Das et~al.(2026)Das, Patwari, Bukka, Guggilla, Delaye, and
  Sirasao]{das2026mince}
Devleena Das, Rajeev Patwari, Vikram~Kumar Bukka, Nithin~Kumar Guggilla,
  Elliott Delaye, and Ashish Sirasao.
\newblock {MINCE: Shrinking LLM Evaluation Datasets via Few-Model Monte Carlo
  Calibration}.
\newblock In \emph{Proceedings of the 2026 Conference on Empirical Methods in
  Natural Language Processing: Industry Track}, pp.\  1418--1428, 2026.
\newblock \doi{10.18653/2026.emnlp-industry.84}.
\newblock URL \url{https://aclanthology.org/2026.emnlp-industry.84/}.

\bibitem[de~Heide(2026)]{deheide2026unionbound}
Rianne de~Heide.
\newblock {Where Does the Union Bound Go? Best-Arm Identification and Strong
  FWER Control}.
\newblock \emph{arXiv preprint arXiv:2608.19903}, 2026.
\newblock URL \url{https://arxiv.org/abs/2608.19903}.

\bibitem[Deshpande et~al.(2016)Deshpande, Hellerstein, and
  Kletenik]{deshpande2013sbfe}
Amol Deshpande, Lisa Hellerstein, and Devorah Kletenik.
\newblock {Approximation Algorithms for Stochastic Submodular Set Cover with
  Applications to Boolean Function Evaluation and Min-Knapsack}.
\newblock \emph{ACM Transactions on Algorithms}, 12\penalty0 (3):\penalty0
  1--28, 2016.
\newblock \doi{10.1145/2876506}.

\bibitem[Divekar(2026)]{divekar2026ppiranking}
Abhishek Divekar.
\newblock {Statistically Reliable LLM-Based Ranking Evaluation via
  Prediction-Powered Inference}.
\newblock \emph{arXiv preprint arXiv:2606.05308}, 2026.
\newblock URL \url{https://arxiv.org/abs/2606.05308}.

\bibitem[Duan et~al.(2026)Duan, Liu, Tang, Chen, Zhou, Liu, Xu, Wu, Chen, Zhou,
  Wang, Yu, Han, Zhu, Zhou, Lu, Zhou, Liu, Fang, Tian, Chen, Li, Zuo, Zhang,
  Yang, Wang, Qiu, He, Li, Zhou, Liu, Wen, Kang, Zhou, and
  Wu]{duan2026recursive}
Yi~Duan, Ying Liu, Zirui Tang, Haodong Chen, Jun Zhou, Yumou Liu, Bangrui Xu,
  Yukai Wu, Sidi Chen, Yuhan Zhou, Haoyu Wang, Xiaoyou Yu, Shaokun Han, Xuzhou
  Zhu, Le~Zhou, Bolin Lu, Wei Zhou, Jiachen Liu, Nuozhou Fang, Jiaxin Tian,
  Ruoyu Chen, Yuxuan Li, Kai Zuo, Kaiyan Zhang, Qianyu Yang, Zijie Wang,
  Jiantao Qiu, Conghui He, Guoliang Li, Bowen Zhou, Zhiyuan Liu, Zhoufutu Wen,
  Jihua Kang, Xuanhe Zhou, and Fan Wu.
\newblock {The Last AI Built by Humans: Toward Genuine Recursive
  Self-Improvement}.
\newblock \emph{arXiv preprint arXiv:2609.11873}, 2026.
\newblock URL \url{https://arxiv.org/abs/2609.11873}.

\bibitem[Dunnett(1955)]{dunnett1955multiple}
Charles~W. Dunnett.
\newblock {A Multiple Comparison Procedure for Comparing Several Treatments
  with a Control}.
\newblock \emph{Journal of the American Statistical Association}, 50\penalty0
  (272):\penalty0 1096--1121, 1955.
\newblock \doi{10.1080/01621459.1955.10501294}.

\bibitem[Fan \& Huang(2026)Fan and Huang]{helix}
Tianyu Fan and Chao Huang.
\newblock {HELIX: Model-Harness Co-evolution for Recursive Self-Improvement}.
\newblock \emph{arXiv preprint arXiv:2608.13951}, 2026.
\newblock URL \url{https://arxiv.org/abs/2608.13951}.

\bibitem[Feng et~al.(2026)Feng, Shen, Balashankar, Gerner-Beuerle, and
  Rodrigues]{feng2026noisyvalid}
Chen Feng, Minghe Shen, Ananth Balashankar, Carsten Gerner-Beuerle, and Miguel
  R.~D. Rodrigues.
\newblock {Noisy but Valid: Robust Statistical Evaluation of LLMs with
  Imperfect Judges}.
\newblock In \emph{International Conference on Learning Representations}, pp.\
  4643--4691, 2026.
\newblock URL
  \url{https://proceedings.iclr.cc/paper_files/paper/2026/hash/08a67eb74bfb9ded9e949dd52973e997-Abstract-Conference.html}.

\bibitem[Garivier \& Kaufmann(2016)Garivier and
  Kaufmann]{garivier2016trackandstop}
Aur{\'e}lien Garivier and Emilie Kaufmann.
\newblock {Optimal Best Arm Identification with Fixed Confidence}.
\newblock In \emph{Conference on Learning Theory}, volume~49 of
  \emph{Proceedings of Machine Learning Research}, pp.\  998--1027. PMLR, 2016.
\newblock URL \url{https://proceedings.mlr.press/v49/garivier16a.html}.

\bibitem[Ghuge et~al.(2025)Ghuge, Gupta, and Nagarajan]{ghuge2023explainable}
Rohan Ghuge, Anupam Gupta, and Viswanath Nagarajan.
\newblock {Nonadaptive Stochastic Score Classification and Explainable
  Half-Space Evaluation}.
\newblock \emph{Operations Research}, 73\penalty0 (4):\penalty0 2204--2222,
  2025.
\newblock \doi{10.1287/opre.2023.0431}.

\bibitem[Ghuge et~al.(2026)Ghuge, Moondra, and Singh]{ghuge2026correlations}
Rohan Ghuge, Jai Moondra, and Mohit Singh.
\newblock {Stochastic Function Certification with Correlations}.
\newblock \emph{arXiv preprint arXiv:2604.02611}, 2026.
\newblock URL \url{https://arxiv.org/abs/2604.02611}.

\bibitem[{Granite Team, IBM}(2026)]{granite2026model}
{Granite Team, IBM}.
\newblock {Granite-4.2-8B}.
\newblock Model card, 2026.
\newblock URL \url{https://huggingface.co/ibm-granite/granite-4.2-8b}.

\bibitem[Habba et~al.(2026)Habba, Itzhak, Yehudai, Perlitz, Bandel,
  Shmueli-Scheuer, Choshen, and Stanovsky]{habba2026growingpains}
Eliya Habba, Itay Itzhak, Asaf Yehudai, Yotam Perlitz, Elron Bandel, Michal
  Shmueli-Scheuer, Leshem Choshen, and Gabriel Stanovsky.
\newblock {Growing Pains: Extensible and Efficient LLM Benchmarking Via Fixed
  Parameter Calibration}.
\newblock In \emph{Conference on Language Modeling}, 2026.
\newblock URL \url{https://openreview.net/forum?id=cvvnSPpDEd}.

\bibitem[Hendrycks et~al.(2021{\natexlab{a}})Hendrycks, Burns, Basart, Zou,
  Mazeika, Song, and Steinhardt]{hendrycks2021mmlu}
Dan Hendrycks, Collin Burns, Steven Basart, Andy Zou, Mantas Mazeika, Dawn
  Song, and Jacob Steinhardt.
\newblock {Measuring Massive Multitask Language Understanding}.
\newblock In \emph{International Conference on Learning Representations},
  2021{\natexlab{a}}.
\newblock URL \url{https://openreview.net/forum?id=d7KBjmI3GmQ}.

\bibitem[Hendrycks et~al.(2021{\natexlab{b}})Hendrycks, Burns, Kadavath, Arora,
  Basart, Tang, Song, and Steinhardt]{hendrycks2021math}
Dan Hendrycks, Collin Burns, Saurav Kadavath, Akul Arora, Steven Basart, Eric
  Tang, Dawn Song, and Jacob Steinhardt.
\newblock {Measuring Mathematical Problem Solving With the {MATH} Dataset}.
\newblock In \emph{Proceedings of the Neural Information Processing Systems
  Track on Datasets and Benchmarks}, volume~1, 2021{\natexlab{b}}.
\newblock URL
  \url{https://datasets-benchmarks-proceedings.neurips.cc/paper_files/paper/2021/hash/be83ab3ecd0db773eb2dc1b0a17836a1-Abstract-round2.html}.

\bibitem[Hoeffding(1963)]{hoeffding1963probability}
Wassily Hoeffding.
\newblock {Probability Inequalities for Sums of Bounded Random Variables}.
\newblock \emph{Journal of the American Statistical Association}, 58\penalty0
  (301):\penalty0 13--30, 1963.
\newblock \doi{10.1080/01621459.1963.10500830}.

\bibitem[Howard et~al.(2021)Howard, Ramdas, McAuliffe, and
  Sekhon]{howard2021confseq}
Steven~R. Howard, Aaditya Ramdas, Jon McAuliffe, and Jasjeet Sekhon.
\newblock {Time-Uniform, Nonparametric, Nonasymptotic Confidence Sequences}.
\newblock \emph{The Annals of Statistics}, 49\penalty0 (2):\penalty0
  1055--1080, 2021.
\newblock \doi{10.1214/20-AOS1991}.

\bibitem[Hsu \& Shekhar(2026)Hsu and Shekhar]{hsu2026sequentialeval}
Chia-Yu Hsu and Shubhanshu Shekhar.
\newblock {Efficient Sequential Evaluation of Large Language Models}.
\newblock \emph{arXiv preprint arXiv:2607.17409}, 2026.
\newblock URL \url{https://arxiv.org/abs/2607.17409}.

\bibitem[Huang et~al.(2026)Huang, Li, Ling, Lin, Zhong, and
  Wu]{huang2026zipbench}
Zhongzhan Huang, Junxin Li, Guoming Ling, Yupei Lin, Shanshan Zhong, and Hefeng
  Wu.
\newblock {{ZipBench}: Low-Cost Framework for Compressing Comprehensive
  Benchmarks of Large Language Models}.
\newblock In \emph{Proceedings of the 2026 Conference on Empirical Methods in
  Natural Language Processing}, pp.\  42839--42869, 2026.
\newblock \doi{10.18653/2026.emnlp-main.1879}.
\newblock URL \url{https://aclanthology.org/2026.emnlp-main.1879/}.

\bibitem[Iacob et~al.(2026)Iacob, Jovanovi{\'{c}}, Shen, Burkhardt, Kurmanji,
  Tastan, Sani, Venanzi, Nie, Odonnat, Cao, Marino, Qiu, Antonova, and
  Lane]{iacob2026redqueen}
Alex Iacob, Andrej Jovanovi{\'{c}}, William~F. Shen, Daniel Burkhardt, Meghdad
  Kurmanji, Nurbek Tastan, Lorenzo Sani, Niccol{\`{o}} Alberto~Elia Venanzi,
  Jiayi Nie, Ambroise Odonnat, Zeyu Cao, Bill Marino, Xinchi Qiu, Rika
  Antonova, and Nicholas~D. Lane.
\newblock {The Red Queen G{\"o}del Machine: Co-Evolving Agents and Their
  Evaluators}.
\newblock \emph{arXiv preprint arXiv:2606.26294}, 2026.
\newblock URL \url{https://arxiv.org/abs/2606.26294}.

\bibitem[Jiang et~al.(2026)Jiang, Chu, Tian, Zhang, Yang, Yang, Liu, Lv, and
  Li]{jiang2026harnessevolve}
Wen Jiang, Mingmin Chu, Yimeng Tian, Qianxin Zhang, Haofei Yang, Rui Yang, Yang
  Liu, Tao Lv, and Fangming Li.
\newblock {{HarnessEvolve}: Learning from Reference Trajectories for Reliable
  Agent Self-Evolution}.
\newblock \emph{arXiv preprint arXiv:2609.00829}, 2026.
\newblock URL \url{https://arxiv.org/abs/2609.00829}.

\bibitem[Jimenez et~al.(2024)Jimenez, Yang, Wettig, Yao, Pei, Press, and
  Narasimhan]{jimenez2024swebench}
Carlos~E. Jimenez, John Yang, Alexander Wettig, Shunyu Yao, Kexin Pei, Ofir
  Press, and Karthik Narasimhan.
\newblock {{SWE}-bench: Can Language Models Resolve Real-World {GitHub}
  Issues?}
\newblock In \emph{International Conference on Learning Representations}, pp.\
  54107--54157, 2024.
\newblock URL
  \url{https://proceedings.iclr.cc/paper_files/paper/2024/hash/edac78c3e300629acfe6cbe9ca88fb84-Abstract-Conference.html}.

\bibitem[Kalyanakrishnan et~al.(2012)Kalyanakrishnan, Tewari, Auer, and
  Stone]{kalyanakrishnan2012lucb}
Shivaram Kalyanakrishnan, Ambuj Tewari, Peter Auer, and Peter Stone.
\newblock {{PAC} Subset Selection in Stochastic Multi-armed Bandits}.
\newblock In \emph{International Conference on Machine Learning}, pp.\
  655--662, 2012.
\newblock URL \url{https://icml.cc/2012/papers/359.pdf}.

\bibitem[Kano et~al.(2019)Kano, Honda, Sakamaki, Matsuura, Nakamura, and
  Sugiyama]{kano2019goodarm}
Hideaki Kano, Junya Honda, Kentaro Sakamaki, Kentaro Matsuura, Atsuyoshi
  Nakamura, and Masashi Sugiyama.
\newblock {Good Arm Identification via Bandit Feedback}.
\newblock \emph{Machine Learning}, 108\penalty0 (5):\penalty0 721--745, 2019.
\newblock \doi{10.1007/s10994-019-05784-4}.

\bibitem[Karnin et~al.(2013)Karnin, Koren, and
  Somekh]{karnin2013successivehalving}
Zohar Karnin, Tomer Koren, and Oren Somekh.
\newblock {Almost Optimal Exploration in Multi-Armed Bandits}.
\newblock In \emph{International Conference on Machine Learning}, volume~28 of
  \emph{Proceedings of Machine Learning Research}, pp.\  1238--1246. PMLR,
  2013.
\newblock URL \url{https://proceedings.mlr.press/v28/karnin13.html}.

\bibitem[Kilian et~al.(2026)Kilian, Cortinovis, and
  Caron]{kilian2026bayesassistedcs}
Valentin Kilian, Stefano Cortinovis, and Fran{\c{c}}ois Caron.
\newblock {Asymptotically Log-Optimal Bayes-Assisted Confidence Sequences for
  Bounded Means}.
\newblock In \emph{Advances in Neural Information Processing Systems}, 2026.
\newblock URL \url{https://openreview.net/forum?id=I6z9IukSBB}.

\bibitem[Kotawala(2026)]{kotawala2026resolution}
Anany Kotawala.
\newblock {Resolution Diagnostics for Paired LLM Evaluation}.
\newblock \emph{arXiv preprint arXiv:2605.30315}, 2026.
\newblock Accepted to the ICML 2026 Workshop on Hypothesis Testing.

\bibitem[Krumdick et~al.(2026)Krumdick, Wiemerslage, Ebner, Lovering, and
  Tanner]{krumdick2026generalabilities}
Michael Krumdick, Adam Wiemerslage, Seth Ebner, Charles Lovering, and Chris
  Tanner.
\newblock {Cost-Efficient Estimation of General Abilities Across Benchmarks}.
\newblock In \emph{Conference on Language Modeling}, 2026.
\newblock URL \url{https://openreview.net/forum?id=BX1XYGDXEg}.

\bibitem[Li et~al.(2026)Li, Chen, and Huang]{li2026avaivat}
Boning Li, Yu~Chen, and Longbo Huang.
\newblock {AV-AIVAT: 74x Cheaper Agent Evaluation with Certified Anytime-Valid
  Stopping in Imperfect-Information Games}.
\newblock \emph{arXiv preprint arXiv:2608.06362}, 2026.
\newblock URL \url{https://arxiv.org/abs/2608.06362}.

\bibitem[Li(2026)]{li2026whodrifted}
Yitao Li.
\newblock {Who Drifted: the System or the Judge? Anytime-Valid Attribution in
  LLM Evaluation Pipelines}.
\newblock \emph{arXiv preprint arXiv:2606.15474}, 2026.
\newblock URL \url{https://arxiv.org/abs/2606.15474}.

\bibitem[Li \& Cheung(2026)Li and Cheung]{li2026oneidentification}
Zitian Li and Wang~Chi Cheung.
\newblock {Closing the Gap on the Sample Complexity of 1-Identification}.
\newblock In \emph{Advances in Neural Information Processing Systems}, 2026.
\newblock URL \url{https://openreview.net/forum?id=yRrNwwvdTn}.

\bibitem[Liang et~al.(2023{\natexlab{a}})Liang, Bommasani, Lee, Tsipras, Soylu,
  Yasunaga, Zhang, Narayanan, Wu, Kumar, Newman, Yuan, Yan, Zhang, Cosgrove,
  Manning, R{\'e}, Acosta-Navas, Hudson, Zelikman, Durmus, Ladhak, Rong, Ren,
  Yao, Wang, Santhanam, Orr, Zheng, Yuksekgonul, Suzgun, Kim, Guha, Chatterji,
  Khattab, Henderson, Huang, Chi, Xie, Santurkar, Ganguli, Hashimoto, Icard,
  Zhang, Chaudhary, Wang, Li, Mai, Zhang, and Koreeda]{liang2023helm}
Percy Liang, Rishi Bommasani, Tony Lee, Dimitris Tsipras, Dilara Soylu,
  Michihiro Yasunaga, Yian Zhang, Deepak Narayanan, Yuhuai Wu, Ananya Kumar,
  Benjamin Newman, Binhang Yuan, Bobby Yan, Ce~Zhang, Christian Cosgrove,
  Christopher~D. Manning, Christopher R{\'e}, Diana Acosta-Navas, Drew~A.
  Hudson, Eric Zelikman, Esin Durmus, Faisal Ladhak, Frieda Rong, Hongyu Ren,
  Huaxiu Yao, Jue Wang, Keshav Santhanam, Laurel Orr, Lucia Zheng, Mert
  Yuksekgonul, Mirac Suzgun, Nathan Kim, Neel Guha, Niladri Chatterji, Omar
  Khattab, Peter Henderson, Qian Huang, Ryan Chi, Sang~Michael Xie, Shibani
  Santurkar, Surya Ganguli, Tatsunori Hashimoto, Thomas Icard, Tianyi Zhang,
  Vishrav Chaudhary, William Wang, Xuechen Li, Yifan Mai, Yuhui Zhang, and Yuta
  Koreeda.
\newblock {Holistic Evaluation of Language Models}.
\newblock \emph{Transactions on Machine Learning Research}, 2023{\natexlab{a}}.
\newblock URL \url{https://openreview.net/forum?id=iO4LZibEqW}.

\bibitem[Liang et~al.(2023{\natexlab{b}})Liang, Mai, Somerville, Kaiyom, Lee,
  and Bommasani]{liang2023helmlite}
Percy Liang, Yifan Mai, Josselin Somerville, Farzaan Kaiyom, Tony Lee, and
  Rishi Bommasani.
\newblock {HELM Lite}: Lightweight and broad capabilities evaluation.
\newblock Stanford CRFM blog, December 2023{\natexlab{b}}.
\newblock URL \url{https://crfm.stanford.edu/2023/12/19/helm-lite.html}.

\bibitem[Liu et~al.(2026)Liu, Wang, Wang, and Wei]{liu2026pairwisevalidator}
Minghao Liu, Yu~Wang, Jiayun Wang, and Wei Wei.
\newblock {Reward-Free Evolving Agents via Pairwise Validator}.
\newblock \emph{arXiv preprint arXiv:2607.14408}, 2026.
\newblock URL \url{https://arxiv.org/abs/2607.14408}.

\bibitem[Ma \& Wu(2026)Ma and Wu]{ma2026validation}
Qinzhen Ma and Ruihai Wu.
\newblock {When Validation Stops Learning: Auditing Update Admission for
  Continual Embodied Agents}.
\newblock \emph{arXiv preprint arXiv:2609.10873}, 2026.
\newblock URL \url{https://arxiv.org/abs/2609.10873}.

\bibitem[Maia~Polo et~al.(2024)Maia~Polo, Weber, Choshen, Sun, Xu, and
  Yurochkin]{polo2024tinybenchmarks}
Felipe Maia~Polo, Lucas Weber, Leshem Choshen, Yuekai Sun, Gongjun Xu, and
  Mikhail Yurochkin.
\newblock {tinyBenchmarks: evaluating LLMs with fewer examples}.
\newblock In \emph{International Conference on Machine Learning}, volume 235 of
  \emph{Proceedings of Machine Learning Research}, pp.\  34303--34326. PMLR,
  2024.
\newblock URL \url{https://proceedings.mlr.press/v235/maia-polo24a.html}.

\bibitem[Maiorano(2026)]{maiorano2026selftestinggate}
Alexandre~Cristov{\~{a}}o Maiorano.
\newblock {Automated Self-Testing as a Quality Gate: Evidence-Driven Release
  Management for LLM Applications}.
\newblock \emph{arXiv preprint arXiv:2603.15676}, 2026.
\newblock URL \url{https://arxiv.org/abs/2603.15676}.

\bibitem[Mao et~al.(2026)Mao, Qu, Guo, Yuan, Liu, Li, Chen, Fu, Lin, Hu, and
  Liu]{mao2026safeevolve}
Qinghua Mao, Wanying Qu, Dadi Guo, Leitao Yuan, Qingyu Liu, Yu~Li, Guanxu Chen,
  Yanwei Fu, Xi~Lin, Xia Hu, and Dongrui Liu.
\newblock {{SafeEvolve}: Harness-Policy Co-Evolution from Agent Experience for
  Safety Alignment}.
\newblock \emph{arXiv preprint arXiv:2609.02786}, 2026.
\newblock URL \url{https://arxiv.org/abs/2609.02786}.

\bibitem[Martinez-Taboada \& Ramdas(2026)Martinez-Taboada and
  Ramdas]{martineztaboada2026gaussianbetting}
Diego Martinez-Taboada and Aaditya Ramdas.
\newblock {Gaussian-efficient testing by betting on the mean of bounded data}.
\newblock \emph{arXiv preprint arXiv:2608.21694}, 2026.
\newblock URL \url{https://arxiv.org/abs/2608.21694}.

\bibitem[Mukherjee \& Moharir(2026)Mukherjee and Moharir]{mukherjee2026grouped}
Raunak Mukherjee and Sharayu Moharir.
\newblock {Fixed-Budget Constrained Best Arm Identification in Grouped
  Bandits}.
\newblock \emph{arXiv preprint arXiv:2603.04007}, 2026.

\bibitem[Nguyen et~al.(2026)Nguyen, Nguyen, and
  Vuong]{nguyen2026heldoutselection}
Michael Nguyen, Quoc Nguyen, and Paul Vuong.
\newblock {Recursive Self-Evolving Agents via Held-Out Selection}.
\newblock \emph{arXiv preprint arXiv:2606.28374}, 2026.
\newblock URL \url{https://arxiv.org/abs/2606.28374}.

\bibitem[Novikov et~al.(2025)Novikov, V{\~u}, Eisenberger, Dupont, Huang,
  Wagner, Shirobokov, Kozlovskii, Ruiz, Mehrabian, Kumar, See, Chaudhuri,
  Holland, Davies, Nowozin, Kohli, and Balog]{novikov2025alphaevolve}
Alexander Novikov, Ng{\^a}n V{\~u}, Marvin Eisenberger, Emilien Dupont, Po-Sen
  Huang, Adam~Zsolt Wagner, Sergey Shirobokov, Borislav Kozlovskii, Francisco
  J.~R. Ruiz, Abbas Mehrabian, M.~Pawan Kumar, Abigail See, Swarat Chaudhuri,
  George Holland, Alex Davies, Sebastian Nowozin, Pushmeet Kohli, and Matej
  Balog.
\newblock {{AlphaEvolve}: A coding agent for scientific and algorithmic
  discovery}.
\newblock \emph{arXiv preprint arXiv:2506.13131}, 2025.
\newblock URL \url{https://arxiv.org/abs/2506.13131}.

\bibitem[Park et~al.(2025)Park, Zecchin, and Simeone]{park2025autoeval}
Sangwoo Park, Matteo Zecchin, and Osvaldo Simeone.
\newblock {Adaptive Prediction-Powered AutoEval with Reliability and Efficiency
  Guarantees}.
\newblock In \emph{Advances in Neural Information Processing Systems},
  volume~38, pp.\  158758--158784, 2025.
\newblock \doi{10.52202/085713-5303}.
\newblock URL
  \url{https://proceedings.neurips.cc/paper_files/paper/2025/hash/e8e81e35728f7559e42ceb265e104d36-Abstract-Conference.html}.

\bibitem[Pilditch(2026)]{pilditch2026optimalstopping}
Toby~D. Pilditch.
\newblock {Knowing When to Stop: Bayesian Optimal Stopping for LLM
  Evaluations}.
\newblock \emph{arXiv preprint arXiv:2608.14425}, 2026.
\newblock URL \url{https://arxiv.org/abs/2608.14425}.

\bibitem[{Qwen Team}(2026)]{qwen2026qwen35}
{Qwen Team}.
\newblock {{Qwen3.5}: Towards Native Multimodal Agents}, 2026.
\newblock URL \url{https://qwen.ai/blog?id=qwen3.5}.

\bibitem[Ramdas \& Wang(2025)Ramdas and Wang]{ramdas2025evalues}
Aaditya Ramdas and Ruodu Wang.
\newblock {Hypothesis Testing with E-values}.
\newblock \emph{Foundations and Trends in Statistics}, 1\penalty0
  (1--2):\penalty0 1--390, 2025.
\newblock \doi{10.1561/3600000002}.

\bibitem[Ramdas et~al.(2023)Ramdas, Gr{\"{u}}nwald, Vovk, and
  Shafer]{ramdas2022savi}
Aaditya Ramdas, Peter Gr{\"{u}}nwald, Vladimir Vovk, and Glenn Shafer.
\newblock {Game-Theoretic Statistics and Safe Anytime-Valid Inference}.
\newblock \emph{Statistical Science}, 38\penalty0 (4):\penalty0 576--601, 2023.
\newblock \doi{10.1214/23-STS894}.

\bibitem[Razzhigaev et~al.(2026)Razzhigaev, Gritsaev, Kaznacheev, Dragunov,
  Yampolskiy, and Kuznetsov]{ouroboros}
Anton Razzhigaev, Andrei Gritsaev, Andrei Kaznacheev, Nikita Dragunov, Roman
  Yampolskiy, and Andrei Kuznetsov.
\newblock {Ouroboros: A Self-Developing Frontier Coding Agent with Reviewed
  Core Evolution}.
\newblock \emph{arXiv preprint arXiv:2608.08311}, 2026.
\newblock URL \url{https://arxiv.org/abs/2608.08311}.

\bibitem[Robeyns et~al.(2025)Robeyns, Szummer, and Aitchison]{robeyns2025sica}
Maxime Robeyns, Martin Szummer, and Laurence Aitchison.
\newblock {A Self-Improving Coding Agent}.
\newblock In \emph{ICLR 2025 Workshop on Scaling Self-Improving Foundation
  Models without Human Supervision}, 2025.
\newblock URL \url{https://openreview.net/forum?id=rShJCyLsOr}.

\bibitem[Romera-Paredes et~al.(2024)Romera-Paredes, Barekatain, Novikov, Balog,
  Kumar, Dupont, Ruiz, Ellenberg, Wang, Fawzi, Kohli, and
  Fawzi]{romeroparedes2023funsearch}
Bernardino Romera-Paredes, Mohammadamin Barekatain, Alexander Novikov, Matej
  Balog, M.~Pawan Kumar, Emilien Dupont, Francisco J.~R. Ruiz, Jordan~S.
  Ellenberg, Pengming Wang, Omar Fawzi, Pushmeet Kohli, and Alhussein Fawzi.
\newblock {Mathematical Discoveries from Program Search with Large Language
  Models}.
\newblock \emph{Nature}, 625\penalty0 (7995):\penalty0 468--475, 2024.
\newblock \doi{10.1038/s41586-023-06924-6}.

\bibitem[Sandoval et~al.(2026)Sandoval, Waudby-Smith, and
  Jordan]{sandoval2026multiarmedbetting}
Ricardo~J. Sandoval, Ian Waudby-Smith, and Michael~I. Jordan.
\newblock {Multi-Armed Sequential Hypothesis Testing by Betting}.
\newblock \emph{arXiv preprint arXiv:2603.17925}, 2026.
\newblock URL \url{https://arxiv.org/abs/2603.17925}.

\bibitem[Sengupta(2026)]{sea}
Biswa Sengupta.
\newblock {Self-Evolving Agents with Anytime-Valid Certificates}.
\newblock \emph{arXiv preprint arXiv:2607.00871}, 2026.
\newblock URL \url{https://arxiv.org/abs/2607.00871}.

\bibitem[Serfling(1974)]{serfling1974sampling}
R.~J. Serfling.
\newblock {Probability Inequalities for the Sum in Sampling Without
  Replacement}.
\newblock \emph{The Annals of Statistics}, 2\penalty0 (1):\penalty0 39--48,
  1974.
\newblock \doi{10.1214/aos/1176342611}.

\bibitem[Shawn(2026)]{shawn2026pace}
Zayx Shawn.
\newblock {{PACE}: Anytime-Valid Acceptance Tests for Self-Evolving Agents}.
\newblock \emph{arXiv preprint arXiv:2606.08106}, 2026.
\newblock URL \url{https://arxiv.org/abs/2606.08106}.

\bibitem[Shekhar et~al.(2023)Shekhar, Xu, Lipton, Liang, and
  Ramdas]{shekhar2023riskaudits}
Shubhanshu Shekhar, Ziyu Xu, Zachary Lipton, Pierre Liang, and Aaditya Ramdas.
\newblock {Risk-limiting Financial Audits via Weighted Sampling without
  Replacement}.
\newblock In \emph{Conference on Uncertainty in Artificial Intelligence},
  volume 216 of \emph{Proceedings of Machine Learning Research}, pp.\
  1932--1941. PMLR, 2023.
\newblock URL \url{https://proceedings.mlr.press/v216/shekhar23a.html}.

\bibitem[Soni(2026)]{toolfailbench2026}
Harsh Soni.
\newblock {{ToolFailBench}: Diagnosing Tool-Use Failures in {LLM} Agents}.
\newblock In \emph{ICML 2026 Workshop on Agents in the Wild: Safety, Security,
  and Beyond}, 2026.
\newblock URL \url{https://arxiv.org/abs/2607.04686}.

\bibitem[Stark(2023)]{stark2022alpha}
Philip~B. Stark.
\newblock {{ALPHA}: Audit that Learns from Previously Hand-Audited Ballots}.
\newblock \emph{The Annals of Applied Statistics}, 17\penalty0 (1):\penalty0
  641--679, 2023.
\newblock \doi{10.1214/22-AOAS1646}.

\bibitem[Taga et~al.(2026)Taga, Oymak, and Shekhar]{taga2026learningtobet}
Ege~Onur Taga, Samet Oymak, and Shubhanshu Shekhar.
\newblock {Learning to Bet for Horizon-Aware Anytime-Valid Testing}.
\newblock In \emph{International Conference on Machine Learning}, volume 306 of
  \emph{Proceedings of Machine Learning Research}, pp.\  117939--117967. PMLR,
  2026.
\newblock URL \url{https://proceedings.mlr.press/v306/taga26a.html}.

\bibitem[Tenzer et~al.(2026)Tenzer, Tolochinsky, and
  Romano]{tenzer2026semisupervisedbetting}
Yaniv Tenzer, Elad Tolochinsky, and Yaniv Romano.
\newblock {Semi-Supervised Hypothesis Testing by Betting on Predictions}.
\newblock In \emph{International Conference on Machine Learning}, volume 306 of
  \emph{Proceedings of Machine Learning Research}, pp.\  120089--120119. PMLR,
  2026.
\newblock URL \url{https://proceedings.mlr.press/v306/tenzer26a.html}.

\bibitem[Tolochinsky et~al.(2026)Tolochinsky, Tenzer, and
  Romano]{tolochinsky2026bestmodel}
Elad Tolochinsky, Yaniv Tenzer, and Yaniv Romano.
\newblock {Valid Best-Model Identification for LLM Evaluation via Low-Rank
  Factorization}.
\newblock In \emph{Advances in Neural Information Processing Systems}, 2026.
\newblock URL \url{https://openreview.net/forum?id=6uU921FkRV}.

\bibitem[Ville(1939)]{ville1939etude}
Jean Ville.
\newblock {\'E}tude critique de la notion de collectif.
\newblock Th{\`e}ses de l'entre-deux-guerres, no.~218, 1939.
\newblock URL \url{https://www.numdam.org/item/THESE_1939__218__1_0/}.
\newblock Doctoral thesis; zbMATH 0021.14505.

\bibitem[Vovk \& Wang(2021)Vovk and Wang]{vovk2021evalues}
Vladimir Vovk and Ruodu Wang.
\newblock {E-Values: Calibration, Combination and Applications}.
\newblock \emph{The Annals of Statistics}, 49\penalty0 (3):\penalty0
  1736--1754, 2021.
\newblock \doi{10.1214/20-AOS2020}.

\bibitem[Wan et~al.(2026)Wan, Xu, Zorc, Baucells, Hu, Wang, and
  Li]{wan2025beacon}
Guangya Wan, Zixin~Stephen Xu, Sasa Zorc, Manel Baucells, Mengxuan Hu, Hao
  Wang, and Sheng Li.
\newblock {{BEACON}: Bayesian Optimal Stopping for Efficient {LLM} Sampling}.
\newblock In \emph{Findings of the Association for Computational Linguistics:
  EMNLP 2026}, pp.\  2147--2172, 2026.
\newblock \doi{10.18653/2026.findings-emnlp.105}.
\newblock URL \url{https://aclanthology.org/2026.findings-emnlp.105/}.

\bibitem[Wang et~al.(2025)Wang, Pi{\k{e}}kos, Nanbo, Laakom, Chen, Ostaszewski,
  Zhuge, and Schmidhuber]{wang2025hgm}
Wenyi Wang, Piotr Pi{\k{e}}kos, Li~Nanbo, Firas Laakom, Yimeng Chen, Mateusz
  Ostaszewski, Mingchen Zhuge, and J{\"u}rgen Schmidhuber.
\newblock {Huxley-G\"odel Machine}: Human-level coding agent development by an
  approximation of the optimal self-improving machine.
\newblock \emph{arXiv preprint arXiv:2510.21614}, 2025.

\bibitem[Waudby-Smith \& Ramdas(2020)Waudby-Smith and
  Ramdas]{waudbysmith2020wor}
Ian Waudby-Smith and Aaditya Ramdas.
\newblock {Confidence sequences for sampling without replacement}.
\newblock In \emph{Advances in Neural Information Processing Systems},
  volume~33, pp.\  20204--20214, 2020.
\newblock URL
  \url{https://proceedings.neurips.cc/paper/2020/hash/e96c7de8f6390b1e6c71556e4e0a4959-Abstract.html}.

\bibitem[Waudby-Smith \& Ramdas(2024)Waudby-Smith and
  Ramdas]{waudbysmith2023betting}
Ian Waudby-Smith and Aaditya Ramdas.
\newblock {Estimating Means of Bounded Random Variables by Betting}.
\newblock \emph{Journal of the Royal Statistical Society Series B: Statistical
  Methodology}, 86\penalty0 (1):\penalty0 1--27, 2024.
\newblock \doi{10.1093/jrsssb/qkad009}.

\bibitem[Wu et~al.(2026)Wu, Nair, and Cand{\`e}s]{wu2026faq}
Skyler Wu, Yash Nair, and Emmanuel~J. Cand{\`e}s.
\newblock {Efficient Evaluation of {LLM} Performance with Statistical
  Guarantees}.
\newblock In \emph{Advances in Neural Information Processing Systems}, 2026.
\newblock URL \url{https://openreview.net/forum?id=kmsvWSZXrM}.

\bibitem[Wu et~al.(2025)Wu, Yin, Kang, Zhang, Xu, Chen, and Zhang]{wu2025sgm}
Xuening Wu, Shenqin Yin, Yanlan Kang, Xinhang Zhang, Qianya Xu, Zeping Chen,
  and Wenqiang Zhang.
\newblock {{SGM}: A Statistical {G\"odel} Machine for Risk-Controlled Recursive
  Self-Modification}.
\newblock \emph{arXiv preprint arXiv:2510.10232}, 2025.
\newblock URL \url{https://arxiv.org/abs/2510.10232}.

\bibitem[Xu \& Wu(2026)Xu and Wu]{xu2026itemregressions}
Xiaonan Xu and Wenjing Wu.
\newblock {What Aggregate Scores Miss: Measuring Item-Level Regressions in
  Commercial LLM API Migrations}.
\newblock \emph{arXiv preprint arXiv:2608.17719}, 2026.

\bibitem[Yue et~al.(2026)Yue, Cui, Sun, Pan, Xue, Li, Li, Zhu, Chen, Liu,
  Huang, Cui, Zhang, and Zuo]{yue2026ecdysis}
Ruiqing Yue, Yu~Cui, Zhuoyu Sun, Sicheng Pan, Xianhong Xue, Tingyu Li, Ting Li,
  Wenzhuo Zhu, Yi~Chen, Yifei Liu, Baohan Huang, Zhe Cui, Haibin Zhang, and
  Cong Zuo.
\newblock {Ecdysis: Efficient and Effective Training of Runtime Harnesses for
  LLM Agents}.
\newblock \emph{arXiv preprint arXiv:2609.11677}, 2026.
\newblock URL \url{https://arxiv.org/abs/2609.11677}.

\bibitem[Zelikman et~al.(2024)Zelikman, Lorch, Mackey, and
  Kalai]{zelikman2023stop}
Eric Zelikman, Eliana Lorch, Lester Mackey, and Adam~Tauman Kalai.
\newblock {Self-Taught Optimizer ({STOP}): Recursively Self-Improving Code
  Generation}.
\newblock In \emph{Conference on Language Modeling}, 2024.
\newblock URL \url{https://openreview.net/forum?id=46Zgqo4QIU}.

\bibitem[Zhang et~al.(2026{\natexlab{a}})Zhang, Hu, Lu, Lange, and
  Clune]{zhang2026dgm}
Jenny Zhang, Shengran Hu, Cong Lu, Robert Lange, and Jeff Clune.
\newblock {{Darwin G\"odel Machine}: Open-Ended Evolution of Self-Improving
  Agents}.
\newblock In \emph{International Conference on Learning Representations}, pp.\
  104223--104294, 2026{\natexlab{a}}.
\newblock URL
  \url{https://proceedings.iclr.cc/paper_files/paper/2026/hash/aa5f5e6eb6f613ec412f1d948dfa21a5-Abstract-Conference.html}.

\bibitem[Zhang et~al.(2026{\natexlab{b}})Zhang, Wu, and
  Qian]{zhang2026slicewise}
Shengwei Zhang, Tao Wu, and Fei Qian.
\newblock {Certifying Model Upgrades with Slice-Wise Non-Regression and
  Incumbent Fallback}.
\newblock \emph{arXiv preprint arXiv:2609.13714}, 2026{\natexlab{b}}.
\newblock URL \url{https://arxiv.org/abs/2609.13714}.

\bibitem[Zheng et~al.(2026)Zheng, Xue, Liang, Yang, and Zhang]{zheng2026seagym}
Congjie Zheng, Chuanyi Xue, Bin Liang, Jun Yang, and Changshui Zhang.
\newblock {SEAGym: An Evaluation Environment for Self-Evolving LLM Agents}.
\newblock \emph{arXiv preprint arXiv:2606.17546}, 2026.
\newblock URL \url{https://arxiv.org/abs/2606.17546}.

\bibitem[Zhou et~al.(2026{\natexlab{a}})Zhou, Vossler, Sivaraman, Mai, and
  Feng]{zhou2026adaptiveauditing}
Siyu Zhou, Patrick Vossler, Venkatesh Sivaraman, Yifan Mai, and Jean Feng.
\newblock {Adaptive auditing of AI systems with anytime-valid guarantees}.
\newblock In \emph{Advances in Neural Information Processing Systems},
  2026{\natexlab{a}}.
\newblock URL \url{https://openreview.net/forum?id=Bfz7nxHizI}.

\bibitem[Zhou et~al.(2025)Zhou, Song, and Zanette]{zhou2025cveval}
Zhaoyi Zhou, Yuda Song, and Andrea Zanette.
\newblock {Accelerating Unbiased LLM Evaluation via Synthetic Feedback}.
\newblock \emph{arXiv preprint arXiv:2502.10563}, 2025.

\bibitem[Zhou et~al.(2026{\natexlab{b}})Zhou, Ye, Chen, Li, and
  Liu]{zhou2026celeus}
Zhijian Zhou, Zesheng Ye, Zhaorun Chen, Bo~Li, and Feng Liu.
\newblock {CELEUS: Certifiable and Efficient LLM Evaluation via E-Processes}.
\newblock In \emph{Advances in Neural Information Processing Systems},
  2026{\natexlab{b}}.
\newblock URL \url{https://openreview.net/forum?id=z1gbJ1JWes}.

\bibitem[Zrnic \& Cand{\`{e}}s(2024)Zrnic and
  Cand{\`{e}}s]{zrnic2024activeinference}
Tijana Zrnic and Emmanuel~J. Cand{\`{e}}s.
\newblock {Active Statistical Inference}.
\newblock In \emph{International Conference on Machine Learning}, volume 235 of
  \emph{Proceedings of Machine Learning Research}, pp.\  62993--63010. PMLR,
  2024.
\newblock URL \url{https://proceedings.mlr.press/v235/zrnic24a.html}.

\end{thebibliography}
\bibliographystyle{iclr2027_conference}

\appendix
\pretocmd{\section}{\FloatBarrier}{}{}
\makeatletter\setlength{\@fptop}{0pt}\setlength{\@fpsep}{\floatsep}\makeatother
\section{Proofs}
\label{app:proofs}

This appendix proves the results of \Cref{sec:problem,sec:method}, each restated before its proof. A table $x'$ is \emph{consistent} with history $h$ when it agrees with $h$ on
every read cell and keeps every unread score in $[0,1]$.

\subsection{Proof of \texorpdfstring{\Cref{thm:certificates}}{Theorem 1}}
\label{app:proof-certificates}

\restatecertificates*

\begin{proof}
\emph{Soundness.} For any consistent $x'$, each unread candidate score is at least $0$ and each unread parent
score is at most $1$, so $D_{kq}(x')\ge L_{kq}(h)$, and symmetrically $D_{kq}(x')\le U_{kq}(h)$. If $L_{kq}(h)\ge T_q$
for every $q$, then $k\in\mathcal S(x')$ for every consistent $x'$. If every candidate has a target with
$U_{kq}(h)<T_q$, then no candidate belongs to $\mathcal S(x')$ for any consistent $x'$.

\emph{Necessity.} Let $x^{-}$ assign $0$ to every unread candidate score and $1$ to every unread parent score.
It is consistent and attains $D_{kq}(x^-)=L_{kq}(h)$ for every $k,q$. If $k\in\mathcal S(x')$ for all consistent
$x'$, then $L_{kq}(h)=D_{kq}(x^-)\ge T_q$ for every $q$. Let $x^{+}$ assign $1$ to unread candidate scores and $0$
to unread parent scores. It is consistent and attains every upper bound. If $\mathcal S(x')=\varnothing$ for all
consistent $x'$, then $\mathcal S(x^+)=\varnothing$, so every candidate $k$ has a target $q$ with
$U_{kq}(h)=D_{kq}(x^+)<T_q$.

\emph{Adaptivity and completeness.} Both equivalences quantify over the history alone, so they hold at any step of any
score-dependent rule. After a full read $L_{kq}(h)=U_{kq}(h)=D_{kq}(x)$, so some candidate clears every target (\accept{})
or every candidate misses one (\nonebetter{}).
\end{proof}

\subsection{Proof of \texorpdfstring{\Cref{prop:resampling}}{Proposition 1}}
\label{app:proof-resampling}

\restateresampling*

\begin{proof}[Proof of \Cref{prop:resampling}]
Write $I_1,\ldots,I_g$ for the drawn indices, $d_i$ for the paired outcome of task $i$ and $J=\{I_1,\ldots,I_g\}$; the
observations $(I_j,d_{I_j})_{j\le g}$ are a function of the draw sequence and $(d_i)_{i\in J}$. The draw rule is independent of the table by
hypothesis, so conditioning on the draw sequence leaves the table's contribution equal to $(d_i)_{i\in J}$, which
proves the first claim. Since $\lvert J\rvert\le\min\{g,N\}$, at most $\min\{g,N\}$ distinct comparisons occur, each reading the candidate and
the parent cell of one task. A further read of a task already in
$O_{kq}$ or $O_q$ leaves both index sets and both sums of \Cref{eq:endpoints} unchanged, so it moves no bound, and the
distinct cells read number at most $2\lvert J\rvert\le 2\min\{g,N\}$.
\end{proof}

\subsection{Proof of \texorpdfstring{\Cref{thm:sigma}}{Theorem 2}}
\label{app:sigma}

\restatesigma*

\begin{proof}[Proof of \Cref{thm:sigma}]
Fix candidate $k$ and target $q$, and write $N=N_q$, $D=D_{kq}$ and $d_i=y_{kqi}-z_{qi}$. Before the orders are drawn
the round fixes a prediction $p_i$ of each $d_i$, a range $[a_i,c_i]\ni d_i$ for each task, a grid of betting
fractions and nonnegative mixture weights summing to one. \method{}-$\Sigma$ is the case $p_i=0$, $a_i=-1$, $c_i=1$.
Write $P=\sum_ip_i$, which is known. Row $k$ reads a prefix of the uniformly random order $\pi$ of target $q$; let
$\mathcal F_t$ be generated by the side information and the first $t$ entries of $\pi$ with their differences.

\emph{Conditional mean of the residual.} Given $\mathcal F_{t-1}$, the order $\pi$ is uniform over the permutations
consistent with the prefix, so $\pi(t)$ is uniform over the $N-t+1$ unread tasks. Their differences sum to
$D-S_{t-1}$ and their predictions to $P-P_{t-1}$, where $S_{t-1}$ and $P_{t-1}$ are the prefix sums. Under the
hypothesis $D=m$ the residual $e_t=d_{\pi(t)}-p_{\pi(t)}$ therefore has the known conditional mean
$\nu_t(m)=\bigl((m-S_{t-1})-(P-P_{t-1})\bigr)/(N-t+1)$.

\emph{Martingales.} The residual lies in $[a_{\pi(t)}-p_{\pi(t)},\,c_{\pi(t)}-p_{\pi(t)}]$, and the extremes $\underline
e_t$ and $\overline e_t$ of these ranges over the unread tasks are $\mathcal F_{t-1}$-measurable. Each betting factor
$1+\lambda_t(e_t-\nu_t(m))$ with $0\le\lambda_t\le 1/(\nu_t(m)-\underline e_t)$ for $M^{+}$, which defines the lower set, and
$1-\lambda_t(e_t-\nu_t(m))$ with $0\le\lambda_t\le 1/(\overline e_t-\nu_t(m))$ for $M^{-}$, which defines the upper set, is nonnegative
and has conditional mean one at $m=D$, since $\lambda_t$ is predictable; a fraction whose bound is not
positive is set to zero. The product of such factors is a
nonnegative martingale starting at one, and so is the average over the grid with the fixed mixture weights. Every
range contains the true difference, so $\nu_t(D)$ lies inside the range of the unread residuals and the admissible
fractions are well defined. Ville's inequality \citep{ville1939etude,howard2021confseq} gives
$\Pr[\exists t\le N:M^{+}_t(D)\ge2K/\alpha]\le\alpha/(2K)$ and $\Pr[\exists t\le N:M^{-}_t(D)\ge2Q/\alpha]\le\alpha/(2Q)$.
The exact interval $[L_{kq},U_{kq}]$ always contains $D$, so intersecting with it, with fallback to it when the
intersection is empty, keeps $D$ in the set on every path where the betting set keeps it. Lazily read parent values enter the exact interval and leave the martingale untouched. This proves (i) in the row's own filtration.

\emph{Set-level verdicts.} Condition on the complete table and the frozen side information. For each
$k\notin\mathcal S(x)$, define the fixed witness
$q_k=\min\{q:D_{kq}<T_q\}$. Let $E^+_{kq}$ denote the event that the lower set of row $(k,q)$ excludes
$D_{kq}$ at some prefix. Acceptance of $k$ requires every lower set to clear its target, including the set at $q_k$.
Consequently,
\begin{equation}
 \{\text{wrong }\accept{}\}\subseteq\bigcup_{k\notin\mathcal S(x)}E^+_{kq_k},\qquad
 \Pr[\text{wrong }\accept{}]\le |\{k:k\notin\mathcal S(x)\}|\frac{\alpha}{2K}\le\frac\alpha2.
 \label{eq:accept-witness}
\end{equation}
The complete table fixes the witnesses independently of the reading orders, and the evaluator never needs them.

When $\mathcal S(x)$ is nonempty, fix $k^\ast=\min\mathcal S(x)$ and let $E^-_{kq}$ denote the analogous upper-set
exclusion event. Returning \nonebetter{} requires excluding $k^\ast$ through at least one target, giving
\begin{equation}
 \{\text{wrong }\nonebetter{}\}\subseteq\bigcup_{q=1}^Q E^-_{k^\ast q},\qquad
 \Pr[\text{wrong }\nonebetter{}]\le\sum_{q=1}^Q\frac{\alpha}{2Q}=\frac\alpha2.
 \label{eq:none-witness}
\end{equation}
When $\mathcal S(x)$ is empty, the event of a wrong \nonebetter{} is empty. \Cref{lem:path} applies each
row-wise bound at its globally selected prefix. Dependence among targets and shared parent observations leaves
these event inclusions intact, which proves (ii).

\emph{Exact nesting.} Since $\mathcal C^{\pm}_t\subseteq[L_{kq},U_{kq}]$, every exact certificate of
\Cref{thm:certificates} on the cells read is a certificate for the confidence sets, and after a full read with
the parent row complete $[L_{kq},U_{kq}]=\{D\}$, so both sets equal $\{D\}$, which proves (iii).

\emph{Adaptive rounds.} When round $\rho$ fixes its predictions, ranges and weights from the paid history of all
earlier rounds before drawing its orders, the argument above holds conditionally on that history, so round $\rho$ errs
with conditional probability at most $\alpha_\rho$. The union bound over rounds gives total error at most
$\sum_\rho\alpha_\rho\le\alpha$, which proves (iv).
\end{proof}

\subsection{Proof of \texorpdfstring{\Cref{lem:path}}{Lemma 1}}
\label{app:proof-path}

\restatepath*

\begin{proof}
The inclusion $\{\xi\notin\mathcal C_\tau\}\subseteq\{\exists t\le N:\xi\notin\mathcal C_t\}$ holds pathwise, at
$t=\tau$, so $\Pr[\xi\notin\mathcal C_\tau]\le\delta$ for every random index $\tau$ in any filtration.
\end{proof}

\section{Algorithms and implementation}
\label{app:method}

The algorithms and parameters below implement the certificates of \Cref{sec:problem,sec:method}.

\subsection{Algorithms}
\label{app:loop}
\Cref{alg:branchcert} is the exact certificate with acquisition planning, and \Cref{alg:sigma} is \method{}-$\Sigma$ with the
lazy parent row; \method{}-$\Sigma$+H runs \Cref{alg:sigma} with the frozen predictions and ranges of \Cref{eq:prediction}. In
\Cref{alg:sigma} a paid cell is read from the cache at no charge, an unpaid cell is charged at its first read, and the run
returns \undecided{} as soon as a charged read would exceed $B$. No other step uses $B$, so a budgeted run follows the
unbudgeted sequence of cells until its $B$-th charged cell, a stopping time that \Cref{thm:sigma}(ii) covers.

\begin{algorithm}[t]
\caption{\method{}-E for one round}
\label{alg:branchcert}
\footnotesize
\begin{algorithmic}[1]
\Require parent $E_0$, candidates $E_{1:K}$, targets $(N_q,T_q)_{q\le Q}$, budget $B$, repeats $r$, product prior
\State $h\gets\varnothing$,\quad $\beta\gets B$
\Loop
  \If{$\exists k:\ \forall q,\ L_{kq}(h)\ge T_q$} \State \Return \accept{} with candidate $k$ \EndIf
  \If{$\forall k\ \exists q:\ U_{kq}(h)<T_q$} \State \Return \nonebetter{} \EndIf
  \If{$\beta<r$} \State \Return \undecided{} \EndIf
  \State exclude every $k$ with $\Delta_k(h)>\lfloor \beta/r\rfloor$ from acceptance planning
  \State compute $W_{\lfloor \beta/r\rfloor}(h)$ and a maximizing read plan
  \If{the plan contains a cell} \State $c\gets$ its first cell
  \Else \State $c\gets$ the next unscored cell in the fixed row-prefix order
  \EndIf
  \State execute $c$ $r$ times, append its majority score to $h$, and set $\beta\gets \beta-r$
\EndLoop
\end{algorithmic}
\end{algorithm}

\begin{algorithm}[t]
\caption{\method{}-$\Sigma$ for one round}
\label{alg:sigma}
\footnotesize
\begin{algorithmic}[1]
\Require parent $E_0$, candidates $E_{1:K}$, targets $(N_q,T_q)_{q\le Q}$, level $\alpha$, grid $\Theta$, budget $B$
\State draw one uniform ordering $\pi_q$ of the tasks of each target; set every $M^{\pm}\gets1$
\Loop
  \For{each candidate $k$ not excluded}
    \If{$\forall q:\ \min\mathcal C^{+}_{kq}\ge T_q$} \State \Return \accept{} with candidate $k$ \EndIf
    \If{$\exists q:\ \max\mathcal C^{-}_{kq}<T_q$} \State exclude $k$ \EndIf
  \EndFor
  \If{every candidate is excluded} \State \Return \nonebetter{} \EndIf
  \If{the budget is spent} \State \Return \undecided{} \EndIf
  \State $k\gets$ the first candidate not excluded with an uncertified target; $q\gets$ its target with the smallest $(\max\mathcal C^{-}_{kq}-T_q)/N_q$
  \If{$k$ has read at least $N_q/2$ tasks of $q$ and the parent row of $q$ is incomplete}
    \State read the rest of the parent row of $q$ and tighten every $[L_{k'q},U_{k'q}]$; \textbf{continue}
  \EndIf
  \State $i\gets$ the next task of $\pi_q$ in the row of $k$; read $z_{qi}$ if unread, then read $y_{kqi}$
  \State update $M^{\pm}_{kq}(m)$ for every $m$ by \Cref{eq:sigma}, drop every $m$ whose wealth reaches $2K/\alpha$
  from $\mathcal C^{+}_{kq}$ or $2Q/\alpha$ from $\mathcal C^{-}_{kq}$, and intersect both with $[L_{kq},U_{kq}]$
  (unread tasks with a known parent contribute $[-z_{qi},1-z_{qi}]$, the others $[-1,1]$), falling back to
  $[L_{kq},U_{kq}]$ when a set empties
\EndLoop
\end{algorithmic}
\end{algorithm}

\subsection{Betting parameters and implementation}
\label{app:betting-parameters}
The reported certificates use the fixed grid
$\Theta=\{0.05,0.1,0.2,0.3,0.5,0.7,0.9\}$ with equal weights $w_\vartheta=1/7$.
For unread residual extremes $\underline e_t$ and $\overline e_t$, the range-scaled bets are
\begin{equation}
 \lambda^+_{t,\vartheta}(m)=\frac{\vartheta}{\max\{\nu_t(m)-\underline e_t,0.05\}},\qquad
 \lambda^-_{t,\vartheta}(m)=\frac{\vartheta}{\max\{\overline e_t-\nu_t(m),0.05\}}.
 \label{eq:implemented-bets}
\end{equation}
The floor bounds numerical amplification near an endpoint; at the true total it enlarges the denominator and keeps every
betting factor nonnegative. The implementation averages the wealths of the grid components with a log-sum-exp calculation,
the integer hypothesis grid holds every total from $-N_q$ to $N_q$, and the exact interval removes infeasible totals after
each read. The chain predictor uses the pseudo-count $\kappa=2$. The HELM Lite predictor weights the other models by
$\exp(20\,g)$, with $g$ a model's share of agreeing scores with the program on the families outside the round, and fits its
shrinkage factor by least squares, within each task, of the other models' scores on their leave-one-out predictions,
clipped to $[0,1]$. Both residual bets use
the prediction at multiplier one, $e_t=d_{\pi(t)}-p_{\pi(t)}$. \Cref{thm:sigma} covers any prespecified finite grid
with normalized nonnegative weights that satisfy the factor bounds.

\subsection{Acquisition order of \method{}-E}
\label{app:planner}

Call $\Delta_k(h)=\sum_q\max\{0,\lceil T_q-L_{kq}(h)\rceil\}$ the \emph{shortfall} of candidate $k$, the favorable
cells it needs across targets in the most generous case. One cell read supplies at most one favorable unit, so a
candidate whose shortfall exceeds the remaining budget $\beta$ leaves acceptance planning and stays in the set-level
exclusion test (\Cref{prop:obstruction}).

\begin{proposition}[A lower bound on certificate cost]
\label{prop:obstruction}
Let $V_\beta(h)$ be the largest probability, under any law on the unread scores, of reaching \accept{} within $\beta$
further cell reads. If $\Delta_k(h)>\beta$ for every candidate $k$, then $V_\beta(h)=0$.
\end{proposition}

\begin{proof}
Fix candidate $k$. Revealing a candidate score $y_{kqi}\in\{0,1\}$ raises $L_{kq}$ by $y_{kqi}\le1$ and leaves every
other target of $k$ unchanged. Revealing a parent score $z_{qi}\in\{0,1\}$ raises $L_{kq}$ by $1-z_{qi}\le 1$ on
target $q$ alone. So one cell read lowers $\Delta_k(h)$ by at most one, \accept{} for $k$ needs $\Delta_k=0$, and if
$\Delta_k(h)>\beta$ for every $k$ no sequence of $\beta$ reads reaches \accept{}, so $V_\beta(h)=0$.
\end{proof}

Validity comes from \Cref{thm:certificates}, so every acquisition order leaves both certificates sound, and \method{}-E
ranks the next cell by a completion probability under a frozen product prior on the unread scores. Each row of the table, one
candidate or the parent on one target, is read in a fixed order, and the prior gives every unread score a probability of
returning a favorable unit. A \emph{read plan} $\varphi$ for candidate $k$ picks, for each target $q$, how many more candidate and parent cells to read,
at total cost at most the remaining budget $\beta$; $\mathcal P_k(\beta)$ collects these plans. Under the prior the favorable
units $F_q(\varphi)$ a plan collects on target $q$ follow a Poisson-binomial law, and the plan clears $q$ when they reach the
shortfall $\lceil T_q-L_{kq}(h)\rceil$. The cells of different targets are disjoint, so the per-target probabilities multiply,
and maximizing the product over plans and candidates gives the \emph{completion probability}
\begin{equation}
 W_\beta(h)=\max_{k}\ \max_{\varphi\in\mathcal P_k(\beta)}\ \prod_{q}
 \Pr\bigl[F_q(\varphi)\ \ge\ \lceil T_q-L_{kq}(h)\rceil\bigr].
 \label{eq:commitment}
\end{equation}

A budget knapsack over the $Q$ targets computes it in $O(KQ(N_{\max}+B^3))$ arithmetic operations
(\Cref{prop:planner-cost}), where $N_{\max}=\max_q N_q$ is the longest row and $B$ is the budget.
\method{}-E reads the first cell of the maximizing plan, observes its score and recomputes the plan from the new history.

\begin{theorem}[Completion probability under adaptive replanning]
\label{thm:commitment}
For binary scores under a fixed product prior, the replanning schedule reaches \accept{} with probability at least
$W_\beta(h)$ from every undecided history $h$. The one-step lookahead schedule that reads the legal cell $a$
maximizing $\E[W_{\beta-1}(h,a,X_a)]$ reaches \accept{} with probability $\tilde V_\beta(h)$, and
$W_\beta(h)\le \tilde V_\beta(h)\le V^\star_\beta(h)$, where $V^\star_\beta$ is the exact finite-horizon optimum over
all adaptive plans.
\end{theorem}

\begin{proof}
\emph{Lower bound.} If $W_\beta(h)=0$, nonnegativity proves the claim. At an undecided history with $W_\beta(h)>0$,
let $\varphi^\star$ be a read plan attaining $W_\beta(h)$ for candidate $k$, and let $a$ be its first cell.
Conditioning $\varphi^\star$ on the observed value $X_a\in\{0,1\}$ yields a plan $\varphi^\star_{X_a}$ for the same
candidate with the same remaining cells, cost at most $\beta-1$, and shortfalls updated by the observation. This
residual is feasible for the successor planner, so
$W_{\beta-1}(h,a,X_a)\ge \Pr[\varphi^\star\text{ clears every target}\mid X_a]$. Taking expectations over $X_a$ under
the product prior gives $\E[W_{\beta-1}(h,a,X_a)]\ge W_\beta(h)$. Induction on the budget makes the success probability
of the replanning schedule from $h$ at least $W_\beta(h)$. At an accepted history both values equal one. At a
\nonebetter{} history both equal zero. At zero budget, $W_0(h)\in\{0,1\}$ equals the terminal reward.

\emph{Sandwich.} Nonnegativity handles $W_\beta(h)=0$. For $W_\beta(h)>0$, let $a_g$ maximize
$\E[W_{\beta-1}(h,a,X_a)]$ over legal cells. By the argument above,
$\E[W_{\beta-1}(h,a_g,X_a)]\ge\E[W_{\beta-1}(h,a,X_a)]\ge W_\beta(h)$ for the plan head $a$. Induction on $\beta$
gives $\tilde V_\beta(h)=\E[\tilde V_{\beta-1}(h,a_g,X_{a_g})]\ge\E[W_{\beta-1}(h,a_g,X_{a_g})]\ge W_\beta(h)$.
The lookahead schedule is itself a feasible adaptive plan, so $\tilde V_\beta(h)\le V^\star_\beta(h)$.
\end{proof}

\begin{proposition}[Cost of the completion probability]
\label{prop:planner-cost}
The completion probability $W_\beta(h)$ of \Cref{eq:commitment} takes $O(KQ(N_{\max}+B^3))$ arithmetic
operations.
\end{proposition}

\begin{proof}
For a fixed candidate and target, the Poisson-binomial law of favorable units over a prefix of length $\ell$ follows
from the law over length $\ell-1$ by one convolution with a Bernoulli variable. One sweep along the row therefore
produces every prefix law, and the remaining budget truncates each to at most $B$ entries. Scanning a row costs
$O(N_{\max})$ and constructing its budget-truncated prefix laws costs $O(B^2)$. For one candidate and target,
enumerating candidate and parent prefix lengths and evaluating their combined success tail costs $O(B^3)$. The
budget knapsack combines the $Q$ target option tables in $O(QB^2)$ operations. Summing over candidates gives
$O(KQ(N_{\max}+B^3))$ arithmetic operations; the bit complexity of exact rational arithmetic is additional.
\end{proof}

\subsection{Repeated rollouts per cell}
\label{app:rollout-bounds}

\method{} can score a cell by the majority score over $r$ rollouts, with $r$ odd and declared with the round; the repeat
count multiplies the cost of every cell by $r$, and \Cref{thm:certificates} applies unchanged to the table of majority scores.
A stochastic evaluator defines a finite score table once the evaluator version, prompt, task, program, decoding settings and
rollout seeds of every cell are fixed and its score is cached; the reading orders supply the probability in \Cref{thm:sigma}
conditional on this table, and reusing an earlier parent score requires the same table identity. Under a sampling model of the rollouts, independent
rollouts with minority probability $\eta<1/2$, the majority score also recovers the score that the latent success probability
of a cell defines, with the error bound of \Cref{prop:reads}.

\begin{proposition}[Repeated rollouts bound the chance of a wrong verdict]
\label{prop:reads}
Let each rollout of cell $c$ return one with probability $\theta_c$, independently across rollouts, and let the true
score of the cell be $\mathbf 1\{\theta_c>1/2\}$. Let a verdict depend on the true scores of a set $A$ of cells and
write $\gamma=\min_{c\in A}\lvert\theta_c-\tfrac12\rvert$ for their rollout gap. For odd $r$, the $r$-rollout majority
table reproduces every true score in $A$, and therefore the verdict, with probability at least
$1-\lvert A\rvert e^{-2r\gamma^2}$. The single-rollout table reproduces them with probability at least
$1-\sum_{c\in A}\min\{\theta_c,1-\theta_c\}$. In particular, at a common flip probability $\eta<\tfrac12$ per rollout,
one rollout errs on a cell with probability $\eta$, the $r$-rollout majority errs on a cell with probability at most
$e^{-2r(1/2-\eta)^2}$, and the chance of a wrong verdict is at most $\lvert A\rvert\,e^{-2r(1/2-\eta)^2}$.
\end{proposition}

\begin{lemma}[Majority error of one cell]
\label{lem:majority}
Let $X_1,\ldots,X_r$ be independent with $\Pr[X_j=1]=\theta$ and $r$ odd, and let
$M=\mathbf 1\{\frac1r\sum_j X_j>\frac12\}$. Then $\Pr[M\ne\mathbf 1\{\theta>1/2\}]\le e^{-2r(\theta-1/2)^2}$.
\end{lemma}

\begin{proof}
Take $\theta>1/2$ and set $t=\theta-\frac12>0$. The majority errs exactly when $\frac1r\sum_jX_j<\frac12$, which
is $\frac1r\sum_jX_j-\theta<-t$; odd $r$ excludes the boundary. Hoeffding's inequality \citep{hoeffding1963probability} for independent
variables in $[0,1]$ bounds that probability by $e^{-2rt^2}$. For $\theta<1/2$ apply the same argument to $1-X_j$. For
$\theta=1/2$ the bound reads one and holds trivially.
\end{proof}

\begin{proof}[Proof of \Cref{prop:reads}]
Each cell of $A$ satisfies $\lvert\theta_c-\frac12\rvert\ge\gamma$, so \Cref{lem:majority} bounds its
majority error by $e^{-2r\gamma^2}$. Write $v_c=\mathbf 1\{\theta_c>1/2\}$ for the true score of cell $c$ and
$\hat v_c$ for the score a majority of its $r$ rollouts returns. A union bound over $A$ gives
$\Pr[\exists c\in A:\ \hat v_c\ne v_c]\le\lvert A\rvert e^{-2r\gamma^2}$. On the complement every true score in $A$
is reproduced, and the verdict depends on the true scores of $A$, so the verdict is reproduced as well. For a single
rollout, $\Pr[X_1\ne\mathbf 1\{\theta_c>1/2\}]=\min\{\theta_c,1-\theta_c\}$, and a union bound over $A$ gives the
second claim. Solving $\lvert A\rvert e^{-2r\gamma^2}\le\delta$ for $r$ yields $r\ge\gamma^{-2}\ln(\lvert
A\rvert/\delta)/2$.

At a common flip probability $\eta<\frac12$, a rollout returns the true score of its cell with probability $1-\eta$,
so $\theta_c\in\{\eta,1-\eta\}$, $\min\{\theta_c,1-\theta_c\}=\eta$ and $\gamma=\frac12-\eta$, which gives the stated
specialization.
\end{proof}

Under these assumptions $r\ge\gamma^{-2}\ln(\lvert A\rvert/\delta)/2$ rollouts per cell hold the chance of a wrong verdict
below $\delta$ at $r$ times the cost. At $\eta=2/11$, the largest minority share of the audited cells of \Cref{app:rollouts},
a 40-cell table needs \RolloutsUnion{} rollouts per cell by this bound and \RolloutsBinomial{} by the exact binomial
calculation with independent cells to bring the chance of a wrong cell below 0.05.

\subsection{Behavior checks of the paid history}
\label{app:behavior}
We check the conditions of \Cref{sec:history-conditions} on \BehaviourScenarios{} scenarios, and all \BehaviourHolding{} hold.
Flipping a cell that earlier rounds have not read, or a cell of the current or a later round, leaves the predictions and the
run unchanged, and changes the predictions once the mask is dropped; an empty cache reproduces the run without a cache, and a
rerun on its own paid cells charges nothing. A candidate that gains on the required target and fails the control target stays
unaccepted, and an empty qualifying set, two qualifying candidates and a dominant one give the right verdicts. Over every reading order of seven-task tables at the
qualification boundary, the share of wrong verdicts stays at or below $\alpha/2$ for \method{}-$\Sigma$ and
\method{}-$\Sigma$+H, with informed and with misleading predictions. A declared range that misses a paired difference raises an
error, an empty intersection falls back to the exact interval, a full read returns the exact totals, zero betting fractions
reproduce the exact certificate, and on the cells \method{}-$\Sigma$ reads the exact certificate decides no earlier than
\method{}-$\Sigma$.

\section{Extended related work}
\label{app:related}

Self-improving agents choose their next parent by reviewed commits \citep{ouroboros}, by training a runtime harness that
separates model failures from harness failures \citep{yue2026ecdysis}, by selecting on held-out tasks across generations
\citep{nguyen2026heldoutselection}, by the productivity of a program's descendants \citep{wang2025hgm}, or by co-evolving the
evaluator with the agent \citep{iacob2026redqueen}. These approaches develop the proposal process, and the certificate of
\Cref{sec:method} controls the acceptance error and adds the verdict that no candidate of the round qualifies. Evaluation
environments for such loops record training, validation, replay and cost for every harness change \citep{zheng2026seagym},
measure whether an agent can design training algorithms \citep{chi2026ai4aibench}, and gate releases on several quality
dimensions with promote, hold and rollback decisions \citep{maiorano2026selftestinggate}. They define what a round measures,
and the certificate decides when that measurement may stop.

Subset evaluation also estimates scores through fixed item-response parameters \citep{habba2026growingpains} or latent
abilities shared across benchmarks \citep{krumdick2026generalabilities}. The resolution of a paired evaluation depends on its
task set and its dependence structure \citep{kotawala2026resolution}, and item-level studies of model updates find gains and
regressions inside one benchmark \citep{xu2026itemregressions}; a requirement on every target of a round states which
averages the loop protects.

Prediction-powered inference has been extended to combine several prediction models under a cost budget
\citep{cowenbreen2026multipleppi}, to rankings \citep{divekar2026ppiranking} and to automatic evaluation
\citep{park2025autoeval}, active testing chooses which points to label from a
surrogate \citep{berrada2025activetestingllm,ashouritaklimi2026ppactivetesting}, control-variate evaluation corrects synthetic
judge feedback with human labels \citep{zhou2025cveval}, and judge-based certification corrects for judge error rates or
attributes drift between a system and its judge \citep{feng2026noisyvalid,li2026whodrifted}. \method{}-$\Sigma$+H uses the
loop's own paid history as a control variate on a finite table read without replacement, with the prediction frozen before
the reading order is drawn.

Testing by betting \citep{ramdas2025evalues} supplies the confidence sequences used here. Recent work tightens betting
intervals for bounded means \citep{martineztaboada2026gaussianbetting,kilian2026bayesassistedcs}, adapts the bet to a deadline
or a time preference \citep{taga2026learningtobet,clerico2026timesensitive}, tests a global null over several arms
\citep{sandoval2026multiarmedbetting}, and locates the union bound in best-arm identification \citep{deheide2026unionbound}.
Betting also audits AI systems adaptively \citep{zhou2026adaptiveauditing} and transfers simulated robot performance to real
trials \citep{mahboob2026simtoreal}. The certificate allocates its level over a candidate set with two one-sided verdicts on a
finite table.

Best-arm identification has been run with LLM judges and selective human audits \citep{ao2026llmjudgebai}, valid best-model
identification completes a partially observed score matrix by low-rank factorization \citep{tolochinsky2026bestmodel},
Bayesian optimal stopping sets per-item sampling for LLM evaluations \citep{pilditch2026optimalstopping}, and fixed-budget
grouped-bandit identification imposes a threshold on every attribute \citep{mukherjee2026grouped}. A candidate of a round
qualifies through a conjunction of requirements on several targets, its differences share one parent row, and its tasks are
read without replacement.

\section{Experimental setup}
\label{app:setup}

The settings below complete the experiments of \Cref{sec:experiments}.

\subsection{HELM Lite rounds}
\label{app:helm}
HELM Lite publishes per-instance scores for every model it evaluates \citep{liang2023helmlite}. A task scores one when the primary metric of its family equals one: final-number exact match on
GSM8K, chain-of-thought equivalence on MATH, quasi-exact match on MedQA and LegalBench, and exact match on MMLU and OpenBookQA.
The (required, control) pairs are GSM8K--MMLU, MATH--OpenBookQA, MedQA--LegalBench, MMLU--MedQA, LegalBench--GSM8K and
OpenBookQA--MATH. The parent sits at the 20th, 40th, 60th or 80th percentile of required accuracy among the models scored on
both families. The difficulty sweep of \Cref{sec:exp-helm}, drawn in \Cref{fig:structure}, crosses three ways of choosing
the candidates (nearest neighbors, a uniform draw and an ability-stratified draw) with mean requirements of 0.02, 0.05 and
0.10 under 8 reading orders, for 216 rounds; its cell of nearest candidates at $b=0.02$ holds the 24 rounds of
\Cref{sec:exp-helm}.

\subsection{Self-improvement chains}
\label{app:chain}
The chains of \Cref{sec:exp-chain} use MATH tasks of levels 3 to 5 as the required target. Task responses decode at
temperature zero with at most 4{,}096 output tokens in an 8{,}192-token context for chains 1 and 2 and at most 8{,}192 tokens in
a 10{,}240-token context for chains 3 to 6, and the proposer samples at temperature 0.8 with top-$p$ 0.95. Each chain also
holds out 300 MATH tasks that no round reads, the held-out split of \Cref{tab:chain}, and the fresh tasks of
\Cref{sec:exp-live} lie outside every partition of every chain. The live loop executed the complete table of each round and ran
\method{}-$\Sigma$ on it under reading order 0; the first candidate that this run certified became the next parent, and
\nonebetter{} kept the parent. Every replay reads the recorded table of each round from that recorded parent, so all rules run
on the same chains, and a replay under another order that accepts a different qualifying candidate counts as correct and
leaves the later tables unchanged. Its paid history is the cells it read itself along its own orders, each charged at its
first read.

\subsection{Live rounds and tasks}
\label{app:live}
ToolFailBench \citep{toolfailbench2026} holds 1{,}000 single-turn tool-use tasks in five domains (finance, medicine, law,
cybersecurity and real estate). In its 750 tool-required tasks the tool returns a value the model would not guess, and its 250
control tasks attach the same tools and are meant to be answered directly; failures such as skipping the tool, ignoring its result
or fabricating its output are labels the benchmark assigns to responses. These two sets form the required and the control
target. A
hash-ranked stratified split assigns 400 tasks to candidate construction, 300 to admission and 300 to a hidden audit, and every
round draws its 15 required and 5 control tasks from the admission split with its own seed. A task scores one when the
trajectory reaches its declared success condition. The mean requirements are $b_q=1/225$ on the required and $b_q=-1/75$ on the
control target, so the total requirements $T_q=1/15$ and $-1/15$ are fractional, and a round with $K$ candidates holds
$20(K+1)$ cells.

Granite-4.2-8B \citep{granite2026model} proposes and evaluates four rounds, Granite R1 to R4. Each round decodes eight
proposal slots under two prompting conditions at temperature 0.8 and top-$p$ 1.0, after evaluating the parent on 75
construction tasks, and decodes every slot twice with the same prompt and seed, at 8{,}192 and at 16{,}384 output tokens; the
length limit cuts 20 of the 32 responses at 8{,}192 tokens and none at 16{,}384, and one slot yields a qualifying edit at
8{,}192 tokens and nine at 16{,}384. A response enters the candidate set when it
parses to a program that differs from the parent, which gives 9, 7, 12 and 10 candidates, of which 0, 7, 0 and 3 qualify. Granite R3 ran live at a budget of 220 of its 260 cells and Granite R4 at 80 of 220. Qwen3.5-9B
scores the ten candidates of Granite R4 and their parent on the same tasks, the Qwen3.5 R4 table, on which all ten qualify, at
a budget of 80. Granite R6 and R8 have one and two candidates and none qualifying. Rollouts decode at temperature zero with
top-$p$ 1.0 and at most 2{,}048 output tokens, and each cell is executed once ($r=1$). \Cref{tab:live} reports the median over
16 reading orders for \method{}-$\Sigma$ and the other seeded rules, \Cref{tab:budgets} the median over 32 for the seeded rules
of its budget replay, and \method{}-E and the fixed orders are deterministic.

\subsection{Compared rules}
\label{app:baselines}
Every compared rule runs on the same tables, reading orders, level and distinct-cell accounting as \method{}-$\Sigma$. A rule
that judges one candidate enters the Bonferroni wrapper of \Cref{sec:experiments}, which gives it level $\alpha/(KQ)$ on each
candidate--target pair, audits the candidates in order, accepts the first one certified on every target and returns
\nonebetter{} when every candidate is excluded.

The without-replacement betting sequences run in two wrappers: the Bonferroni wrapper with an eager parent, which reads the parent row first, and two-sided levels
$\alpha/(KQ)$, and the wrapper of \method{} of \Cref{sec:history}, with the shared orders, the lazy parent row, the parent
cache, first-read charging and one-sided levels $\alpha/(2K)$ and $\alpha/(2Q)$. Both use the certificate driver of
\method{}-$\Sigma$, with the grid of hypothesized totals and the exact-interval fallback, and map paired differences to
$[0,1]$ by $d\mapsto(d+1)/2$. Hedged-capital betting \citep{waudbysmith2023betting} uses predictable plug-in bets truncated at $1/2$, and
the empirical-Bernstein sequence \citep{waudbysmith2020wor} unions its two one-sided bounds. ALPHA \citep{stark2022alpha} runs
the shrink-trunc test with shrinkage parameter 100 and its prior mean set from the frozen history prediction or by the
adaptive rule of the paper. Control-variate betting \citep{shekhar2023riskaudits} uses ApproxKelly bets with uniform weights
and the frozen history prediction as control variate. \method{}-$\Sigma$ and \method{}-$\Sigma$+H also run inside the
Bonferroni wrapper, so the spread within that block measures the effect of the sequence and the gap to the wrapper of
\method{} the effect of the set-level wrapper.

Among the sequential and active evaluators, CELEUS \citep{zhou2026celeus}, the active confidence sequences of \citet{hsu2026sequentialeval}, FAQ \citep{wu2026faq},
AV-AIVAT \citep{li2026avaivat} and Best-model ID \citep{tolochinsky2026bestmodel} run one process per candidate and target at
level $\alpha/(KQ)$. Each receives a logistic low-rank factor model of the other models' scores on the round's tasks, with its
rank chosen from $\{8,16,32,64\}$ and its weight decay from $10^{1}$ to $10^{-5}$ on a held-out fifth of the entries and
trained with Adam for 2{,}000 epochs at learning rate $10^{-2}$; each program of the round starts from a Gaussian prior on its
factor fitted to the other models' factors and updates it by a rank-one Laplace step after every read: CELEUS uses it as the surrogate of its hedged e-process, the active sequences bet with its predicted
distribution, FAQ samples tasks by its predicted spread at a planned budget of a quarter of the table, AV-AIVAT subtracts it as
control variate, and Best-model ID combines bandit sampling with doubly robust estimates on it.

Of the per-candidate gates, DISCERN \citep{balachandran2026discern} and slice-wise certification \citep{zhang2026slicewise} release no code and are
reproduced from their text. DISCERN certifies one candidate against a retained parent through the disagreement identity
$\Delta=\E[D\,A]$ with a predictable plug-in empirical-Bernstein confidence sequence; a disagreement read reads both cells,
each target draws tasks with replacement, verdicts are shifted to our integer requirements, the level is split as
$\alpha/(KQ)$, and a stream stops at the 60{,}000 draws of the paper's longest horizon. Slice-wise
certification runs a one-sided Hoeffding non-inferiority test per target with an intersection-union rule and a Bonferroni split
over a candidate library frozen on a tenth of each target's tasks (the share most favorable to it on these tables), sized by
its planning rule at power $0.8$; it falls back to the parent when certification fails, which we record as \undecided{}, since
it certifies no exclusion of every candidate. The PACE-style gate$^{\dagger}$ \citep{shawn2026pace} runs the paired betting
test of \method{}-$\Sigma$ with the with-replacement mean and no exact interval. The Two-Gate rule of \citet{cai2026safeharness}
runs per candidate with an even risk split and rejects every candidate of the live rounds, its gain bound being non-positive at
15 required tasks.

The fixed-batch paired test of \citet{ma2026validation}, a paired-binomial test, draws 256 required and 512 control pairs uniformly with replacement,
executes the candidate and the parent with a shared reset for every pair, and decides from the positive and negative
disagreement counts at spending level $\delta=0.1$, with $\alpha=0.0608$ at the first attempt and $0.0152$ at the second. Its
candidate ranks first on the complete table, a choice that favors the test. Its 1{,}536 rollouts are charged as 40 distinct
cells covering 20 comparisons (\Cref{prop:resampling}), and a deferral is \undecided{}.

Among the bandit, subset, racing and stopping rules, good-arm identification with the HDoC index \citep{kano2019goodarm} and 1-identification \citep{li2026oneidentification}
certify whether some candidate clears a threshold, one pull reading one task of a candidate and its parent; with LUCB bounds,
good-arm identification reads \BanditGoodArmLucbShare\% and decides \BanditGoodArmLucbDecides\% of runs.
tinyBenchmarks \citep{polo2024tinybenchmarks}, calibrated on many earlier models, and MINCE \citep{das2026mince}, calibrated on
a few, estimate each candidate's score from a task subset and accept the best estimate above the requirement, with no error
control; MINCE on a fixed subset reads \BanditMinceShare\% with \BanditMinceWrong\% wrong verdicts, and \Cref{tab:helm} reports MINCE
sized to the requirement. Successive halving \citep{karnin2013successivehalving} names its surviving leader,
and the LUCB race \citep{kalyanakrishnan2012lucb,dunnett1955multiple} races the candidates against the shared parent with time
and candidate error spending. BEACON-style stopping$^{\dagger}$ borrows the Bayesian stopping idea of BEACON, a stop rule for best-of-$N$ sampling
\citep{wan2025beacon}: it accepts when a Dirichlet posterior over the paired differences gives qualifying a probability of
$1-\alpha$, with no frequentist error control.

Among the minibatch gates of practical loops, GEPA's gate \citep{agrawal2025gepa} accepts the first candidate whose sum on a fresh required minibatch strictly exceeds the
parent's with the control minibatch not below the parent's, with minibatches of 3 tasks (the default of the \texttt{gepa}
package), 10 and 30. GEPA's pipeline passes the candidates through a required minibatch of 3, scores every passed candidate and
the parent on every task, and accepts the qualifier with the largest required gain. The random subset of matched size reads one
shared random subset per target for every program, sized to the cells \method{}-$\Sigma$+H spent on the same round and reading order, and
accepts the largest estimated required gain among the estimates that meet both requirements. All three run on the 48 recorded
chain rounds under the 8 reading orders of \Cref{tab:chain}.

\subsection{Loops at an equal budget}
\label{app:gate-loops}
For each of three seeds, four loops start from the initial prompt and the task split of the recorded chain with that seed and
propose eight candidate system prompts per round with the proposer of the chains. Only the gate differs, among
\method{}-$\Sigma$+H with the cells its own loop paid as history, full evaluation on the complete table, the GEPA pipeline and GEPA's gate with
three tasks. Every loop has a budget of 108{,}000 acceptance cells, four complete tables, counting every executed cell
including unread prefetch, and at most 12 rounds. Qwen3.5-9B \citep{qwen2026qwen35} proposes at temperature 0.8 with
top-$p$ 0.95, as in the chains, and answers at temperature zero with at most 8{,}192 output tokens, served by vLLM on one H100
per loop. The proposer's development set and each new parent's accuracy
on 1{,}000 MATH and 1{,}000 MMLU test tasks, disjoint from the acceptance tasks, run outside the budget, and after the last round
every candidate accepted by a gate that did not read its complete table is audited on that table. One rule compares the final parents. \method{}-$\Sigma$+H is ahead of a gate when its mean required test accuracy over the
seeds exceeds the gate's by more than twice the standard error of the per-seed differences and its mean control accuracy is at
most 2 points lower, tied when the difference lies within twice that standard error, and behind otherwise.

\subsection{The SWE-bench Verified stream}
\label{app:swe}
The per-issue resolution of \SweSystems{} SWE-bench Verified submissions \citep{jimenez2024swebench,chowdhury2024swebenchverified}
from the public experiments repository of the benchmark, those with per-instance results, ordered by date, form the stream.
The \SweRequiredTasks{} issues annotated as taking at least fifteen minutes to fix form the required target ($b=0.02$) and the
\SweControlTasks{} shorter issues the control target ($b=-0.02$). The oldest submission is the initial parent, and the others
arrive eight per round over \SweRounds{} rounds; the loop generates no proposals. Each of six gates runs its own loop, whose
accepted candidate becomes its next parent: \method{}-$\Sigma$, \method{}-$\Sigma$+H with the cells its own loop paid, the
exact certificate, full evaluation, the GEPA pipeline and GEPA's gate with three tasks. Every loop spends
$\alpha_\rho=0.05/(\zeta(2)\rho^2)$ in round $\rho$, charges every agent--issue result once per loop and is checked against
the complete table of each round it decides, under \SweOrders{} reading orders per gate.

\section{Additional results}
\label{app:results}

These results extend the comparisons of \Cref{sec:experiments}.

\subsection{Every rule on HELM Lite}
\label{app:helm-results}
\begin{figure}[!htb]
\centering
\includegraphics[width=\textwidth]{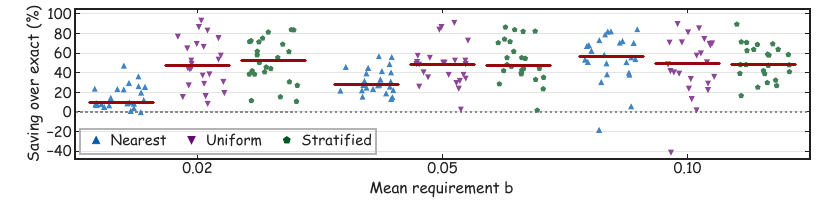}
\caption{Saving of \method{}-$\Sigma$ over the exact certificate on each of the 216 HELM Lite rounds of the difficulty
sweep, by the way the ten candidates are chosen and the mean requirement $b$; bars are medians. The families and their task
counts stay fixed.}
\label{fig:structure}
\end{figure}

\Cref{tab:helm} lists every rule of \Cref{fig:helm} with its citation, the published sequences in both wrappers, and the
share each rule replayed on the chains reads there; \Cref{tab:levels} gives the level sweep of \Cref{sec:exp-helm}, and
\Cref{fig:structure} shows its difficulty sweep. The exact certificate reads more than \method{}-$\Sigma$ in
\HelmSigmaBelowExact{} of the \HelmRoundCount{} rounds. Inside the wrapper of \method{}, \method{}-$\Sigma$ reads fewer median
cells than hedged capital in \WrappedHedgedRoundsLower{} of the 24 rounds and fewer than adaptive ALPHA in
\WrappedAlphaRoundsLower{}.

\Cref{tab:helm} charges each rule its own stopping cost. A replay at one budget for every rule caps each run at a fixed
share of its table, truncating the anytime rules and giving tinyBenchmarks and MINCE the same share of the tasks of every
target. At \LadderBudget\% of each table \method{}-$\Sigma$ returns the complete-table verdict in \LadderSigmaCorrect\% of the
runs, and no bandit or subset rule in more than \LadderOthersCorrect\%.
Good-arm identification and 1-identification decide no run below \LadderBanditFirstDecision\% of a table and return the
complete-table verdict in at most \LadderBanditFullCorrect\% of the runs at the full table; tinyBenchmarks and MINCE err in \LadderHighWrongTiny\% and
\LadderHighWrongMince\% of the runs at \LadderHighBudget\%, and reach the complete-table verdict only at the full read.

\begin{table}[!htb]
\caption{Every rule on the 24 HELM Lite rounds at $\alpha=0.05$, 16 reading orders each. Read is the share of the table
read, its median over the orders of a round averaged over rounds; Decides is the share of runs that return \accept{} or
\nonebetter{}; Wrong is the share of decided runs that disagree with the complete table. Among the rules that decide every
run with at most $\alpha/2$ wrong verdicts, bold marks the lowest share and underline the second lowest. Averaged over all runs,
\method{}-$\Sigma$ reads \HelmRunMeanSigma\% (standard error \HelmOrderErrorSigma{} points over the 16 orders), and within a
round its share varies across orders with a standard deviation of \HelmOrderSpreadSigma{} points. Chains is the mean share read
over the 384 runs of the six recorded chains, for the rules replayed there, none of which errs.}
\label{tab:helm}
\centering
\small
\setlength{\tabcolsep}{4pt}
\begin{threeparttable}
\begin{tabular}{l r r r r}
\toprule
& \multicolumn{3}{c}{HELM Lite (\%)} & \\
\cmidrule(lr){2-4}
Rule & Read\,$\downarrow$ & Decides\,$\uparrow$ & Wrong\,$\downarrow$ & Chains (\%)\,$\downarrow$ \\
\midrule
\multicolumn{5}{@{}l}{\textit{\method{}}} \\
\method{}-$\Sigma$+H$^{\ast}$ & \textbf{37.8} & 100 & 0.3 & \textbf{19.4} \\
\method{}-$\Sigma$ & {\secondbest{38.2}} & 100 & 0.5 & {\secondbest{23.2}} \\
\midrule
\multicolumn{5}{@{}l}{\textit{Exact certificate}} \\
Exact, parent first & 44.7 & 100 & 0.0 & \textemdash \\
\midrule
\multicolumn{5}{@{}l}{\textit{Without-replacement betting sequences inside the \method{} wrapper}} \\
WoR hedged capital~\citep{waudbysmith2023betting} & 40.5 & 100 & 0.3 & 25.2 \\
WoR empirical Bernstein~\citep{waudbysmith2020wor} & 41.2 & 100 & 0.0 & 27.5 \\
ALPHA, history prior$^{\ast}$~\citep{stark2022alpha} & 41.4 & 100 & 0.0 & 32.6 \\
ALPHA, adaptive prior~\citep{stark2022alpha} & 39.4 & 100 & 0.0 & 27.4 \\
Control-variate betting$^{\ast}$~\citep{shekhar2023riskaudits} & 40.0 & 100 & 0.0 & 23.7 \\
\midrule
\multicolumn{5}{@{}l}{\textit{Without-replacement betting sequences, Bonferroni wrapper with eager parent}} \\
WoR hedged capital~\citep{waudbysmith2023betting} & 44.4 & 100 & 0.0 & 37.7 \\
WoR empirical Bernstein~\citep{waudbysmith2020wor} & 44.7 & 100 & 0.0 & 40.6 \\
ALPHA, history prior$^{\ast}$~\citep{stark2022alpha} & 44.4 & 100 & 0.0 & 46.0 \\
ALPHA, adaptive prior~\citep{stark2022alpha} & 43.6 & 100 & 0.0 & 41.7 \\
Control-variate betting$^{\ast}$~\citep{shekhar2023riskaudits} & 44.1 & 100 & 0.0 & 36.1 \\
\method{}-$\Sigma$, Bonferroni wrapper & 43.1 & 100 & 0.0 & 34.8 \\
\method{}-$\Sigma$+H, Bonferroni wrapper$^{\ast}$ & 42.9 & 100 & 0.0 & 30.3 \\
\midrule
\multicolumn{5}{@{}l}{\textit{Sequential and active evaluators}} \\
CELEUS$^{\ast}$~\citep{zhou2026celeus} & 61.9 & 100 & 0.0 & \textemdash \\
Active CS$^{\ast}$~\citep{hsu2026sequentialeval} & 46.8 & 9 & 0.0 & \textemdash \\
AV-AIVAT, asymptotic$^{\ast}$~\citep{li2026avaivat} & 60.3 & 57 & 0.0 & \textemdash \\
Best-model ID$^{\ast}$~\citep{tolochinsky2026bestmodel} & 56.5 & 58 & 0.0 & \textemdash \\
FAQ$^{\ast}$~\citep{wu2026faq} & 30.6 & 5 & 0.0 & \textemdash \\
AV-AIVAT, empirical Bernstein$^{\ast}$~\citep{li2026avaivat} & 71.5 & 51 & 0.0 & \textemdash \\
\midrule
\multicolumn{5}{@{}l}{\textit{Per-candidate gates}} \\
DISCERN~\citep{balachandran2026discern} & 69.3 & 40 & 0.0 & \textemdash \\
Slice-wise~\citep{zhang2026slicewise} & 28.8 & 22 & 0.0 & \textemdash \\
PACE-style gate$^{\dagger}$~\citep{shawn2026pace} & 77.7 & 4 & 0.0 & \textemdash \\
\midrule
\multicolumn{5}{@{}l}{\textit{Bandit, subset and racing rules}} \\
HDoC good-arm~\citep{kano2019goodarm} & 93.8 & 54 & 0.0 & \textemdash \\
1-identification~\citep{li2026oneidentification} & 100.0 & 16 & 0.0 & \textemdash \\
tinyBenchmarks$^{\ast}$~\citep{polo2024tinybenchmarks} & 13.6 & 100 & 65.1 & \textemdash \\
MINCE, sized$^{\ast}$~\citep{das2026mince} & 67.0 & 100 & 19.8 & \textemdash \\
Successive halving~\citep{karnin2013successivehalving} & 32.6 & 100 & 75.5 & \textemdash \\
LUCB race~\citep{kalyanakrishnan2012lucb} & 44.2 & 0 & \textemdash & \textemdash \\
\midrule
\multicolumn{5}{@{}l}{\textit{No error control}} \\
BEACON-style$^{\dagger}$~\citep{wan2025beacon} & 16.8 & 63 & 39.3 & \textemdash \\
\bottomrule
\end{tabular}
\begin{tablenotes}\footnotesize
\item[] $^{\ast}$Uses the scores of other models on HELM Lite and the paid history on the chains. $^{\dagger}$Built by the authors from the idea of the cited work. A dash marks a rule that never decides or was not replayed on the chains.
\end{tablenotes}
\end{threeparttable}
\end{table}

\begin{table}[!htb]
\caption{The level sweep on the 24 HELM Lite rounds, 16 reading orders each: share read, runs that decide and wrong verdicts
among the decided runs (\%) at five levels $\alpha$. A dash marks a level at which a rule decides no run.}
\label{tab:levels}
\centering
\small
\setlength{\tabcolsep}{3pt}
\begin{tabular}{@{}l rrrrr rrrrr rrrrr@{}}
\toprule
& \multicolumn{5}{c}{Read (\%) at $\alpha=$} & \multicolumn{5}{c}{Decides (\%) at $\alpha=$} & \multicolumn{5}{c}{Wrong (\%) at $\alpha=$} \\
\cmidrule(lr){2-6}\cmidrule(lr){7-11}\cmidrule(l){12-16}
Rule & 0.01 & 0.02 & 0.05 & 0.1 & 0.2 & 0.01 & 0.02 & 0.05 & 0.1 & 0.2 & 0.01 & 0.02 & 0.05 & 0.1 & 0.2 \\
\midrule
\method{}-$\Sigma$ & 39.8 & 39.2 & 38.2 & 37.4 & 35.9 & 100 & 100 & 100 & 100 & 100 & 0.0 & 0.0 & 0.5 & 0.5 & 1.0 \\
\method{}-$\Sigma$+H & 39.5 & 39.0 & 37.8 & 37.0 & 35.9 & 100 & 100 & 100 & 100 & 100 & 0.0 & 0.0 & 0.3 & 0.5 & 1.6 \\
Exact, parent first & 44.7 & 44.7 & 44.7 & 44.7 & 44.7 & 100 & 100 & 100 & 100 & 100 & 0.0 & 0.0 & 0.0 & 0.0 & 0.0 \\
Active CS & 49.0 & 48.1 & 46.7 & 45.6 & 44.1 & 7 & 8 & 9 & 10 & 11 & 0.0 & 0.0 & 0.0 & 0.0 & 0.0 \\
CELEUS & 64.6 & 63.7 & 61.9 & 60.4 & 58.7 & 100 & 100 & 100 & 100 & 100 & 0.0 & 0.0 & 0.0 & 0.0 & 0.0 \\
FAQ & 31.0 & 30.9 & 30.7 & 30.4 & 29.8 & 4 & 4 & 6 & 7 & 8 & 0.0 & 0.0 & 0.0 & 0.0 & 0.0 \\
DISCERN & 72.1 & 70.8 & 69.3 & 68.0 & 66.3 & 38 & 39 & 40 & 42 & 45 & 0.0 & 0.0 & 0.0 & 0.0 & 0.0 \\
PACE-style gate$^{\dagger}$ & 80.4 & 79.4 & 77.7 & 76.2 & 74.3 & 0 & 0 & 4 & 4 & 4 & \textemdash & \textemdash & 0.0 & 0.0 & 0.0 \\
Slice-wise & 28.9 & 28.8 & 28.8 & 28.7 & 28.6 & 22 & 22 & 22 & 22 & 22 & 0.0 & 0.0 & 0.0 & 0.0 & 0.0 \\
LUCB race & 44.2 & 44.2 & 44.2 & 44.2 & 44.2 & 0 & 0 & 0 & 0 & 0 & \textemdash & \textemdash & \textemdash & \textemdash & \textemdash \\
Successive halving & 32.6 & 32.6 & 32.6 & 32.6 & 32.6 & 100 & 100 & 100 & 100 & 100 & 75.5 & 75.5 & 75.5 & 75.5 & 75.5 \\
BEACON-style$^{\dagger}$ & 47.7 & 37.6 & 16.8 & 5.3 & 1.2 & 26 & 35 & 63 & 88 & 100 & 32.7 & 34.6 & 39.3 & 45.4 & 52.5 \\
\bottomrule
\end{tabular}
\end{table}

\Cref{tab:batch_width} rebuilds the rounds of \Cref{tab:helm} with 5, 10, 20 and 40 nearest-ability candidates, and with a
second control target on a third family at $b=-0.02$, the first family after the control family in alphabetical order outside
the round's pair. \method{}-$\Sigma$ reads the least in every setting, and its wrong verdicts are at most \WidthWrongMax\% of
the \WidthRuns{} runs of a setting. The share read falls with the batch width mostly through the mix of verdicts: a wider
batch of nearest candidates holds a qualifying candidate in more rounds (\WidthFortyTwoQualifying{} of \WidthRounds{} with 40
candidates and two targets, against \WidthTenTwoQualifying{} with ten), so more rounds end at their first \accept{}.

\begin{table}[!htb]
\caption{Batch width and target count on HELM Lite (24 rounds per setting, 16 reading orders, $\alpha=0.05$): $K$ nearest-ability candidates and $Q$ targets, where $Q=3$ adds a second control target. Qualifying counts the rounds whose complete table holds a qualifying candidate. Shares are the within-round median share read (\%), averaged over rounds, for \method{}-$\Sigma$, the parent-first exact certificate, and hedged capital and adaptive ALPHA in the Bonferroni wrapper with an eager parent; each row's lowest is bold and second underlined. Saving is the mean over rounds of the reduction in median cells of \method{}-$\Sigma$ against each rule (\%), and Cells the mean cells \method{}-$\Sigma$ reads per round.}
\label{tab:batch_width}
\centering
\footnotesize
\setlength{\tabcolsep}{3.2pt}
\begin{tabular}{@{}r r r r r r r r r r r@{}}
\toprule
& & & \multicolumn{4}{c}{Share read (\%)} & \multicolumn{3}{c}{Saving (\%)} & \\
\cmidrule(lr){4-7}\cmidrule(lr){8-10}
$K$ & $Q$ & Qualifying & $\Sigma$ & Exact & Hedged & ALPHA & vs.\ exact & vs.\ hedged & vs.\ ALPHA & Cells \\
\midrule
5 & 2 & 4 & \textbf{41.8} & 51.7 & 51.7 & {\secondbest{51.1}} & 19.5 & 19.7 & 18.8 & 4{,}486 \\
10 & 2 & 9 & \textbf{38.2} & 44.7 & 44.4 & {\secondbest{43.6}} & 14.9 & 14.2 & 12.7 & 7{,}566 \\
20 & 2 & 20 & \textbf{18.8} & 22.3 & 21.2 & {\secondbest{20.7}} & 18.4 & 13.4 & 11.4 & 7{,}274 \\
40 & 2 & 24 & \textbf{5.8} & 8.3 & 7.2 & {\secondbest{7.1}} & 30.9 & 21.7 & 20.4 & 4{,}226 \\
10 & 3 & 6 & \textbf{28.3} & 37.0 & 35.9 & {\secondbest{35.3}} & 24.3 & 22.6 & 21.7 & 7{,}845 \\
20 & 3 & 17 & \textbf{16.0} & 22.7 & 20.2 & {\secondbest{19.8}} & 26.6 & 20.2 & 18.8 & 8{,}473 \\
40 & 3 & 23 & \textbf{5.3} & 9.5 & 7.1 & {\secondbest{7.0}} & 38.1 & 25.8 & 24.9 & 5{,}455 \\
\bottomrule
\end{tabular}
\end{table}

\subsection{Betting grids}
\label{app:grid}
\Cref{tab:betting-grid} reports all \BettingGridTotalRuns{} runs of four grids with equal weights and the history multiplier
at one.

\begin{table}[!htb]
\caption{Four fixed grids of betting fractions on the 24 HELM Lite rounds at $\alpha=0.05$ with the same
16 reading orders, 384 runs per certificate and grid, all decided. Read is
the share read and Wrong the share of wrong verdicts; Change is the share read minus that of the default grid, in points.
For each certificate bold marks the lowest share and underline the second lowest, ties sharing a rank.}
\label{tab:betting-grid}
\centering
\footnotesize
\setlength{\tabcolsep}{4pt}
\begin{tabular}{@{}l l r r r r r r@{}}
\toprule
& & \multicolumn{3}{c}{\method{}-$\Sigma$ (\%)} & \multicolumn{3}{c}{\method{}-$\Sigma$+H (\%)} \\
\cmidrule(lr){3-5}\cmidrule(l){6-8}
Grid & Fractions & Read\,$\downarrow$ & Change & Wrong\,$\downarrow$ & Read\,$\downarrow$ & Change & Wrong\,$\downarrow$ \\
\midrule
Default & $\{0.05, 0.1, 0.2, 0.3, 0.5, 0.7, 0.9\}$ & {\secondbest{38.2}} & +0.0 & 0.5 & \textbf{37.8} & +0.0 & 0.3 \\
Coarse & $\{0.1, 0.3, 0.7\}$ & \textbf{38.1} & $-$0.1 & 0.5 & \textbf{37.8} & +0.0 & 0.5 \\
Small & $\{0.05, 0.1, 0.2\}$ & 39.0 & +0.8 & 0.3 & {\secondbest{39.2}} & +1.4 & 0.3 \\
Large & $\{0.5, 0.7, 0.9\}$ & 40.6 & +2.4 & 0.3 & 39.4 & +1.5 & 0.3 \\
\bottomrule
\end{tabular}
\end{table}

\subsection{Settings, table size and history quality on the chains}
\label{app:chain-results}
\Cref{tab:settings} varies the two constants of \method{}-$\Sigma$+H on the chains and refreshes part of the acceptance
tasks before every round after the first. Every setting of the pseudo-count and of the parent completion keeps the share read
within \ConstantsThreeQuartersHistoryChange{} points of its default, the largest change coming from completing the parent row
after three quarters of a target, concentrated in chains 4 and 6; no setting reads more than \SettingsLargestDrop{} points below
its default. Under refresh the saving of \method{}-$\Sigma$+H over \method{}-$\Sigma$, pooled over runs, is
\RefreshQuarterSaving\% with a quarter of the tasks refreshed (\RefreshQuarterSavingMin\% to \RefreshQuarterSavingMax\% per
chain), \RefreshHalfSaving\% with half and \RefreshFullSaving\% with all, when no paid cell remains.

\begin{table}[!htb]
\caption{The settings of the chains (48 rounds, 8 reading orders, 384 runs per setting): mean share read (\%). \method{}-$\Sigma$ does not use the pseudo-count $\kappa$, and a dash marks those cells. A refreshed task loses its paid cells and its prediction before every round after the first, and its parent cell is paid again. No run returns a wrong verdict.}
\label{tab:settings}
\centering
\small
\setlength{\tabcolsep}{4pt}
\begin{tabular}{@{}l r r@{}}
\toprule
Setting & \method{}-$\Sigma$ & \method{}-$\Sigma$+H \\
\midrule
Defaults: $\kappa=2$, parent row after half, no refresh & 23.2 & 19.4 \\
\midrule
Pseudo-count $\kappa=0.5$ & \textemdash & 19.6 \\
Pseudo-count $\kappa=1$ & \textemdash & 19.5 \\
Pseudo-count $\kappa=4$ & \textemdash & 19.4 \\
Pseudo-count $\kappa=8$ & \textemdash & 19.7 \\
Parent row completed after a quarter & 23.1 & 19.2 \\
Parent row completed after three quarters & 23.9 & 20.4 \\
A quarter of the tasks refreshed & 24.4 & 21.9 \\
Half of the tasks refreshed & 25.5 & 24.0 \\
All of the tasks refreshed & 27.4 & 27.4 \\
\bottomrule
\end{tabular}
\end{table}

\Cref{tab:table_size} cuts each target to its first 375 and 750 tasks. From 375 to 1{,}500 tasks per target the table grows
fourfold and the cells read per round grow by a factor of \SizeSigmaGrowth{} for \method{}-$\Sigma$ and \SizeHistoryGrowth{}
for \method{}-$\Sigma$+H. The share of \method{}-$\Sigma$ does not fall from 375 to 750 tasks, where chain 2 reads
\SizeMediumSigmaShareChainTwo\%, and cutting a table changes which candidates qualify, so the mix of verdicts changes with the
size. The saving of \method{}-$\Sigma$+H over \method{}-$\Sigma$ is positive in every chain at every size.

\begin{table}[!htb]
\caption{Table size on the six chains (48 rounds, 8 reading orders). Each target keeps its first $N$ tasks under one fixed permutation, and the complete-table verdict is recomputed on each truncated table; Qualifying counts the rounds whose truncated table holds a qualifying candidate. Shares are mean shares read (\%), each row's lowest bold and second underlined, with the exact certificate read along the schedule of \method{}-$\Sigma$; Cells is the mean per round; the last columns give the saving of \method{}-$\Sigma$ over the exact certificate in mean cells (\%) and of \method{}-$\Sigma$+H over \method{}-$\Sigma$ with the chains as units (\%, standard error). No run returns a wrong verdict.}
\label{tab:table_size}
\centering
\footnotesize
\setlength{\tabcolsep}{3.4pt}
\begin{tabular}{@{}r r r r r r r r r@{}}
\toprule
& & \multicolumn{3}{c}{Share read (\%)} & \multicolumn{2}{c}{Cells} & \multicolumn{2}{c}{Saving (\%)} \\
\cmidrule(lr){3-5}\cmidrule(lr){6-7}\cmidrule(l){8-9}
$N$ & Qualifying & Exact & $\Sigma$ & $\Sigma$+H & $\Sigma$ & $\Sigma$+H & $\Sigma$ vs.\ exact & $\Sigma$+H vs.\ $\Sigma$ \\
\midrule
375 & 12 & 36.7 & {\secondbest{31.6}} & \textbf{28.7} & 2{,}130 & 1{,}938 & 14.0 & 9.4 (2.3) \\
750 & 11 & 41.2 & {\secondbest{32.1}} & \textbf{28.4} & 4{,}332 & 3{,}831 & 22.2 & 12.0 (2.9) \\
1{,}500 & 7 & 39.9 & {\secondbest{23.2}} & \textbf{19.4} & 6{,}255 & 5{,}232 & 41.9 & 16.7 (2.9) \\
\bottomrule
\end{tabular}
\end{table}

\Cref{tab:history} replays the chains with five predictors that share the recorded ranges, on the unedited tables and on
tables edited to the qualification boundary, where acceptance and the certification of no candidate are both hardest. The
quality of the prediction moves the cost alone. Every one-sided 95\% Clopper--Pearson upper bound of the wrong shares at the
boundary lies below $\alpha/2$, over \BoundaryRuns{} runs.

\begin{table}[!htb]
\caption{\method{}-$\Sigma$+H on the six chains with predictors of different quality, all with the recorded ranges. Unedited: 64 reading orders, 3{,}072 runs per predictor, each fed by the cells its run paid in earlier rounds; Read is the mean share read (\%), Change its change against no prediction (\%) and Wrong the count of wrong verdicts. Boundary: every round edited three ways, under 200 reading orders per edited table; wrong \accept{} is over the two edits in which no candidate qualifies, wrong \nonebetter{} over the edit in which exactly one candidate meets both requirements with no slack (\%). The oracle predicts the true differences and is not ranked.}
\label{tab:history}
\centering
\footnotesize
\setlength{\tabcolsep}{2.6pt}
\begin{tabular}{@{}l r r r r r r r@{}}
\toprule
& \multicolumn{3}{c}{Unedited chains} & \multicolumn{2}{c}{Wrong \accept{}} & \multicolumn{2}{c}{Wrong \nonebetter{}} \\
\cmidrule(lr){2-4}\cmidrule(lr){5-6}\cmidrule(l){7-8}
Predictor & Read\,$\downarrow$ & Change & Wrong & $\alpha=0.2$ & $\alpha=0.05$ & $\alpha=0.2$ & $\alpha=0.05$ \\
\midrule
none (\method{}-$\Sigma$) & {\secondbest{23.4}} & +0.0 & 0 & 0.72 & 0.20 & 5.49 & 1.11 \\
paid history (\method{}-$\Sigma$+H) & \textbf{19.6} & $-$16.1 & 0 & 0.63 & 0.14 & 5.23 & 0.97 \\
history permuted across tasks & 23.7 & +1.3 & 0 & 0.77 & 0.20 & 5.64 & 1.11 \\
adversarial: negated differences & 24.7 & +5.8 & 0 & 0.78 & 0.20 & 5.49 & 1.11 \\
\midrule
oracle: true differences & 12.8 & $-$45.2 & 0 & 0.00 & 0.00 & 0.00 & 0.00 \\
\bottomrule
\end{tabular}
\end{table}

A stale cache breaks the third condition of \Cref{sec:history-conditions}. On the two fresh complete tables of round 2 of
chain 1, with the recorded execution of the parent row as the cache, \StaleDifferLow{} and \StaleDifferHigh{} parent cells
outside the paid set differ between the two executions, so their cached scores give ranges that can exclude the realized
difference. With these narrow ranges the range check halts \StaleHalted{} of the \StaleOrders{} runs on each table before a
verdict, and the remaining run accepts a qualifying candidate before it reads a differing cell. Widening the range of every
stale cell to $[-1,1]$ restores the condition, and every run then accepts a qualifying candidate and reads
\StaleWidenedSavingLow\% to \StaleWidenedSavingHigh\% less than \method{}-$\Sigma$, with no wrong verdict in any arm.

\subsection{Loops at an equal budget and the SWE-bench stream}
\label{app:loop-results}
In the loops of \Cref{tab:gates}, the per-seed differences of the required test accuracy of \method{}-$\Sigma$+H are
\GateLoopCompletePerSeed{} points against full evaluation and \GateLoopGepaScorePerSeed{} points against the GEPA pipeline. A
round of the \method{}-$\Sigma$+H loop, read with growing windows, takes \GateLoopSigmaHMinutesPerRound{} minutes on average
against \GateLoopCompleteMinutesPerRound{} for a complete table, and the GEPA pipeline, which scores only the candidates its
minibatch passes, takes \GateLoopGepaScoreMinutesPerRound{}. \Cref{tab:swe} gives the loops on the SWE-bench Verified stream.

\begin{table}[!htb]
\caption{Loops on the SWE-bench Verified leaderboard: 135 submissions by date, 8 candidates per round, 17 rounds, 512 reading orders per gate. Each gate runs its own loop at level $\alpha_\rho=0.05/(\zeta(2)\rho^2)$ in round $\rho$. Cells count each agent--issue result once per loop, and Share is their share of the cells of full evaluation (\%). Wrong verdicts are over all 8{,}704 runs, one per round and reading order (\%). Final is the mean resolve rate (\%) of the loop's final parent and Range its range over orders. A dash marks a gate whose final parent is the same under every order. A dagger marks the GEPA rules, which we extend with a control target. Among the gates without a wrong verdict, bold marks the lowest share and underline the second lowest; full evaluation reads every cell and is not ranked.}
\label{tab:swe}
\centering
\footnotesize
\setlength{\tabcolsep}{3.4pt}
\begin{tabular}{@{}l r r r r r r@{}}
\toprule
Gate & \stack{Cells}{per loop} & Share\,$\downarrow$ & \stack{Wrong}{\accept{}} & \stack{Wrong}{\nonebetter{}} & Final & Range \\
\midrule
\method{}-$\Sigma$+H & 24{,}224 & \textbf{35.9} & 0.0 & 0.0 & 77.6 & \textemdash \\
\method{}-$\Sigma$ & 25{,}137 & {\secondbest{37.2}} & 0.0 & 0.0 & 77.6 & \textemdash \\
Exact, $\Sigma$ order & 26{,}946 & 39.9 & 0.0 & 0.0 & 77.6 & \textemdash \\
GEPA pipeline$^{\dagger}$~\citep{agrawal2025gepa} & 19{,}161 & 28.4 & 0.0 & 29.9 & 77.5 & 70.4 to 79.2 \\
GEPA gate$^{\dagger}$, 3 tasks & 948 & 1.4 & 26.4 & 17.2 & 76.4 & 45.0 to 79.2 \\
\midrule
Full evaluation & 67{,}500 & 100.0 & 0.0 & 0.0 & 78.8 & \textemdash \\
\bottomrule
\end{tabular}
\end{table}

\subsection{Live tables}
\label{app:live-results}
\Cref{tab:live} gives the cells behind \Cref{fig:live}a; those of the fixed-batch paired test come from its cell audit in
\Cref{app:rollouts}. The fixed-batch paired test \citep{ma2026validation} spends 1{,}536 rollouts on 40 distinct cells of one
candidate per round and, charged per distinct cell, decides nothing. A replay of every shared-cell rule at four budgets, of which
\Cref{tab:budgets} shows the half and the full one, adds the half-space order, the candidate-first order, a
betting screen$^{\dagger}$ after \citet{sea} with wealth $\prod_t(1+0.5d_t)$ and threshold 10 for a zero-mean null per
candidate, and score-and-keep evaluation \citep{zhang2026dgm,robeyns2025sica}. \method{}-E and the parent-first order read the
same cells at every budget on all five tables, so the shared parent row accounts for the cost of \method{}-E. A separate replay on the Qwen3.5 R4 table at its live budget of 80
cells gives the betting screen \MatrixBettingCost{} cells with decisions in \MatrixBettingDecisionRate\% of its orders,
successive halving \MatrixHalvingCost{}, the LUCB race \MatrixRaceCost{} with no decision, and score-and-keep
\MatrixBatchCost{}, past the budget.

\begin{table}[!htb]
\caption{Distinct cells read on the seven complete live tables, the median over 16 reading orders for
the seeded rules. Bold marks the lower of \method{}-$\Sigma$ and the parent-first exact certificate; Saving (\%) is that of
\method{}-$\Sigma$ over the exact certificate.}
\label{tab:live}
\centering
\footnotesize
\setlength{\tabcolsep}{2.3pt}
\begin{threeparttable}
\begin{tabular}{@{}l r r r r r r r@{}}
\toprule
Table & Cells & \method{}-$\Sigma$\,$\downarrow$ & \stack{Exact,}{parent first\,$\downarrow$}
& DISCERN\,$\downarrow$ & Slice-wise\,$\downarrow$ & \stack{BEACON-}{style$^{\dagger}$\,$\downarrow$}
& Saving\,$\uparrow$ \\
\midrule
Granite R1 & 200 & \textbf{130.5} & 154.0 & 200.0 & 30.0$^{\circ}$ & 170.0$^{\circ}$ & 15.3 \\
Granite R2 & 160 & 40.0 & 40.0 & 40.0 & 24.0$^{\circ}$ & 136.0$^{\circ}$ & 0.0 \\
Granite R3 & 260 & \textbf{106.0} & 174.0 & 260.0 & 39.0$^{\circ}$ & 232.5$^{\circ}$ & 39.1 \\
Granite R4 & 220 & \textbf{50.5} & 53.0 & 60.0 & 33.0$^{\circ}$ & 187.0$^{\circ}$ & 4.7 \\
Qwen3.5 R4 & 220 & \textbf{36.5} & 37.0 & 40.0 & 33.0$^{\circ}$ & 187.0$^{\circ}$ & 1.4 \\
Granite R6 & 40 & \textbf{29.0} & 36.0 & 40.0 & 6.0$^{\circ}$ & 38.0$^{\circ}$ & 19.4 \\
Granite R8 & 60 & \textbf{39.5} & 54.0 & 60.0 & 9.0$^{\circ}$ & 54.0$^{\circ}$ & 26.9 \\
\bottomrule
\end{tabular}
\begin{tablenotes}\footnotesize
\item[] $^{\circ}$Decides no run. $^{\dagger}$Built by the authors from the idea of \citet{wan2025beacon}.
\end{tablenotes}
\end{threeparttable}
\end{table}

\begin{table}[!htb]
\caption{Distinct cells per rule on each complete live table at half and all of its cells. Seeded rules report the median over 32 reading orders, with their decision rate and the share of
their decisions that agree with the complete table (\%) in parentheses.}
\label{tab:budgets}
\centering
\footnotesize
\setlength{\tabcolsep}{3.2pt}
\begin{threeparttable}
\begin{tabular}{@{}l r r r r r r r r@{}}
\toprule
& & \multicolumn{3}{c}{Exact certificate} & & & & \\
\cmidrule(lr){3-5}
Table & Budget & \stack{\method{}-E,}{parent first} & \stack{Half-}{space} & \stack{Candidate}{first} & \stack{Betting}{screen$^{\dagger}$}
& \stack{Successive}{halving} & \stack{LUCB}{race} & \stack{Score-}{and-keep} \\
\midrule
Granite R1 & 100 & 100$^{\circ}$ & 100$^{\circ}$ & 100$^{\circ}$ & 43 (6, 0) & 53 (100, 0) & 88$^{\circ}$ (0) & 150$^{\ddagger}$ \\
 & 200 & 154 & 200 & 199 & 45 (9, 0) & 82 (100, 0) & 122$^{\circ}$ (0) & 150 \\
\midrule
Granite R2 & 80 & 40 & 40 & 80$^{\circ}$ & 30 (25, 100) & 44 (100, 100) & 72$^{\circ}$ (0) & 120$^{\ddagger}$ \\
 & 160 & 40 & 40 & 148 & 30 (94, 100) & 70 (100, 100) & 98$^{\circ}$ (0) & 120 \\
\midrule
Granite R3 & 130 & 130$^{\circ}$ & 130$^{\circ}$ & 130$^{\circ}$ & 68$^{\circ}$ (0) & 65 (100, 0) & 110$^{\circ}$ (0) & 195$^{\ddagger}$ \\
 & 260 & 174 & 260 & 254 & 107 (3, 0) & 106 (100, 0) & 155$^{\circ}$ (0) & 195 \\
\midrule
Granite R4 & 110 & 53 & 58 & 110$^{\circ}$ & 42 (75, 100) & 57 (100, 78) & 96$^{\circ}$ (0) & 165$^{\ddagger}$ \\
 & 220 & 53 & 58 & 215 & 42 (100, 100) & 90 (100, 91) & 132$^{\circ}$ (0) & 165 \\
\midrule
Qwen3.5 R4 & 110 & 37 & 38 & 110$^{\circ}$ & 25 (100, 100) & 57 (100, 100) & 96$^{\circ}$ (0) & 165$^{\ddagger}$ \\
 & 220 & 37 & 38 & 204 & 25 (100, 100) & 90 (100, 100) & 132$^{\circ}$ (0) & 165 \\
\bottomrule
\end{tabular}
\begin{tablenotes}\footnotesize
\item[] $^{\circ}$Decides no run. $^{\ddagger}$Cost past the budget. $^{\dagger}$Built by the authors from the idea of \citet{sea}.
\end{tablenotes}
\end{threeparttable}
\end{table}

Under degraded conditions on the Qwen3.5 R4 table and on Granite R3 every verdict of the exact certificate stays sound: a withheld cell ends the run
with a typed failure and no acceptance, a withheld cell outside the schedule changes nothing, a budget below the cost ends in
\undecided{}, and three uniformly random acquisition orders need 150 to 165 and 237 to 252 cells, where \method{}-E needs
\MatrixCertificateCost{} and \SweepNoneBetterCostRoundThree.

\subsection{Exact finite models}
\label{app:finite}
The finite-model studies enumerate every binary score table of a declared setting with its exact probability under the product
prior and run every schedule on each. Fifteen settings with two or three candidates,
one or two targets, one or two cells per row and primitive probabilities in $\{1/5,1/2,4/5\}$ are swept over every budget up to
the full table (\Cref{tab:finite}). The planner of \Cref{app:planner} attains the Bellman optimum on more pairs than any other
schedule. On a three-candidate, two-target setting with candidate probability $1/5$ and parent probability $4/5$, its expected
cost is $358146/78125\approx4.584$ cells against $390146/78125\approx4.994$ for the parent-first order, 8.2\% fewer. The
half-space order specializes Algorithms 1 and 3 of \citet{ghuge2023explainable} to unit weights and costs with
$\epsilon=3/80$.

\begin{table}[!htb]
\caption{Exact finite models: 115 setting and budget pairs over 1{,}920 weighted
tables. At optimum counts the pairs that reach the Bellman optimum $V^\star_\beta$, Largest gap the largest shortfall from it;
bold marks the highest and underline the second highest share.}
\label{tab:finite}
\centering
\small
\setlength{\tabcolsep}{5pt}
\begin{tabular}{l r r r}
\toprule
Schedule & At optimum & Share (\%)\,$\uparrow$ & Largest gap\,$\downarrow$ \\
\midrule
\method{}-E, read-plan replanning & 105 & \textbf{91.3} & $19/125$ \\
Favorable-head order & 104 & {\secondbest{90.4}} & $164/625$ \\
Parent-first order & 102 & 88.7 & $3/5$ \\
Inverted-prior replanning & 97 & 84.3 & $3/5$ \\
Candidate-first order & 58 & 50.4 & $512/625$ \\
Half-space order & 57 & 49.6 & $16/25$ \\
\bottomrule
\end{tabular}
\end{table}

\subsection{Repeated rollouts of the fixed-batch paired test}
\label{app:rollouts}
\begin{figure}[!htb]
\centering
\includegraphics[width=\textwidth]{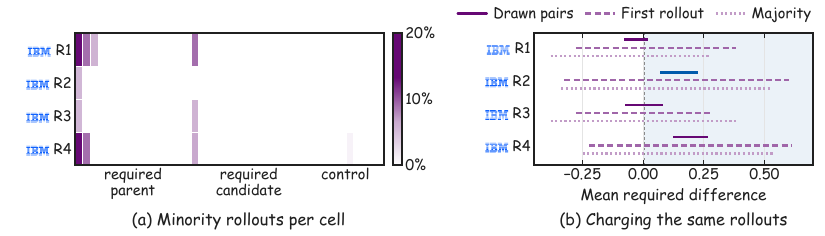}
\caption{Repeated rollouts of the fixed-batch paired test on the four Granite-4.2-8B rounds. (a) Share of rollouts that take the
minority score in each of the 40 distinct cells of a round. (b) The required interval of the test read as drawn pairs, as the
first rollout and as the majority of each cell, blue where the drawn pairs accept; the shaded side of the dashed line
qualifies. The drawn pairs hold \GatesPositive{} positive and \GatesNegative{} negative required disagreements on R1 to R4,
and on R4 one negative control pair of 512 keeps them undecided.}
\label{fig:rollouts}
\end{figure}

An audit executes each of the 40 cells of the fixed-batch paired test on every Granite round many times at temperature zero
(\Cref{fig:rollouts}a); in all 11 unsettled cells, \GatesUnsettled{} on R1 to R4, which return both scores, the first rollout
disagrees with the majority. Under independent rollouts with a minority share of at most $2/11$ that
pattern has probability below $10^{-8}$, so the rollouts of these tests are dependent; the certificate fixes one realized
table of cached scores, and its guarantee holds for that table. A fresh re-execution of the test on Granite R1 executes its cells 11 to 59 times and returns the same four unsettled
cells. Five re-executions of the Qwen3.5 R4, Granite R3 and Granite R4 tables move none of their cells, and the exact certificate
returns the same verdict at \MatrixCertificateCost{}, \SweepNoneBetterCostRoundThree{} and \LiveAcceptCells{} cells under the
majority and under the first rollout. Scored once per distinct cell, the \accept{} of the drawn pairs on Granite R2
becomes \undecided{}, since its 256 required pairs resolve 15 comparisons (\Cref{prop:resampling}, \Cref{fig:rollouts}b); on
the complete tables rounds 2 and 4 improve on the parent and rounds 1 and 3 do not.

\subsection{Re-execution of two complete tables}
\label{app:reexecution}
New executions of the round-4 tables of Qwen3.5-9B and Granite-4.2-8B keep programs, tasks, requirements, model revisions
and evaluator, decode at temperature zero with at most 2{,}048 output tokens and draw new request seeds. Both agree with their
originals on all 220 cells, so four schedules of the exact certificate return the verdicts and costs of \Cref{tab:budgets}; at
the live budget of 80 cells \method{}-E, the parent-first and the half-space orders accept after \ReexecutionQwen{} cells on
the Qwen3.5-9B table and after \ReexecutionGranite{} on the Granite-4.2-8B table, and the candidate-first order returns
\undecided{} on both.

\subsection{Fresh execution of a recorded round}
\label{app:fresh}
\begin{figure}[!htb]
\centering
\includegraphics[width=\textwidth]{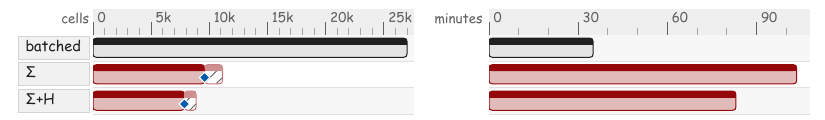}
\caption{Round 2 of chain 1 executed again: the cells each run executes (left) and its wall-clock time (right), for the
batched round and both certificates. Hatched cells were prefetched and left unread; a diamond marks the verdict.}
\label{fig:fresh}
\end{figure}

\Cref{fig:fresh} draws the fresh execution of round 2 of chain 1 of \Cref{sec:exp-live} as two timelines over the same lanes.
Each certificate executes fewer than half the cells of the batched round and, reading in windows of \LiveWindowMin{} to
\LiveWindowMax{} tasks, takes longer in wall-clock time than the recorded batched round. \Cref{tab:cost} runs the same round
again on one H200 GPU per run under one engine setting, with the evaluator cache serving the cells paid in round 1: the
complete table in one batch, twice, and each certificate with windows that grow from \LiveWindowMin{} to \LiveWindowMax{} tasks or with a full first
batch of the first 256 tasks of each target followed by windows that grow from 256 to 1{,}024 tasks. Each run realizes its own table and accepts a candidate that meets
the requirement on that table; candidate c6 lies near the required threshold on the recorded table and falls below it on both
fresh complete tables, and candidate c7 qualifies on all three. The two fresh complete tables differ from the recorded one in
\CostFreshFlipsLow{} and \CostFreshFlipsHigh{} cells, and across eight reading orders on each, both certificates return the
complete-table verdict in all \CostReplayRuns{} runs. Updating the bounds of \method{}-$\Sigma$ takes
\CertificateCpuMillisecondsPerRead{}~ms of single-core CPU time per read.

\begin{table}[!htb]
\caption{End-to-end execution of round 2 of chain 1 on one GPU per run, with round 1's paid cells served from the evaluator cache. Parentheses name the accepted candidate, and executed cells include prefetched cells left unread. Output tokens are in millions and wall-clock time in seconds. A dash marks the complete table, which runs no certificate.}
\label{tab:cost}
\centering
\footnotesize
\setlength{\tabcolsep}{3.4pt}
\begin{tabular}{@{}l r r r r@{}}
\toprule
Run & Cells read & Cells executed & Output tokens & Wall-clock \\
\midrule
complete table, run 1 and run 2 & \textemdash & 25{,}500 & 25.0 and 25.1 & 3{,}212 and 2{,}385 \\
\method{}-$\Sigma$, growing windows (c6) & 9{,}903 & 10{,}744 & 12.4 & 2{,}448 \\
\method{}-$\Sigma$+H, growing windows (c6) & 9{,}359 & 10{,}260 & 11.9 & 2{,}447 \\
\method{}-$\Sigma$, full first batch (c6) & 10{,}292 & 14{,}120 & 15.4 & 1{,}735 \\
\method{}-$\Sigma$+H, full first batch (c7) & 9{,}387 & 12{,}364 & 14.0 & 2{,}200 \\
\bottomrule
\end{tabular}
\end{table}

\end{document}